%% file: arxiv.tex
\documentclass[10pt,twocolumn,letterpaper]{article}
\usepackage{iccv}
\usepackage{times}
\usepackage{epsfig}
\usepackage{graphicx}
\usepackage{amsmath}
\usepackage{amssymb}

\usepackage{multirow} 
\usepackage{multicol} 
\usepackage{makecell} 
\usepackage[dvipsnames]{xcolor} 
\usepackage{colortbl} 
\usepackage{diagbox} 
\usepackage{siunitx}
\usepackage{caption}
\usepackage{enumitem} 
\usepackage{xspace} 
\usepackage[Export]{adjustbox} 
\usepackage{booktabs} 
\usepackage{balance} 

\usepackage{caption} 
\usepackage{tocloft}

\usepackage{titling}
\pretitle{\begin{center}\Large \bf}
\posttitle{\par\end{center} \vskip 0.5em}
\preauthor{\begin{center} \large \begin{tabular}[t]{c}}
\postauthor{\end{tabular}\par\end{center}}
\predate{\begin{center}\large}
\postdate{\par\end{center} \vskip -0.75em}

\usepackage[pagebackref=true,breaklinks=true,colorlinks,bookmarks=false]{hyperref}

\usepackage[capitalize]{cleveref}

\def\eg{\emph{e.g.}\xspace} 
\def\ie{\emph{i.e.}\xspace} 
\def\etc{\emph{etc.}\xspace}

\def\etal{\emph{et al.}\xspace}

\newcommand{\ours}{UncLe-SLAM\xspace}

\newcommand{\boldparagraph}[1]{\vspace{0.3em}\noindent{\bf #1}}

\newcommand{\bplus}{$\boldsymbol{+}$}    

\colorlet{colorFst}{Green!25}       
\colorlet{colorSnd}{SpringGreen!45} 
\colorlet{colorTrd}{Yellow!30}      
\colorlet{colorLow}{darkgray!30}    
\colorlet{colorDeg}{Orange!30}      
\newcommand{\fs}{\cellcolor{colorFst}\bf}   
\newcommand{\nd}{\cellcolor{colorSnd}}      
\newcommand{\rd}{\cellcolor{colorTrd}}      
\newcommand{\degr}{\cellcolor{colorDeg}}          

\usepackage{scrextend} 
\definecolor{gray}{rgb}{0.95,0.95,0.95}
\definecolor{mycol}{rgb}{0.90,0.95,1.0}

\iccvfinalcopy 

\begin{document}

\title{UncLe-SLAM: Uncertainty Learning for Dense Neural SLAM}

\author{
Erik~Sandström$^{1}$\thanks{Equal contribution.} \hspace{3em} 
Kevin Ta$^{1}$\footnotemark[1] \hspace{3em} 
Luc~Van~Gool$^{1,2}$ \hspace{3em} 
Martin~R.~Oswald$^{1,3}$ \\
$^{1}$ETH Zürich, Switzerland \hspace{1.5em}
$^{2}$KU Leuven, Belgium \hspace{1.5em}
$^{3}$University of Amsterdam, Netherlands\\
}

\date{}
\maketitle

\input{tex/abstract.tex}
\input{tex/introduction.tex}

\input{tex/related_work.tex}

\input{tex/background.tex}
\input{tex/method.tex}

\input{tex/experiments.tex}
\input{tex/conclusion.tex}

\clearpage

\title{UncLe-SLAM: Uncertainty Learning for Dense Neural SLAM \\ --- Supplementary Material ---}

\author{
Erik~Sandström$^{1}$\footnotemark[1] \hspace{3em} 
Kevin Ta$^{1}$\footnotemark[1] \hspace{3em} 
Luc~Van~Gool$^{1,2}$ \hspace{3em} 
Martin~R.~Oswald$^{1,3}$ \\
$^{1}$ETH Zürich, Switzerland \hspace{1.5em}
$^{2}$KU Leuven, Belgium \hspace{1.5em}
$^{3}$University of Amsterdam, Netherlands\\
}

\maketitle

\date{}

\input{tex/supplementary_abstract}

\appendix
\input{tex/supplementary}

\clearpage

{\small
\balance
\bibliographystyle{ieee_fullname}
\bibliography{egbib}
}

\end{document}

%% file: tex/abstract.tex
\begin{abstract}
We present an uncertainty learning framework for dense neural simultaneous localization and mapping (SLAM).
Estimating pixel-wise uncertainties for the depth input of dense SLAM methods allows re-weighing the tracking and mapping losses towards image regions that contain more suitable information that is more reliable for SLAM.
To this end, we propose an online framework for sensor uncertainty estimation that can be trained in a self-supervised manner from only 2D input data.
We further discuss the advantages of the uncertainty learning for the case of multi-sensor input.
Extensive analysis, experimentation, and ablations show that our proposed modeling paradigm improves both mapping and tracking accuracy and often performs better than alternatives that require ground truth depth or 3D.
Our experiments show that we achieve a 38\% and 27\% lower absolute trajectory tracking error (ATE) on the 7-Scenes and TUM-RGBD datasets respectively. On the popular Replica dataset using two types of depth sensors, we report an 11\% F1-score improvement on RGBD SLAM compared to the recent state-of-the-art neural implicit approaches.
Source code: \url{https://github.com/kev-in-ta/UncLe-SLAM}.

\end{abstract}


%% file: tex/introduction.tex
\section{Introduction}





Neural scene representations have taken over the 3D reconstruction field by storm~\cite{park2019deepsdf,mescheder2019occupancy,chen2019net,Mildenhall2020NeRF:Synthesis} and have recently also been built into SLAM systems~\cite{Sucar2021IMAP:Real-Time,zhu2022nice,yang2022vox} with excellent results for geometric reconstruction, hole filling, and novel view synthesis.
However, their camera tracking performance is typically inferior to the one of traditional sparse methods~\cite{Campos2021ORB-SLAM3:SLAM} that rely on feature point matching \cite{zhu2022nice,yang2022vox}.
A major difference to sparse methods which focus on a small set of points is that the rendering loss in most dense methods treats all pixels equally although it is plausible that they differ in their amount of useful information for SLAM, due to sensor noise.
In the context of RGBD-cameras, it is well-known that several factors such as surface material type, texture \etc, often affect the sensor's raw output, leading to noisy measurements \cite{gokturk2004time, andersen2012kinect}.
Introducing pixel-wise uncertainties into a dense SLAM approach allows us to model non-uniform weights to focus on tracking and mapping suitable scene parts in a continuous manner. This is akin to the discrete selection of features points in traditional sparse approaches.
\begin{figure}[t]
\centering
\footnotesize
\setlength{\tabcolsep}{1pt}
\renewcommand{\arraystretch}{1}
\includegraphics[valign=c,width=\linewidth]{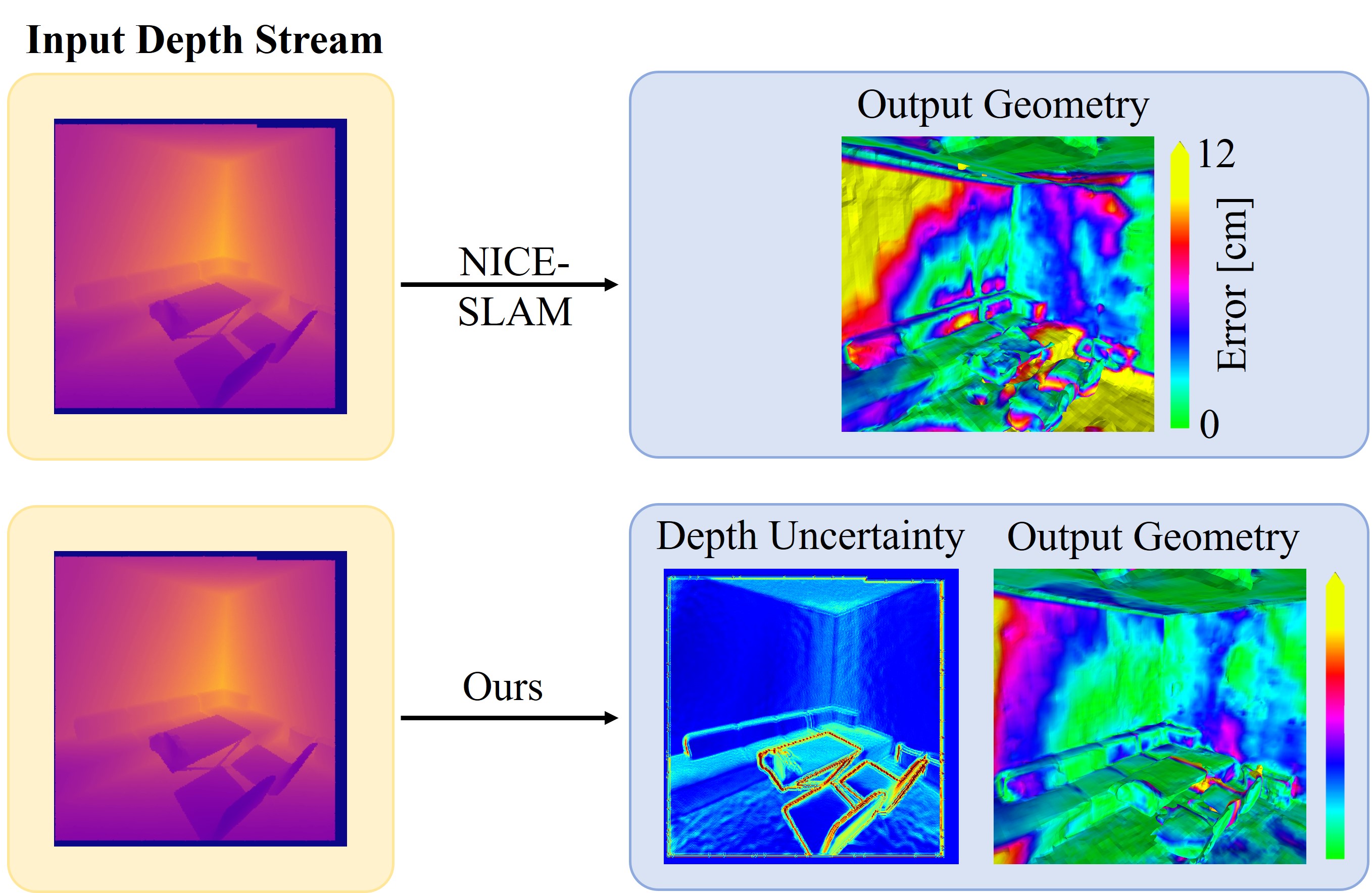} 
\caption{\textbf{\ours benefit.} Our proposed method learns depth uncertainty on the fly in a self-supervised way. We show that our approach yields more accurate 3D mapping and tracking than other dense neural implicit SLAM methods, like NICE-SLAM~\cite{zhu2022nice} which does not model depth uncertainty.}
\label{fig:teaser}
\end{figure}
Currently, the majority of dense neural SLAM approaches employ a uniform weighting for all pixels during mapping ~\cite{zhu2022nice,yang2022vox,mahdi2022eslam,zhu2023nicer} and tracking ~\cite{zhu2022nice,yang2022vox,Sucar2021IMAP:Real-Time,zhu2023nicer}. Some efforts have been made to construct more informed pixel sampling strategies via active resampling or rejection based on the re-rendering loss for mapping~\cite{Sucar2021IMAP:Real-Time} and tracking~\cite{mahdi2022eslam}, but these approaches are ultimately limited by simple heuristics. In this paper, we therefore tackle the task of learning aleatoric depth sensor uncertainty on the fly to weigh scene parts in a non-uniform manner based on the estimated confidence. Furthermore, mobile devices are often equipped with more than one depth sensing modality and it is often observed that different modalities complement each other~\cite{Sandstrom2022LearningFusion}. With these aspects in mind, we design our implicit SLAM system to perform dense SLAM with one or more depth sensors. Additionally, existing depth fusion methods that model single sensor depth uncertainty~\cite{sankowski2017estimation,reynolds2011capturing,poggi2016learning,Weder2020RoutedFusionLR} or fuse multiple depth sensors~\cite{Sandstrom2022LearningFusion} require access to ground-truth depth or 3D at train time. Hence, these methods may not be robust to domain shifts at test time. On the contrary, we learn sensor-agnostic uncertainty online in a self-supervised way without requiring ground truth depth or 3D. For that, we assume a Laplacian error distribution on the depth sensor and derive the corresponding loss function.

Our method, dubbed UncLe-SLAM, jointly learns the aleatoric depth uncertainty and the scene geometry by passing cheaply available 2D features from the depth sensor as input to a small uncertainty decoder, meaning that we stay within real-time run constraints. Our approach thus guides the mapping and tracking process with the implicitly learned uncertainty, see \cref{fig:teaser}. Moreover, we showcase that our formulation generalizes well to the multi-sensor setting where two depth sensors with varying noise distributions are fused into the same 3D representation. Our contributions are:

\begin{itemize}[itemsep=0pt,topsep=2pt,leftmargin=10pt]
    \item A robust approach for estimating aleatoric depth uncertainty for the single and multi-sensor case is proposed. The introduced framework is robust, accurate and can be directly integrated into a dense SLAM system without the need for ground truth depth or 3D.

    \item In the single depth sensor case, we show that our uncertainty-driven approach often improves on standard performance metrics regarding geometric reconstruction and tracking accuracy. In the multi-sensor case, we show for various sensor combinations that our method extracts results that are consistently better than those obtained from the individual sensors.
\end{itemize}

%% file: tex/related_work.tex
\section{Related Work}\label{sec:rel}
The approach proposed in this paper covers a wide range of research topics such as SLAM, sensor fusion, sensor modeling, uncertainty modeling, etc. All of these topics are well-studied with an exhaustive list of literature. Therefore, we narrow our related work discussion to the relevant methods that better helps expose our contributions.

%
%
%
\smallskip
\subsection{Single-Sensor Depth Fusion and Dense SLAM}\label{rel-dense-slam}
Curless and Levoy's seminal work~\cite{curless1996volumetric} is the basis for many dense depth mapping approaches~\cite{newcombe2011kinectfusion, Weder2020RoutedFusionLR}. Subsequent developments include 
scalable techniques with voxel hashing~\cite{niessner2013voxel_hashing,Kahler2015infiniTAM,Oleynikova2017voxblox}, octrees~\cite{6751517}, and pose robustness~\cite{7900211}. Further advancements led to dense SLAM, such as ~\cite{newcombe2011dtam,schops2019bad,Sucar2021IMAP:Real-Time,zhu2022nice}, which can also handle loop closures such as BundleFusion~\cite{dai2017bundlefusion}. To address the issue with noisy depth maps, RoutedFusion~\cite{Weder2020RoutedFusionLR} learns a fusion network that outputs the TSDF update of the volumetric grid. Other works such as NeuralFusion~\cite{weder2021neuralfusion} and DI-Fusion~\cite{huang2021di} extend this concept by learning the scene representation, resulting in better outlier handling. Lately, the work on continuous neural mapping~\cite{yan2021continual} learns the scene representation using continual mapping from a sequence of depth maps. Yet, none of the above-mentioned approaches explicitly study multiple depth modalities or their uncertainty and their fusion in a neural SLAM framework. Further, their extensions to multiple sensor fusion are often not trivial. Nevertheless, by treating all sensors alike, they can be used as simple baselines.

\smallskip
\subsection{Multi-Sensor Depth Fusion}
The fusion of at least two types of depth-sensing devices has been studied in the past. Notably, the fusion of raw depth maps from two different sensors, such as RGB stereo and time-of-flight  (ToF)~\cite{van2012sensor,Choi2012fusion,agresti2017deep,evangelidis2015fusion,dal2015probabilistic,marin2016reliable,agresti2019stereo,deng2021tof}, RGB stereo and Lidar~\cite{maddern2016real}, RGB and Lidar~\cite{qiu2019deepLidar,park2018high,patil2020don}, RGB stereo and monocular depth~\cite{martins2018fusion} and the fusion of multiple RGB stereo algorithms~\cite{poggi2016deep} is well-studied and explored. Yet, these methods study specific sensors and are not inherently equipped with 3D reasoning. Few works consider 3D reconstruction with multiple sensors~\cite{rozumnyi2019learned,kim2009multi,bylow2019combining,yang2020noise,yang2019heterofusion,gu20203d}, but these do not consider the online mapping setting. Conceptually, more closely related to our work is SenFuNet~\cite{Sandstrom2022LearningFusion}, which is an online mapping method for multi-sensor depth fusion. Still, contrary to our approach, \cite{Sandstrom2022LearningFusion} requires access to ground truth 3D data at train time. It does not predict explicit uncertainty per sensor but requires multi-sensor input to weigh the sensors against each other.

\smallskip
\subsection{Uncertainty Modeling for Depth}
Uncertainty modeling for depth estimation has been studied extensively in the past, specifically for multiview stereo (MVS)~\cite{kuhn2020deepc,xu2021digging,zhao2021confidence,su2022uncertainty} and binocular stereo~\cite{poggi2016learning,seki2016patch,tosi2018beyond,kim2019laf}. In addition to the popular Gaussian distribution to model sensor noise~\cite{cao2018real}, the Laplacian noise model has also been employed to analyse depth uncertainty. For instance, Klodt \etal~\cite{klodt2018supervising} assume, like our approach, a Laplacian noise model to explore the advantage of depth uncertainty modeling from short sequences of RGB images. Likewise, Yang \etal~\cite{yang2020d3vo} uses a Laplacian model for monocular depth estimation~\cite{yang2020d3vo}. Furthermore, some works propose self-supervised frameworks for monocular depth estimation, such as~\cite{poggi2020uncertainty, yang2020d3vo}. Aleatoric uncertainty estimation has also been applied for surface normal estimation from RGB~\cite{Bae2021EstimatingEstimation}. This technique was recently used to refine depth estimated from a monocular RGB camera~\cite{Bae2022IronDepth:Uncertainty}. Closer to our setting, RoutedFusion~\cite{Weder2020RoutedFusionLR} trains an encoder-decoder style network to refine depth maps and predict a measure of confidence. Nevertheless, unlike our approach, they require access to ground truth depth for training. Despite impressive progress in depth uncertainty modeling, there has been little focus on uncertainty estimation of the 3D surface. DI-fusion~\cite{huang2021di} proposed a technique to do this by imposing a Gaussian assumption on the signed distance function. Yet, unlike our approach, it needs ground truth 3D for training. 

Regarding uncertainty modeling, our method is related to the treatment of probabilistic depth fusion methods~\cite{duan2012probabilistic,duan2015unified,lefloch2015anisotropic,dong2018psdf,cao2018real}. As studied and observed by several methods, As studied and observed by several methods, explicit uncertainty modeling is helpful\footnote{For a review on uncertainty estimation in deep learning we refer to~\cite{Abdar2021AChallenges}}. In the context of SLAM, Cao \etal~\cite{cao2018real} introduced a probabilistic framework via a Gaussian mixture model for dense visual SLAM based on surfels to address uncertainties in the observed depth. However, it is well-known that Gaussian noise modeling has its practical limitations~\cite{park2013gaussian}.

Overall, to the best of our knowledge, none of the state-of-the-art neural SLAM methods for dense online SLAM consider aleatoric uncertainty modeling along with multiple sensors. Moreover, none of the above works consider estimating uncertainty in an online self-supervised way with implicit neural SLAM.

%% file: tex/background.tex
\section{Preliminaries}\label{sec:background}
To perform online neural implicit SLAM from a sequence of RGBD images, it is necessary to have a 3D representation. Furthermore, due to the self-supervision from the incoming sensor frames, a rendering technique is needed that connects the 3D representation to the 2D observations. By using the 3D representation and 2D rendering technique, the mapping and tracking processes can be constructed. In this paper, we focus on solid (non-transparent) surface reconstruction. We first present background information on implicit surface and volumetric radiance representations, which is then used to develop our online uncertainty modeling approach.

\subsection{Scene Representation}
Convolutional Occupancy Networks \cite{peng2020convolutional} proposes to learn the occupancy $\mathrm{o} \in [0, 1]$ using an encoded 3D grid of features that can be passed, after trilinear interpolation, through an MLP decoder to acquire the occupancy. NICE-SLAM \cite{zhu2022nice} utilizes this idea and encodes the scene in hierarchical voxel grids of features. For any sampled 3D coordinate $\mathbf{p}_i\in\mathbb{R}^3$, feature vectors can be extracted from these voxel grids. The features can then be fed, in a coarse-to-fine manner, through MLP decoders to extract the occupancy of the given point. 

The geometry is encoded in two feature grids -  middle and fine\footnote{There is an additional coarse grid, but it is not used for mapping, and despite claims from the authors, when looking at the source code, it is neither used for tracking. Thus, we do not consider it.}. Each feature grid $\phi^l_\theta$ has an associated pretrained decoder $f^l$, where $l \in \{1,2\}$ and $\theta$ describes the optimizable features. We denote a trilinearly interpolated feature vector at point $\mathbf{p}_i$ as $\phi^l_\theta(\mathbf{p}_i)$. Additionally, the color is encoded in a fourth feature grid $\psi_\omega$ (parameters $\omega$) with decoder $g_\xi$ (parameters $\xi$), and is used for further scene refinement after initial stages of geometric optimization. The observed scene geometry is reconstructed from the middle and fine resolution feature grids, with the fine feature grid output residually added to the middle grid occupancy. In summary, the occupancy $\mathrm{o}_{i}$ and color $\mathbf{c}_i$ are predicted as
\begin{align}
    \mathrm{o}_{i} &= f^1\big(\mathbf{p}_i, \phi^1_\theta(\mathbf{p}_i)\big) + f^2\big(\mathbf{p}_i, \phi^2_\theta(\mathbf{p}_i),\phi^1_\theta(\mathbf{p}_i)\big) \nonumber \\
    \mathbf{c}_i &= g_\xi\big(\mathbf{p}_i, \psi_\omega(\mathbf{p}_i)\big).
    \label{eq:occ-from-net}
\end{align}

\subsection{Depth and Image Rendering}
To link the 3D representation with supervision using 2D RGBD observations, NICE-SLAM uses volume rendering of depth maps and RGB images. This process involves sampled points $\mathbf{p}_i\in\mathbb{R}^3$ at depth $d_i\in\mathbb{R}^1$ along a ray $\mathbf{r}\in\mathbb{R}^3$ cast from origin $\mathbf{O}\in\mathbb{R}^3$, as
\begin{align}
    \mathbf{p}_i = \mathbf{O} + d_i\mathbf{r}, \quad i \in \{1, ..., N\}.
    \label{eq:point-sample}
\end{align}

The occupancies are evaluated along the ray according to \cref{eq:occ-from-net} and volume rendering constructs a weighting function $w_i$ using  \cref{eq:weight-func}. This weight represents the discretized probability that the ray terminates at that particular point.
\begin{align}
    w_i = \mathrm{o}_{i} \prod_{j=1}^{i-1}(1-\mathrm{o}_{j})
    \label{eq:weight-func}
\end{align}

The rendered depth is computed as the weighted average of the depth values along each ray, and equivalently for the color following \cref{eq:rgbd-render} as defined below.
\begin{align}
   \hat{D} = \sum_{i=1}^N w_i d_i, \quad 
   \hat{I} = \sum_{i=1}^N w_i \mathbf{c}_i
   \label{eq:rgbd-render}
\end{align}
This volume rendering method also provides variance from the discretized selection of points. By taking the depth differences with respect to the sensor depth multiplied by the weighting function, a measure of variance can be extracted that is a composite of the model uncertainty and sampling uncertainty, as defined in \cref{eq:epi-unc-render}.
\begin{align} 
   \hat{S}_{D} = \sqrt{\sum_{i=1}^N w_i \big(\hat{D} - d_i \big)^2}
   \label{eq:epi-unc-render}
\end{align}

%% file: tex/method.tex
\section{Method}\label{sec:methods}
\begin{figure*}[t]
  \centering
  \includegraphics[width=\linewidth]{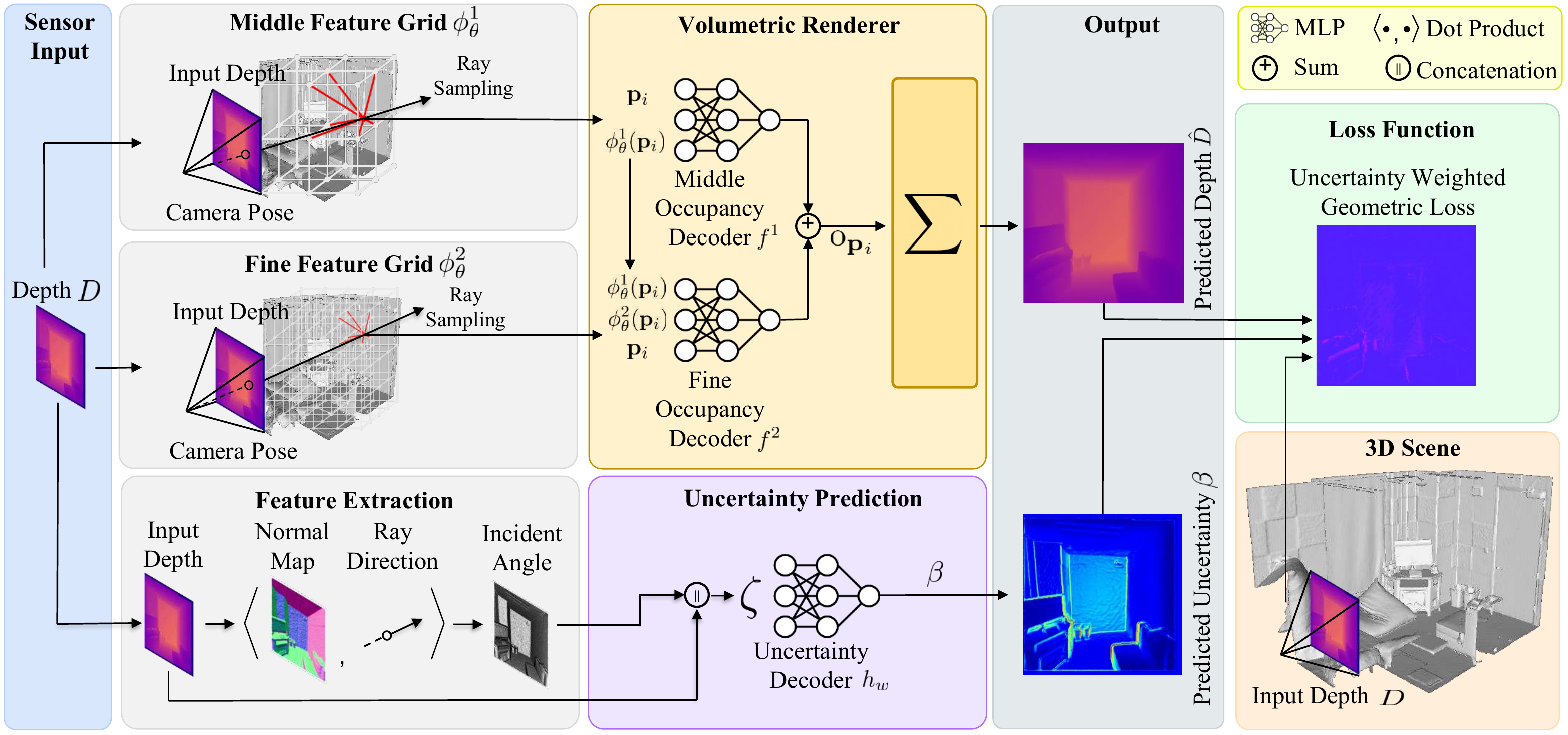}\\
  \caption{\textbf{\ours{} Architecture.} Given an input depth map from an estimated camera pose, mapping and tracking is performed by minimizing a re-rendering loss, by optimizing either the grid features $\theta$ and network parameters $w$ or the camera extrinsics respectively. The depth is estimated using point samples $\mathbf{p}_i$ along rays with a volumetric renderer which decodes geometric multi-scale features $\phi_{\theta}^1(\mathbf{p}_i)$ and $\phi_{\theta}^2(\mathbf{p}_i)$ into occupancies. The uncertainty is estimated by feeding informative features through an uncertainty decoder $h_w$. The architecture can be extended to a multi-sensor setting or with RGB by adding additional uncertainty MLPs. We build the architecture on top of NICE-SLAM~\cite{zhu2022nice}.}
  \label{fig:architecture}
\end{figure*}

This section details how we introduce aleatoric uncertainty modeling based on the preliminaries covered in \cref{sec:background}. 
The rest of our methodology section is arranged as follows: We first present our theoretical assumptions which form the basis for our loss function derivation. Then, we explain how our framework elegantly supports multi-sensor fusion with additional depth sensors and RGBD fusion without relying on heuristic hyperparameters. Finally, we describe our architecture and implementation. For an overview, see \cref{fig:architecture}.

\smallskip
\subsection{Theoretical Assumptions} We motivate our formulation of sensor noise under the assumption of a Laplacian noise distribution on a per-ray basis which was found to perform better on vision tasks than a Gaussian assumption by~\cite{Kendall2017WhatVision}. Further, we assume that the noise is heteroscedastic meaning that the noise variance is a variable for each pixel. That is, each pixel $m$ in the captured depth sensor is treated independently. Consequently, the measured depth is sampled from the probability density function
\begin{equation}
    P(D_m) = \frac{1}{2 \beta_m}\exp\left(-\frac{|D_m - \hat{D}_m|_1}{\beta_m}\right)\enspace.
    \label{eq:pixel-iid-prob}
\end{equation}
We take $\hat{D}_m$ to be the true depth and $\sqrt{2}\beta_m$ to be the standard deviation of the depth reading of a specific pixel, parameterised by some function with parameters $\tau$. When we aggregate all depth sensor information, we get the joint density of the per-ray depth observations
\begin{equation}
    \nonumber P(D_1, ..., D_M)
    = \prod_{m = 1}^M \frac{1}{2 \beta_m} \exp\left(-\frac{|D_m - \hat{D}_m|_1}{\beta_m} \right) \enspace,
    \label{eq:pixel-iid-joint-prob}
\end{equation}
%
where M is the total number of pixel readings. The best estimate of the depth can thus be determined via maximum likelihood estimation 
\begin{align}
    \nonumber
     \arg \max_{\theta, \tau}\ &P(D_1, ..., D_M)
    = \arg \min_{\theta, \tau} -\log\big(P(D_1, ..., D_M)\big) \\ 
    &= \arg \min_{\theta, \tau} \sum_{m = 1}^M \frac{|D_m - \hat{D}_m|_1}{\beta_m} + \log(\beta_m).
    \label{eq:joint-dist-mle}
\end{align}
\subsection{Mapping}
Mapping is performed equivalently to~\cite{zhu2022nice}, but with the revised loss function
\begin{align}
    \mathcal{L}_{map} = \sum_{m = 1}^M \frac{\lvert D_m - \hat{D}_m(\theta) \rvert_1}{\beta_m(\tau)} + \log\big(\beta_m(\tau)\big)
    \label{eq:unc-loss-l1}
\end{align}
A database of keyframes is utilized to regularize the mapping loss. Keyframes are added at a regular frame interval and sampled for each mapping phase to have a significant overlap with the viewing frustum of the current frame. Pixels are then sampled from the keyframes along with the current frame to optimize the map. In terms of optimization, a two-stage approach is taken. For each mapping phase, the middle grid is first optimized and then, once converged, the fine grid is included for further refinement. For more details, we refer to~\cite{zhu2022nice}.

\subsection{Tracking}
Tracking is performed equivalently to~\cite{zhu2022nice}, but with the revised mapping loss function
\begin{align}
  \mathcal{L}_{\mathrm{track}} &= \frac{1}{M_t}
  \sum_{m=1}^{M_t}
  \frac{\lvert D_m - \hat{D}_m(\theta) \rvert_1}
  {\hat{S}_{D}(\theta) + \beta_m(\tau)},
  \label{eq:og-track-depth-loss}
\end{align}
which additionally takes the aleatoric sensor uncertainty into account. $M_t$ is the number of pixels that are sampled during tracking. We optimize the camera extrinsics $\{\mathbf{R}, \mathbf{t}\}$.

\subsection{Multi-Sensor Depth Fusion and RGBD Fusion}
The methods described so far have encompassed implicitly learning uncertainty given a single sensor. We extend this single-sensor approach to incorporate a second sensor. If we again assume that each depth observation is I.I.D., the joint likelihood we wish to maximize is the product of the probability distributions for each pixel in each sensor.

Given two synchronized and aligned sensors, we can sample a set of pixels $m \in \{1,...,M\}$ from two depth sensors yielding the generalized loss function
\begin{align}
    \mathcal{L} = \sum_{m = 1}^M \sum_{i = 1}^2  \frac{\lvert D_{m,i} - \hat{D}_{m} \rvert_1}{ \beta_{m,i}} + \log(\beta_{m,i}).
    \label{eq:two-sensor-l1}
\end{align}
%

One interpretation of this objective function is that the pipeline implicitly learns the weighting between the two sensor observations. The loss function penalizes large uncertainties via the log terms, and implicitly learns the uncertainty for both sets of observations as the model depth is optimized. 
In an analogous fashion, RGBD fusion can be achieved via the loss function
\begin{align}
    \mathcal{L}_{rgbd} &=  \mathcal{L}_{geo} + \mathcal{L}_{rgb} 
    \label{eq:rgbd-loss} \\
    \mathcal{L}_{geo} &= \sum_{m = 1}^M \frac{\lvert D_{m} - 
    \hat{D}_{m} \rvert_1}{ \beta_{m, d}} + \log(\beta_{m, d})  
    \label{eq:d-loss} \\
    \mathcal{L}_{rgb} &= \sum_{m = 1}^M \frac{\lvert I_{m} - \hat{I}_{m} \rvert_1}{ \beta_{m, r}} + \log(\beta_{m, r}),
    \label{eq:rgb-loss}
\end{align}
where $\beta_{m, d}$ and $\beta_{m, r}$ denote the per pixel sensor uncertainty for the depth and rgb sensor respectively.
 This modeling is different to NICE-SLAM where the color and geometry losses are weighted by a heuristic hyperparameter.


\subsection{Design Choices and Architecture Details}
The per-pixel depth and variance is rendered according to \cref{eq:rgbd-render} and \cref{eq:epi-unc-render} respectively.

The variance from \cref{eq:epi-unc-render} could naively be applied to \cref{eq:unc-loss-l1,eq:og-track-depth-loss} with the rendered variance $\hat{S}_D$ representing $2\beta^2$. Unfortunately, such an approach is poorly motivated as this calculated variance is related to the model confidence, as opposed to the sensor-specific noise. In practice, the uncertainty we strive to model is aleatoric uncertainty and should be distinct from the model confidence. One interpretation of the variance from \cref{eq:epi-unc-render} is as the epistemic uncertainty. With an increasing number of observations, the epistemic uncertainty should shrink, driving the model towards sharp bounds. We instead seek a separate process to extract aleatoric uncertainty. We take the concept of implicitly learned aleatoric uncertainty from the work of Kendall and Gal~\cite{Kendall2017WhatVision} and design a patch-based MLP. Our approach takes in spatial information from the specific depth frame to generate uncertainty $\beta$, distinct and decoupled from the rendered variance $\hat{S}_D$.

An additional concern within the framework is the computational overhead. Volume rendering is one of the more intensive operations and an additional rendering for each sensor may be prohibitively expensive. Consequently, we propose a simpler approach to derive a ray-specific uncertainty through the use of 2D features that contain relevant information. We can leverage cheaply available metadata, as was done in \eg~\cite{sayed2022simplerecon}, to capture sensor noise. We investigate plausible per-pixel (per-ray) features and end up with the following inputs to estimate depth uncertainty: the measured depth $D_m \in \mathbb{R}¹$ and the incident angle $\theta \in \mathbb{R}¹$ between the local ray direction and the surface normal, computed as in~\cite{newcombe2011kinectfusion} from the depth map through central difference after bilateral filtering~\cite{tomasi1998bilateral}. For RGB uncertainty, we feed the color instead of the depth and incident angle. Instead of only feeding the features from a single pixel observation, we feed the features from a 5$\times$5 patch, effectively expanding the receptive field of the ray. This patch of pixels gives local context and local correlation of uncertainty for areas near edges or with high frequency content. We denote the concatenation of the features $\zeta$.

 The MLP network, denoted $h_w$, is similar in architecture to the MLPs $f^l$ used for the occupancy decoders. We use a network with 5 intermediate layers with 32 nodes each, activated via ReLU, except for the last layer. Inspired by NeRF-W~\cite{Martin-Brualla2021NeRFCollections}, we apply a softplus activation with a minimum uncertainty value $\beta_{\mathrm{min}}$. The output $\tilde{y}_m \in \mathbb{R}$ from the last layer is thus processed as
\begin{align}
    \beta_m = h_w(\zeta) = \beta_{\mathrm{min}} + \log \left( 1 + \exp \left( \tilde{y}_m \right) \right)
    \label{eq:beta-min}
\end{align}
 The addition of a minimum uncertainty changes the bound of the uncertainty to $(\beta_{\mathrm{min}},\infty)$, and mitigates numerical instability during optimization. Finally, we only update $h_w$ during the fine stage of optimization \ie in the middle stage, we use the same loss as~\cite{zhu2022nice}.

%% file: tex/experiments.tex
\section{Experiments}
\label{sec:exp}
We first describe our experimental setup and then report results on single and multi-sensor experiments. We evaluate our method on the Replica dataset~\cite{straub2019replica} as well as the real-world 7-Scenes~\cite{glocker2013real} and TUM-RGBD~\cite{Sturm2012ASystems} datasets.
All reported results are averages over the respective test scenes and over ten runs, unless otherwise stated.
Further experiments and details are in the supplementary material.

\boldparagraph{Implementation Details.}
We leave many of the hyperparameters from~\cite{zhu2022nice} as is \eg we use 0.32~\si{m} and 0.16~\si{m} voxel size for the middle and fine resolution respectively. The ray sampling strategy remains the same, with 32 points uniformly sampled along the ray and 16 points sampled uniformly near the depth reading. The feature grids store 32-dimensional features and we use the same occupancy decoders and color decoders as~\cite{zhu2022nice}. We leave the learning rates for feature grid optimization under the same schedule\textemdash \ie 0.1 for the middle stage and 0.005 for the fine stage.
On Replica, we map every 5th frame and use 5K pixels uniformly sampled during mapping and tracking. We use 10 tracking iterations and 60 mapping iterations and inlude the fine grid optimization after 60 $\%$ of the total mapping iterations. These parameters were not tuned and may be optimized to further improve performance. Specifically, the learning rates may be adjusted under the new loss formulation to improve stability.

\boldparagraph{Evaluation Metrics.}
The meshes, produced by marching cubes~\cite{lorensen1987marching} from the occupancy grids, are evaluated using the F-score which is the harmonic mean of the Precision (P) and Recall (R). We further provide the mean precision and mean recall along with the depth L1 metric as in~\cite{zhu2022nice}. For tracking accuracy, we use ATE RMSE~\cite{Sturm2012ASystems}.

\boldparagraph{Baseline Methods.} 
We compare our proposed method to existing state-of-the-art online dense neural SLAM methods. The most natural baseline is NICE-SLAM~\cite{zhu2022nice}, which treats all depth observations equally, followed by SenFuNet~\cite{Sandstrom2022LearningFusion}, which performs multi-sensor depth fusion. SenFuNet does not explicitly model per sensor uncertainty, but fuses two depth sensors with a learned weighting network. In the multi-sensor setting, we also compare to Vox-Fusion~\cite{yang2022vox} by weighting all depth readings equally. Additionally, we pretrain a 2D confidence prediction network from the raw depth maps using a slightly modified version of the network proposed by Weder \etal~\cite{Weder2020RoutedFusionLR}. The per pixel learned confidences are used at runtime in NICE-SLAM to scale the importance in the mapping and tracking loss function. We call this baseline ``NICE-SLAM+Pre''. Details are provided in the supplementary material. 

\boldparagraph{Datasets.} 
The Replica dataset~\cite{straub2019replica} comprises high-quality 3D reconstructions of a variety of indoor scenes. We utilize the publicly available dataset collected by Sandström \etal~\cite{Sandstrom2022LearningFusion}, which provides trajectories from a simulated structured light (SL) sensor~\cite{handa2014benchmark}, depth from stereo with semi-global matching~\cite{hirschmuller2007stereo} (SGM) and from a learning-based approach called PSMNet~\cite{chang2018pyramid} as well as color.

The 7-Scenes~\cite{glocker2013real} and TUM-RGBD~\cite{Sturm2012ASystems} datasets comprise a set of RGBD scenes captured with an active depth camera along with ground truth poses.

\begin{table}[tb]
\centering
\setlength{\tabcolsep}{3pt}
\resizebox{\columnwidth}{!}
{
\begin{tabular}{l|lllllll}
\cellcolor{gray}       & \cellcolor{gray}Depth L1$\downarrow$     & \cellcolor{gray}mP$\downarrow$  & \cellcolor{gray}mR$\downarrow$     & \cellcolor{gray}P$\uparrow$ & \cellcolor{gray}R$\uparrow$ & \cellcolor{gray}F$\uparrow$ & \cellcolor{gray}ATE$\downarrow$\\
\multirow{-2}{*}{\cellcolor{gray} ${\text{Model}\downarrow | \text{Metric}\rightarrow}$} & \cellcolor{gray}[cm] & \cellcolor{gray}[cm] & \cellcolor{gray}[cm] & \cellcolor{gray}$[\%]$ & \cellcolor{gray}$[\%]$ & \cellcolor{gray}$[\%]$ & \cellcolor{gray}[cm]\\\hline
\multicolumn{8}{c}{\emph{Depth + Ground Truth Poses}} \\ \hline
NICE-SLAM~\cite{zhu2022nice} & \nd 2.64 & \nd 2.65 & \rd 2.35 & \rd 88.75 & \rd 88.20 & \rd 88.45 & \multicolumn{1}{c}{-} \\
NICE-SLAM+Pre & \rd 2.67 & \nd 2.65 & \nd 2.31 & \nd 89.00 & \nd 88.62 & \nd 88.78 & \multicolumn{1}{c}{-} \\
Ours & \fs \textbf{2.42}  & \fs \textbf{2.58} & \fs \textbf{2.29} & \fs \textbf{89.14} & \fs \textbf{88.70} & \fs \textbf{88.89} & \multicolumn{1}{c}{-} \\ 
\hline
\multicolumn{8}{c}{\emph{Depth + Tracking}} \\ \hline
NICE-SLAM~\cite{zhu2022nice} & \nd 10.65 & \nd 10.04 & \nd 7.17 & \nd 48.46 & \nd 51.43 & \nd 49.80 & \nd 27.90 \\
NICE-SLAM+Pre & \rd 9.90 & \rd 13.99 & \rd 6.84 & \rd 52.43 & \rd 57.72 & \rd 54.54 & \rd 36.95 \\\
Ours & \fs \textbf{7.39} & \fs \textbf{6.56}& \fs \textbf{6.20}& \fs \textbf{57.30} & \fs \textbf{57.57} & \fs \textbf{57.41} & \fs \textbf{19.36} \\ 
\hline
\multicolumn{8}{c}{\emph{RGB-D + Tracking}} \\ \hline
NICE-SLAM~\cite{zhu2022nice} & \nd 8.11 & \nd 7.81 & \nd 6.77 & \nd 51.81 & \nd 53.56 & \nd 52.63 & \nd 20.25 \\
Ours & \fs \textbf{6.49}& \fs \textbf{6.43}& \fs \textbf{5.93}& \fs \textbf{58.89} & \fs \textbf{59.39} & \fs \textbf{59.09} & \fs \textbf{18.92} \\ 
\hline
\end{tabular}
}
\caption{\textbf{{Reconstruction Performance on Replica~\cite{straub2019replica}: PSMNet~\cite{chang2018pyramid}.}} Our model outperforms the baseline methods in the mapping only setting as well as with tracking enabled and when color is available. Best results are highlighted as \colorbox{colorFst}{\bf first}, \colorbox{colorSnd}{second}, and \colorbox{colorTrd}{third}.}
\label{tab:psmnet}
\end{table}

\subsection{Single Sensor Evaluation}
\boldparagraph{Replica.} We provide experimental evaluations on two depth sensors in three different settings: 1. Depth with ground truth poses \ie pure mapping from noisy depth. 2. Depth with estimated camera poses (\ie with tracking) and 3. RGBD with tracking. In \cref{tab:psmnet} for the PSMNet~\cite{chang2018pyramid} sensor, our model shows consistent improvements on all metrics in all three settings. For the SGM~\cite{hirschmuller2007stereo} sensor (in \cref{tab:sgm}) we find consistent improvements in the settings where tracking is enabled. In the mapping only setting, the pretrained confidence model performs marginally better for the SGM sensor. \cref{fig:single_sensor_recon} shows the reconstruction results for two scenes from the Replica dataset with the two sensors. Compared to NICE-SLAM~\cite{zhu2022nice}, we find that \ours on average reconstructs more accurate geometries.

\smallskip
\boldparagraph{Uncertainty Visualization.} To gain insights about the estimated uncertainties that our model produces, we visualize the estimated uncertainties for our two depth sensors in \cref{fig:unc_vis}. For reference, we also plot the absolute ground truth depth error. Compared to the uncertainties produced by the pretrained network, we find that our model produces sharper estimates, see \eg the last row where our model can replicate the error pattern more accurately. This is likely a result of our restricted receptive field while the pretrained model employs a fully convolutional network model with a larger receptive field. Moreover, our model seems to be able to replicate some errors better than the pretrained model, see \eg the red patch for the PSMNet sensor where our model can capture the error while the pretrained model struggles. We believe this is due the the ability of our model to adapt to test time constraints through runtime optimization. Moreover, our network $h_w$ contains only 5409 parameters while the pretrained network contains 360 241.

\begin{table}[tb]
\centering
\setlength{\tabcolsep}{3pt}
\resizebox{\columnwidth}{!}
{
\begin{tabular}{l|lllllll}
\cellcolor{gray}       & \cellcolor{gray}Depth L1$\downarrow$      & \cellcolor{gray}mP$\downarrow$  & \cellcolor{gray}mR$\downarrow$     & \cellcolor{gray}P$\uparrow$ & \cellcolor{gray}R$\uparrow$ & \cellcolor{gray}F$\uparrow$ & \cellcolor{gray}ATE$\downarrow$\\
\multirow{-2}{*}{\cellcolor{gray} ${\text{Model}\downarrow | \text{Metric}\rightarrow}$} & \cellcolor{gray}[cm] & \cellcolor{gray}[cm] & \cellcolor{gray}[cm] & \cellcolor{gray}$[\%]$ & \cellcolor{gray}$[\%]$ & \cellcolor{gray}$[\%]$ & \cellcolor{gray}[cm]\\\hline
\multicolumn{8}{c}{\emph{Depth + Ground Truth Poses}} \\ \hline
NICE-SLAM~\cite{zhu2022nice} & \rd 2.35 & \nd 2.55 & \rd 2.12 & \rd 89.54 & \rd 91.07 & \nd 90.29 & \multicolumn{1}{c}{-} \\
NICE-SLAM+Pre & \fs \textbf{2.25} & \fs \textbf{2.49} & \fs \textbf{2.08} & \fs \textbf{89.86} & \fs \textbf{91.42} & \fs \textbf{90.62} & \multicolumn{1}{c}{-} \\
Ours & \nd 2.27  & \rd 2.56 & \nd 2.10  & \nd 89.59 & \nd 91.24 & \nd 90.40 & \multicolumn{1}{c}{-} \\ \hline 
\multicolumn{8}{c}{\emph{Depth + Tracking}} \\ \hline
NICE-SLAM~\cite{zhu2022nice} & \nd 12.03 & \nd 10.21 & \nd 7.75 & \nd 46.00 & \nd 50.58 & \nd 48.10 & \nd 30.73 \\
NICE-SLAM+Pre & \rd 18.96 & \rd 16.35 & \rd 6.90 & \rd 48.92 & \rd 57.60 & \rd 52.54 & \rd 39.14 \\
Ours & \fs \textbf{10.60}  & \fs \textbf{9.38} & \fs \textbf{6.58} & \fs \textbf{52.62} & \fs \textbf{57.21} & \fs \textbf{54.72} & \fs \textbf{29.11} \\ \hline 
\multicolumn{8}{c}{\emph{RGB-D + Tracking}} \\ \hline
NICE-SLAM~\cite{zhu2022nice} & \nd 9.91 & \fs \textbf{10.37} & \nd 6.82 & \nd 50.12 & \nd 54.51 & \nd 52.00 & \fs \textbf{26.56} \\
Ours & \fs \textbf{7.79} & \nd 11.01 & \fs \textbf{5.80} & \fs \textbf{56.10} & \fs \textbf{61.16} & \fs \textbf{58.19} & \nd 27.41 \\ 
\hline
\end{tabular}
}
\caption{\textbf{Reconstruction Performance on Replica~\cite{straub2019replica}: SGM~\cite{hirschmuller2007stereo}.} Our model outperforms the baseline methods in most settings while being marginally worse than the model using pretrained uncertainties in the mapping only setting.}
\label{tab:sgm}
\end{table}

\begin{figure}[t]
\centering
{\tiny
\setlength{\tabcolsep}{1pt}
\renewcommand{\arraystretch}{1}
\newcommand{\sz}{0.95}
\begin{tabular}{cc}
 &  \hspace{2.8cm} Estimated Uncertainty \hspace{1.65cm} Ground Truth  \\
 & \hspace{-0.2cm} Sensor Depth \hspace{1.1cm} NICE-SLAM+Pre \hspace{0.75cm} Ours \hspace{1.65cm} Depth Error\\[0.0cm]
 \raisebox{-0.75cm}{\rotatebox[origin=c]{90}{\footnotesize PSMNet~\cite{chang2018pyramid}\tiny}}
 & 
\multirow{3}{*}{\includegraphics[width=\sz\linewidth]{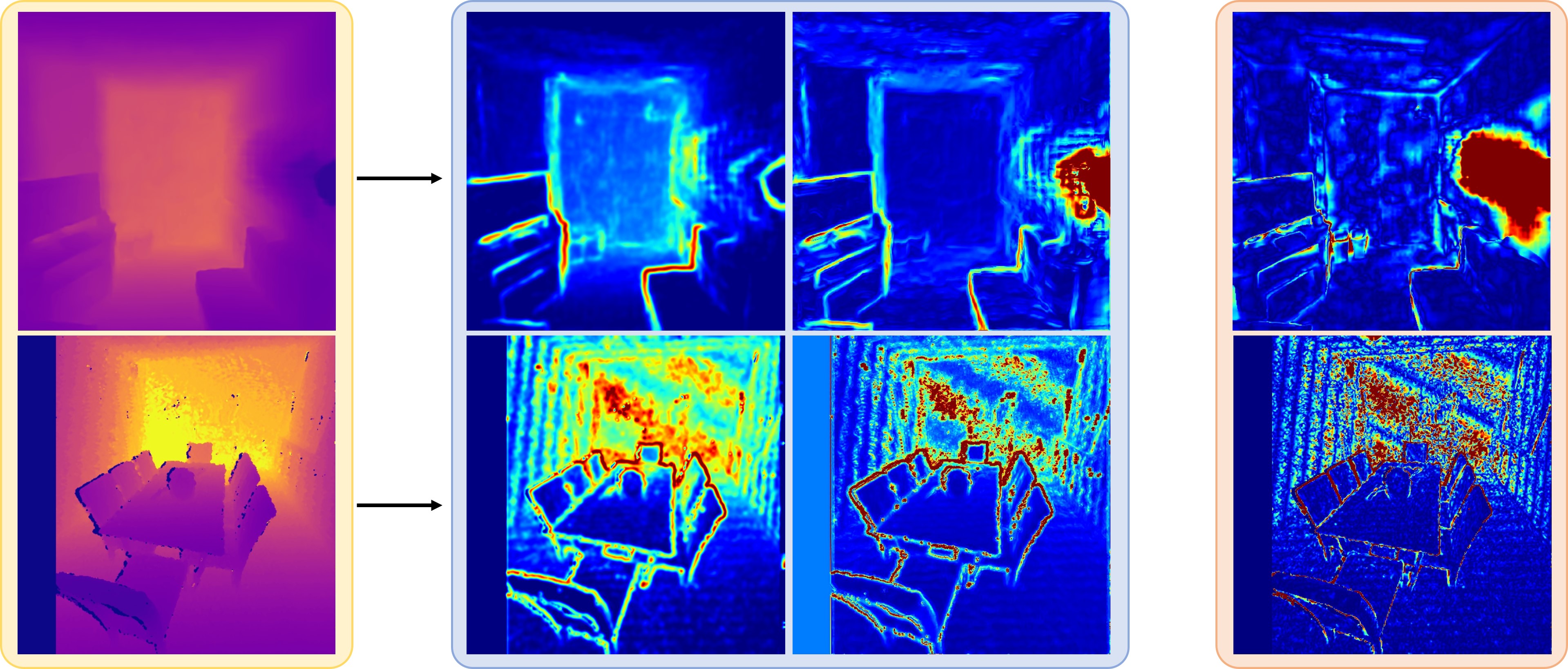}} \\[1.75cm]
\rotatebox[origin=c]{90}{\footnotesize SGM~\cite{hirschmuller2007stereo}\tiny} & 
  \\
\end{tabular}
}
\vspace{0.5cm}
\caption{\textbf{Uncertainty Visualization.} Each row shows a depth map from a specific sensor with the associated uncertainty estimation from the pretrained network model and ours. As reference, the ground truth absolute depth error is shown in the last column. We find our model reproduces the error map with less smoothing than the pretrained model while capturing more details, \eg the red patch from the PSMNet sensor. Blue: low uncertainty, red: high uncertainty.}
\label{fig:unc_vis}
\end{figure}

\begin{figure}[t]
\centering
{\footnotesize
\setlength{\tabcolsep}{1pt}
\renewcommand{\arraystretch}{1}
\newcommand{\sz}{0.29}
\begin{tabular}{ccccc}
\rotatebox[origin=c]{90}{\texttt{Off 0} $|$ \tiny PSMNet\cite{chang2018pyramid}\footnotesize}  & 
\includegraphics[trim={0 2cm 0 1cm},clip=true, valign=c, width=\sz\linewidth]{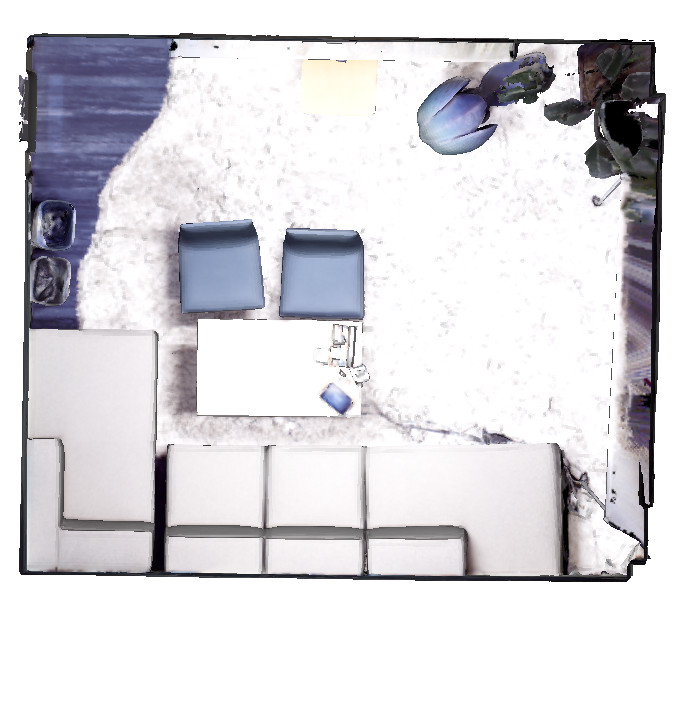} & 
\includegraphics[trim={0 2cm 0 1cm},clip=true, valign=c, width=\sz\linewidth]{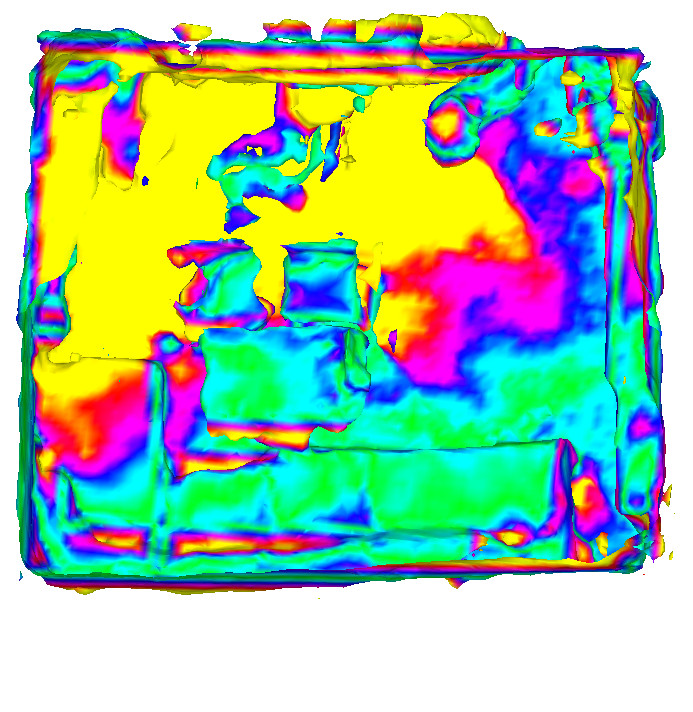} &
\includegraphics[trim={0 2cm 0 1cm},clip=true, valign=c, width=\sz\linewidth]{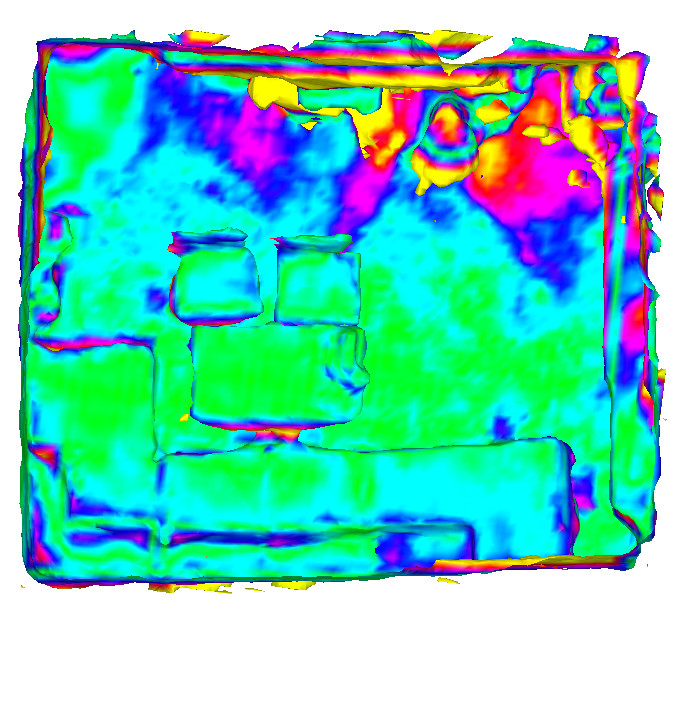} & \multirow{1}{*}[15.0pt]{\includegraphics[width=.1\linewidth]{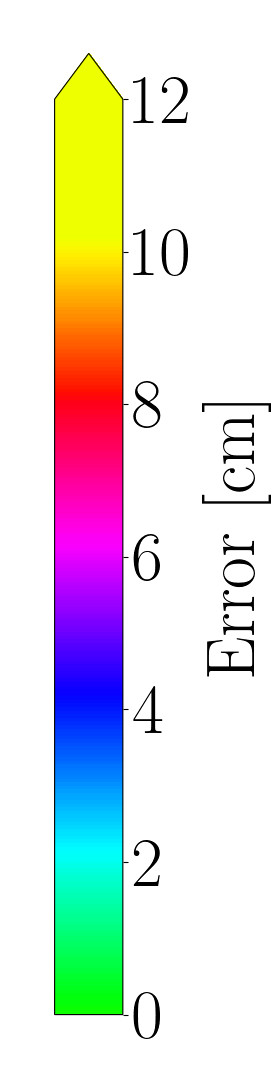}} \\
\rotatebox[origin=c]{90}{\texttt{Rm 2} $|$ \tiny SGM\cite{hirschmuller2007stereo}\footnotesize} & 
\includegraphics[trim={0 1cm 0 1cm},clip=true, valign=c, width=\sz\linewidth]{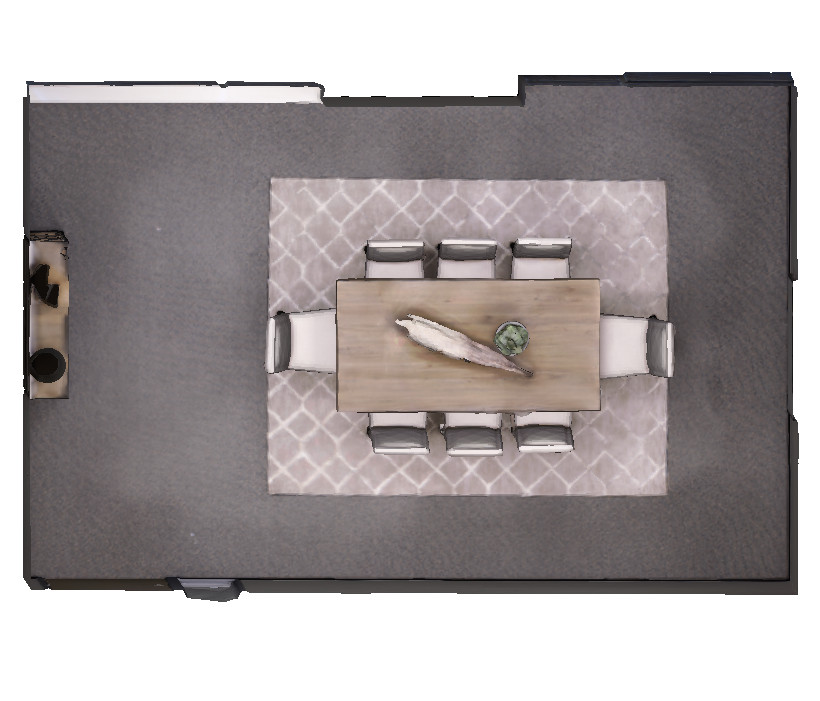} & 
\includegraphics[trim={0 1cm 0 1cm},clip=true, valign=c, width=\sz\linewidth]{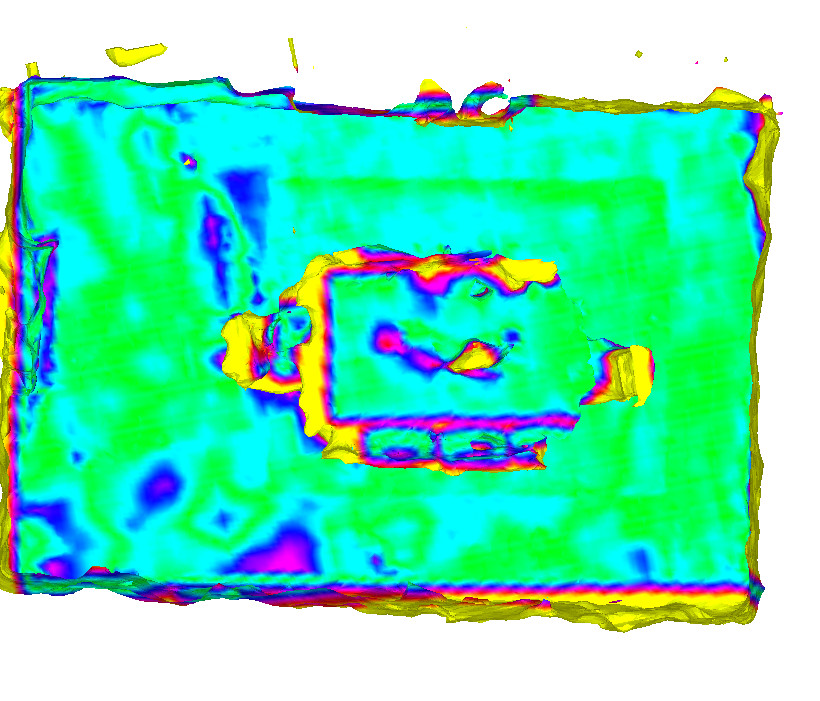} &
\includegraphics[trim={0 1cm 0 1cm},clip=true, valign=c, width=\sz\linewidth]{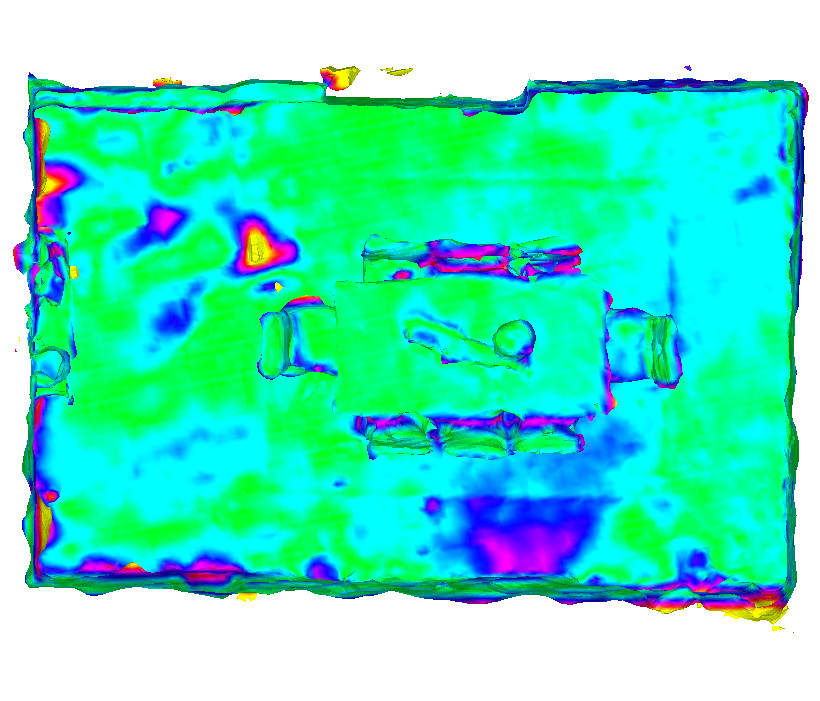} &  \\
 & Ground Truth & NICE-SLAM~\cite{zhu2022nice} & Ours & 
\end{tabular}
}
\caption{\textbf{Single Sensor Reconstruction on Replica~\cite{straub2019replica}.} We show that our uncertainty modeling on average helps to achieve more accurate reconstructions when noisy depth sensors are provided as input. The \texttt{office 0} scene uses only depth as input while the \texttt{room 2} scene is provided RGBD input. Tracking is enabled for all experiments. The colorbar displays the deviation from the ground truth mesh.}
\label{fig:single_sensor_recon}
\end{figure}



\boldparagraph{7-Scenes.}
In \cref{tab:track-7}, we evaluate our framework on the 7-Scenes dataset~\cite{glocker2013real}. We use sequence 1 for all scenes. We find that NICE-SLAM~\cite{zhu2022nice} consistently yields worse tracking results suggesting the effectiveness of our depth uncertainty when it comes to maintaining robust camera pose tracking. On average, our method yields a 38 $\%$ gain in terms of the mean ATE.

\begin{table}[tb]
\centering
\setlength{\tabcolsep}{2pt}
\renewcommand{\arraystretch}{1.05}
\resizebox{\columnwidth}{!}
{
\begin{tabular}{lccccccccc}
\toprule
Method & \texttt{Chess} & \texttt{Fire} & \texttt{Head} & \texttt{Off.} & \texttt{Pump.} & \texttt{Kitch.} & \texttt{Stairs} & Avg.\\
\midrule
\multirow{1}{*}{NICE-SLAM~\cite{zhu2022nice}} & \nd 40.30
& \nd 47.67 & \nd 20.55  & \nd 8.49  & \nd 33.11  & \nd 24.39  &\nd 9.18 & \nd 24.24 \\[0.8pt] \noalign{\vskip 1pt}
\multirow{1}{*}{Ours
} 
& \fs 14.85   & \fs 25.47  & \fs 13.12 &  \fs 7.83 & \fs 29.32  & \fs 6.21 & \fs 8.53 & \fs 15.05 \\
\bottomrule
\end{tabular}
}
\caption{\textbf{Tracking Evaluation on 7-Scenes.} We report the average ATE RMSE [cm] over 5 runs for each scene. With our depth uncertainty modeling, we achieve significantly better tracking compared to NICE-SLAM. On average, our method yields a 38 $\%$ gain in terms of the mean ATE.}
\label{tab:track-7}
\end{table}

\boldparagraph{TUM-RGBD.}
In \cref{tab:track-TUM}, we evaluate our framework on the real-world TUM-RGBD dataset~\cite{Sturm2012ASystems}. Our conclusions on this dataset is similar to the 7-Scenes dataset. On average, camera pose tracking is greatly benefited by our uncertainty aware strategy.

\begin{table}[tb]
\centering
\setlength{\tabcolsep}{2pt}
\renewcommand{\arraystretch}{1.05}
\resizebox{0.75\columnwidth}{!}
{
\begin{tabular}{lllll}
\toprule
   \multirow{2}{*}{Method}& \texttt{fr1/} &  \texttt{fr1/} & \texttt{fr1/} & \multirow{2}{*}{Avg.} \\
  & \texttt{desk} &  \texttt{desk2} & \texttt{xyz} & \\ 
\midrule
NICE-SLAM~\cite{zhu2022nice} & \nd 40.40  & \nd 47.81  & \nd 5.11 & \nd 31.11 \\
Ours & \fs 29.04  & \fs 36.57  & \fs 2.71 & \fs 22.77 \\
\bottomrule
\end{tabular}
}
\caption{\textbf{Tracking Evaluation on TUM-RGBD.} We report the average ATE RMSE [cm] by mapping every 2nd frame. 
}
\label{tab:track-TUM}
\end{table}

\subsection{Multi-Sensor Evaluation}
We conduct experiments in the multi-sensor setting. We compare to Vox-Fusion~\cite{yang2022vox}, a dense neural SLAM system and SenFuNet~\cite{Sandstrom2022LearningFusion}, which is a mapping only framework. To learn sensor specific uncertainties, we use one uncertainty decoder $h_w$ per sensor. In \cref{tab:sgm_psmnet} we show for SGM\bplus{}PSMNet fusion that we are able to consistently improve over the single-sensor reconstructions in isolation and over SenFuNet~\cite{Sandstrom2022LearningFusion} and VoxFusion~\cite{yang2022vox}. When ground truth poses are provided, we find that original NICE-SLAM performs very similar to our proposed uncertainty aware model. On a closer look, the PSMNet and SGM sensors are quite similar and we believe that when both sensors yield similar depth characteristics, simple averaging works well, \ie putting equal weight to both sensors as done by NICE-SLAM. We find, however, that uncertainty modeling is very important to obtain robust tracking which greatly improves the reconstruction accuracy. Finally, \cref{fig:multi_sensor_recon} shows visualizations of the reconstruction accuracy comparing the single sensor reconstructions to the geometry attained by \ours. We find that the most accurate sensor is on average favored. For more results, see the supplementary material.

\begin{table}[tb]
\centering
\setlength{\tabcolsep}{3pt}
\resizebox{\columnwidth}{!}
{
\begin{tabular}{l|lllllll}
\cellcolor{gray}       & \cellcolor{gray}Depth L1$\downarrow$      & \cellcolor{gray}mP$\downarrow$  & \cellcolor{gray}mR$\downarrow$     & \cellcolor{gray}P$\uparrow$ & \cellcolor{gray}R$\uparrow$ & \cellcolor{gray}F$\uparrow$ & \cellcolor{gray}ATE$\downarrow$ \\
\multirow{-2}{*}{\cellcolor{gray} ${\text{Model}\downarrow | \text{Metric}\rightarrow}$} & \cellcolor{gray}[cm] & \cellcolor{gray}[cm] & \cellcolor{gray}[cm] & \cellcolor{gray}$[\%]$ & \cellcolor{gray}$[\%]$ & \cellcolor{gray}$[\%]$ & \cellcolor{gray}[cm] \\\hline
\multicolumn{8}{c}{\emph{Single Sensor Ours: Depth + Ground Truth Poses}} \\ \hline
PSMNet~\cite{chang2018pyramid} & 2.42  & 2.58 & 2.29 & 89.14 & 88.70 & 88.89 & - \\ 
SGM~\cite{hirschmuller2007stereo}    & 2.27  & 2.56 & 2.10  & 89.59 & \fs \textbf{91.24} & \rd 90.40 & - \\ 
\hline
\multicolumn{8}{c}{\emph{Multi-Sensor: Depth + Ground Truth Poses}} \\ \hline
NICE-SLAM~\cite{zhu2022nice} & \nd  2.03 & \fs 2.34 & \fs \textbf{1.99} & \fs \textbf{90.57} & \nd 90.86 & \fs \textbf{90.69} & - \\
SenFuNet~\cite{Sandstrom2022LearningFusion} & 23.49  & 15.62   & 12.66  & 32.74 & 28.32 & 30.22 & -\\ 
Vox-Fusion~\cite{yang2022vox} & 6.52  & 48.76  & 30.72  & 28.01 & 49.36 & 35.65 & -\\ 
NICE-SLAM+Pre & \rd 2.19 & \rd 2.44 & \nd 2.01 & \rd 89.93 & \rd 90.76 & 90.31 & - \\
Ours & \fs \textbf{1.97}  & \nd 2.36 & \nd 2.01  & \nd 90.15 & \rd 90.76 & \nd 90.42 & - \\ 
\hline
\multicolumn{8}{c}{\emph{Single Sensor Ours: Depth + Tracking}} \\ \hline
PSMNet~\cite{chang2018pyramid} & \nd 7.39 & \nd 6.56 & \rd 6.20 & \rd 57.30 & \rd 57.57 & \rd 57.41 & \fs 19.36 \\ 
SGM~\cite{hirschmuller2007stereo} & \rd 10.60  & \rd 9.38 &  6.58 & 52.62 & 57.21 & 54.72 & \rd 29.11 \\ \hline 
\multicolumn{8}{c}{\emph{Multi-Sensor: Depth + Tracking}} \\ \hline
NICE-SLAM~\cite{zhu2022nice} & 13.58 & 16.76 & 7.84 & 51.19 & 55.45 & 52.81 & 40.37 \\
NICE-SLAM+Pre & 11.29  &  13.59 & \nd 6.12 & \nd 62.02 & \nd 65.95 & \nd 63.30 & 35.55 \\
Ours & \fs 4.13 & \fs 4.60 & \fs 4.35 & \fs 70.30 & \fs 69.30 & \fs 69.76 & \nd 19.88 \\
\hline
\end{tabular}
}
\caption{\textbf{Reconstruction Performance on Replica~\cite{straub2019replica}: SGM~\cite{hirschmuller2007stereo}\bplus{}PSMNet~\cite{chang2018pyramid}.} Our multi-sensor reconstruction performance improves over the single sensor results in isolation and we outperform most of the baseline methods. The experiment was conducted in the depth only setting with known camera poses.}
\label{tab:sgm_psmnet}
\end{table}

\begin{figure}[t]
\centering
{\footnotesize
\setlength{\tabcolsep}{1pt}
\renewcommand{\arraystretch}{1}
\newcommand{\sz}{0.21}
\begin{tabular}{cccccc}
\multirow{1}{*}[-8.5pt]{\rotatebox[origin=c]{90}{\texttt{Office 0}}}
& 
\includegraphics[trim={0 0cm 0 0cm},clip=true, valign=c, width=\sz\linewidth]{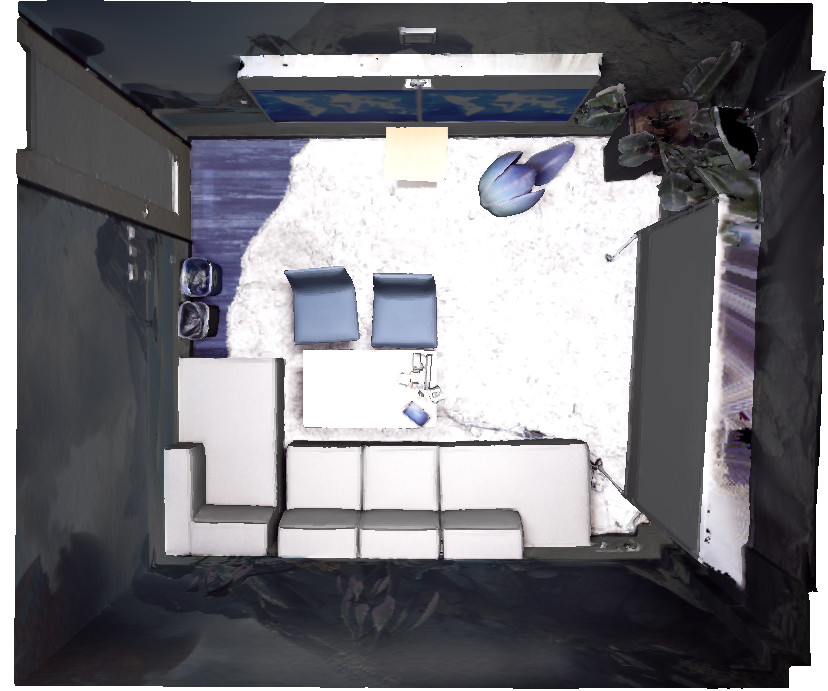} & 
\includegraphics[trim={0 0cm 0 0cm},clip=true, valign=c, width=\sz\linewidth]{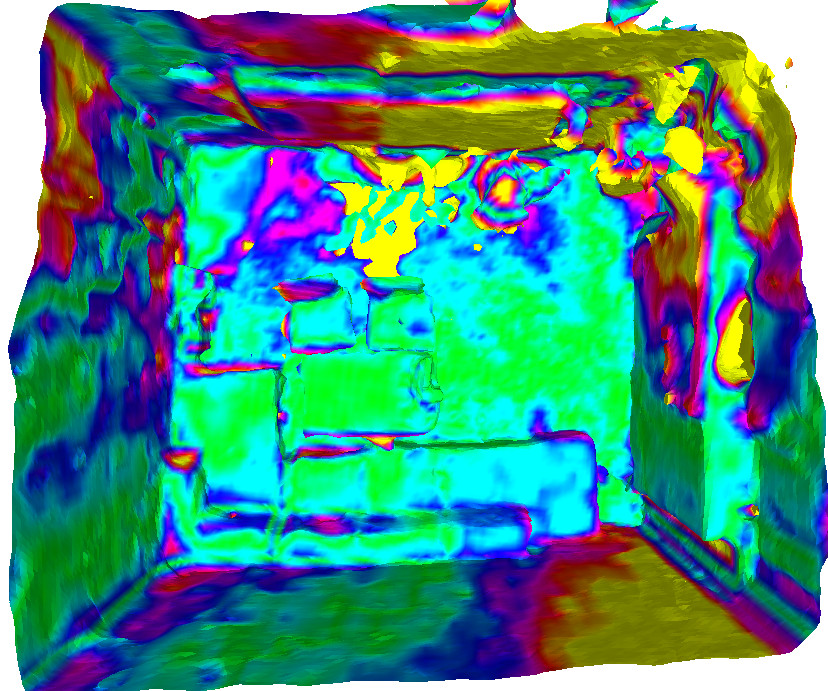} &
\includegraphics[trim={0 0cm 0 0cm},clip=true, valign=c, width=\sz\linewidth]{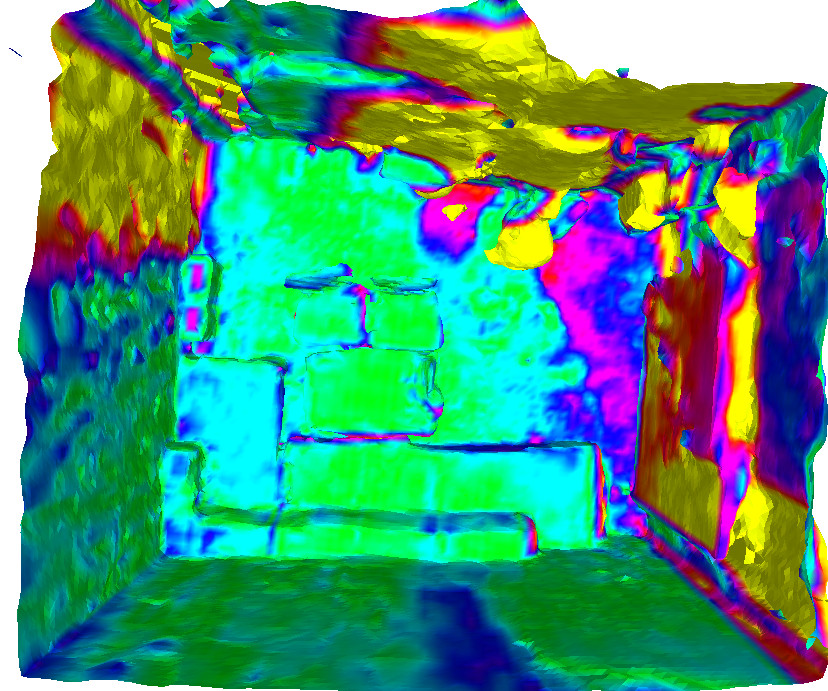} & 
\includegraphics[trim={0 0cm 0 0cm},clip=true, valign=c, width=\sz\linewidth]{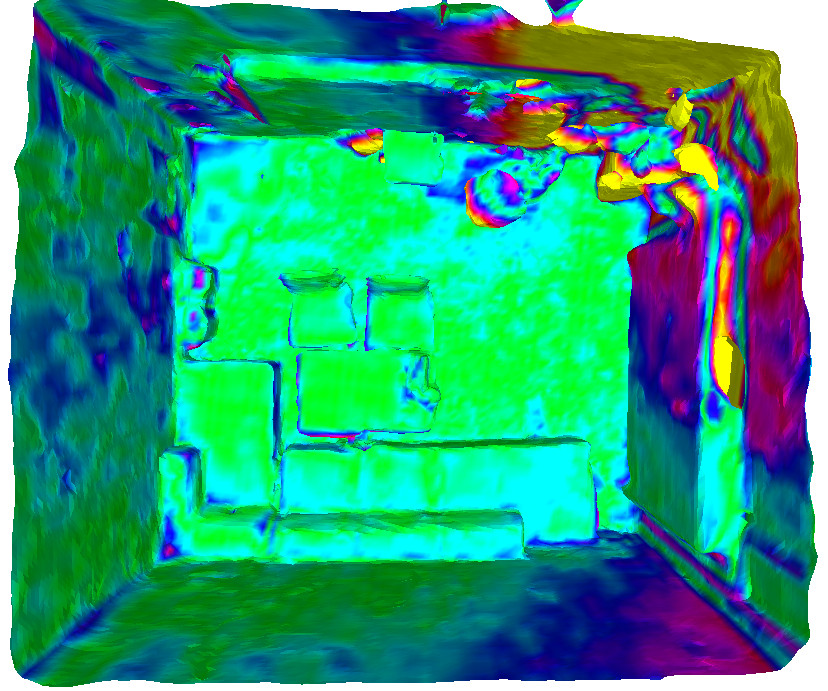} &
\multirow{1}{*}[-58.5pt]{\includegraphics[width=.1\linewidth]{figs/colorbar_outlier_filter.jpg}} \\
&  \includegraphics[trim={0 0cm 0 0cm},clip=true, valign=c, width=\sz\linewidth]{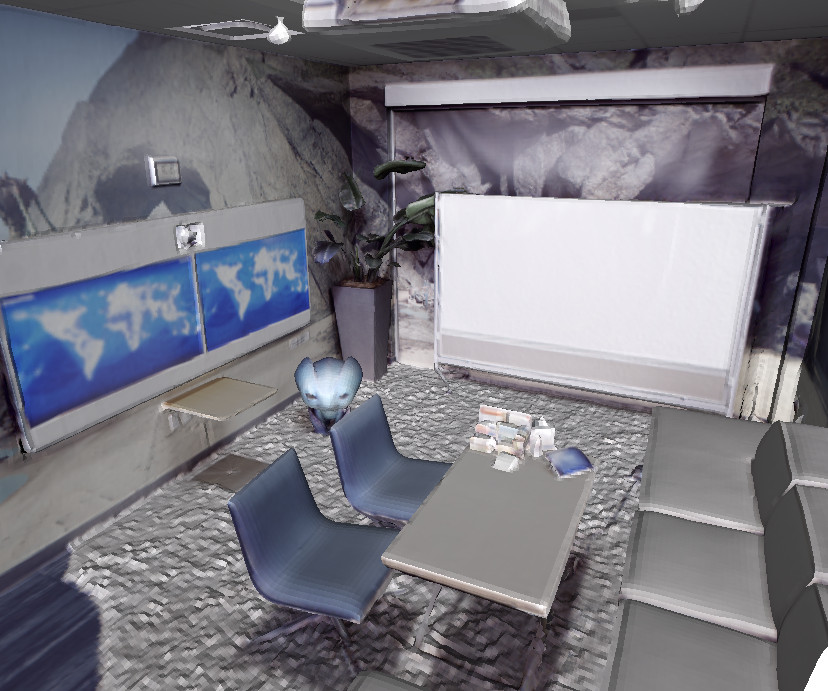} & 
\includegraphics[trim={0 0cm 0 0cm},clip=true, valign=c, width=\sz\linewidth]{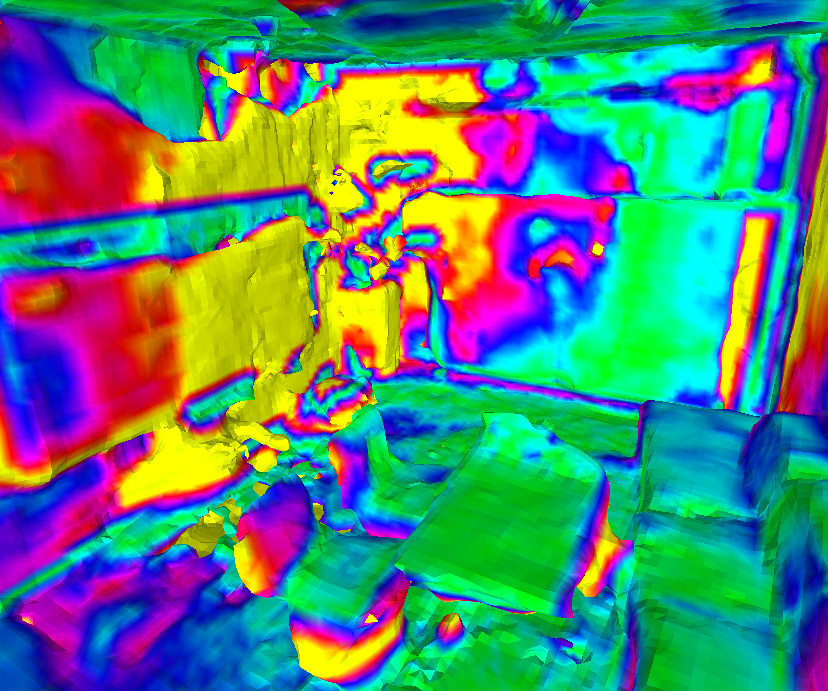} &
\includegraphics[trim={0 0cm 0 0cm},clip=true, valign=c, width=\sz\linewidth]{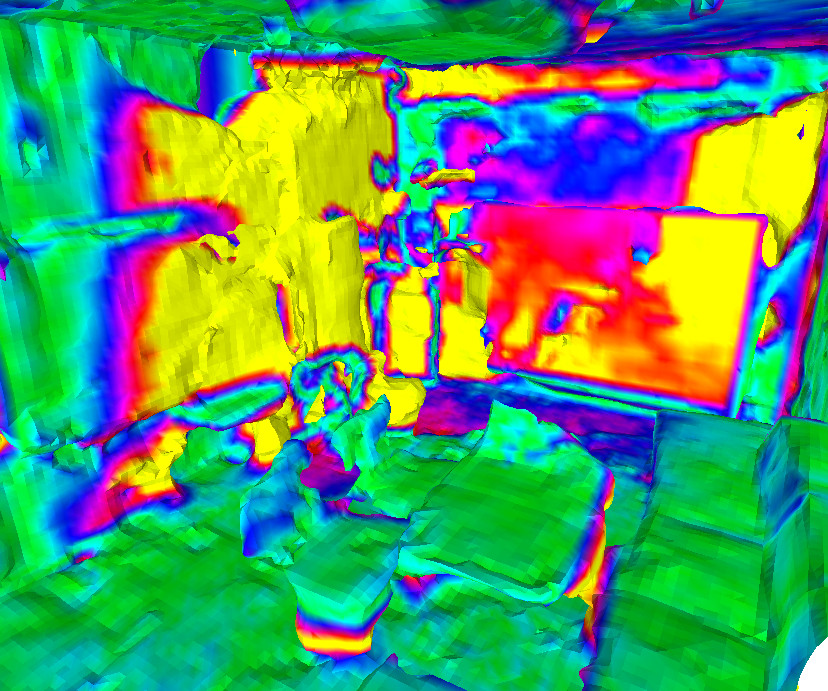} & 
\includegraphics[trim={0 0cm 0 0cm},clip=true, valign=c, width=\sz\linewidth]{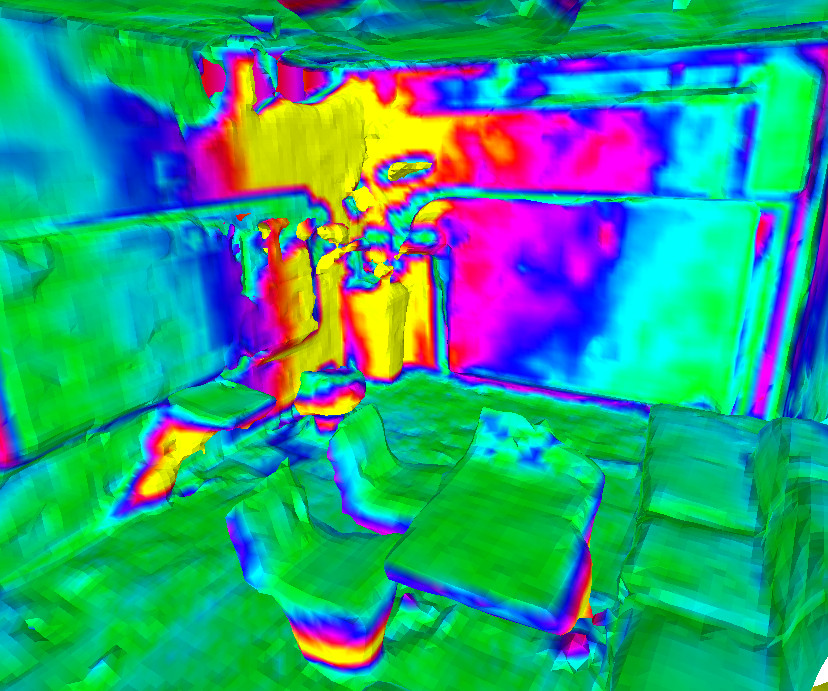} & \\
\multirow{1}{*}[-8.5pt]{\rotatebox[origin=c]{90}{\texttt{Office 1}}} & 
\includegraphics[trim={0 0cm 0 0cm},clip=true, valign=c, width=\sz\linewidth]{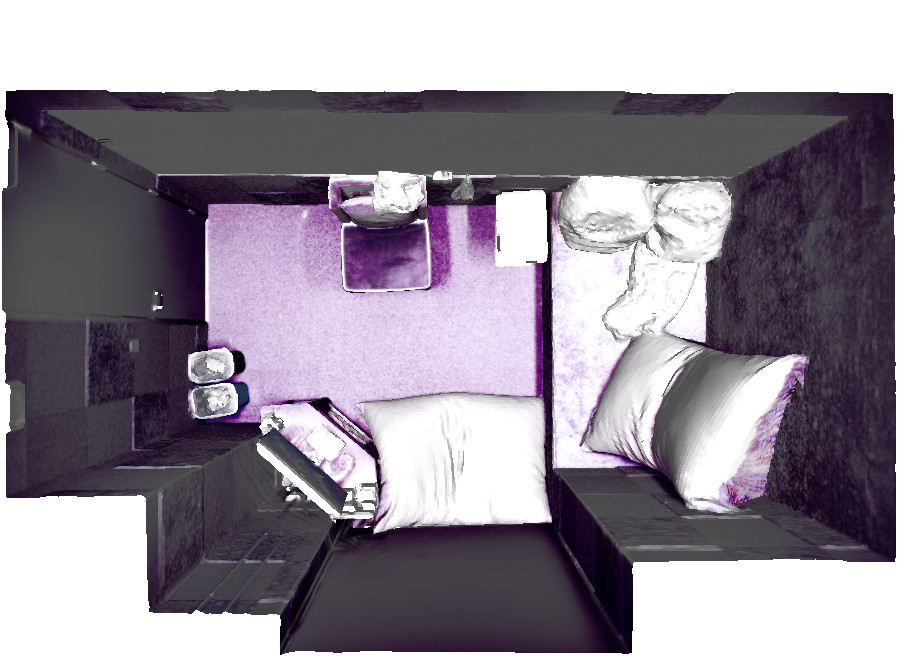} & 
\includegraphics[trim={0 0cm 0 0cm},clip=true, valign=c, width=\sz\linewidth]{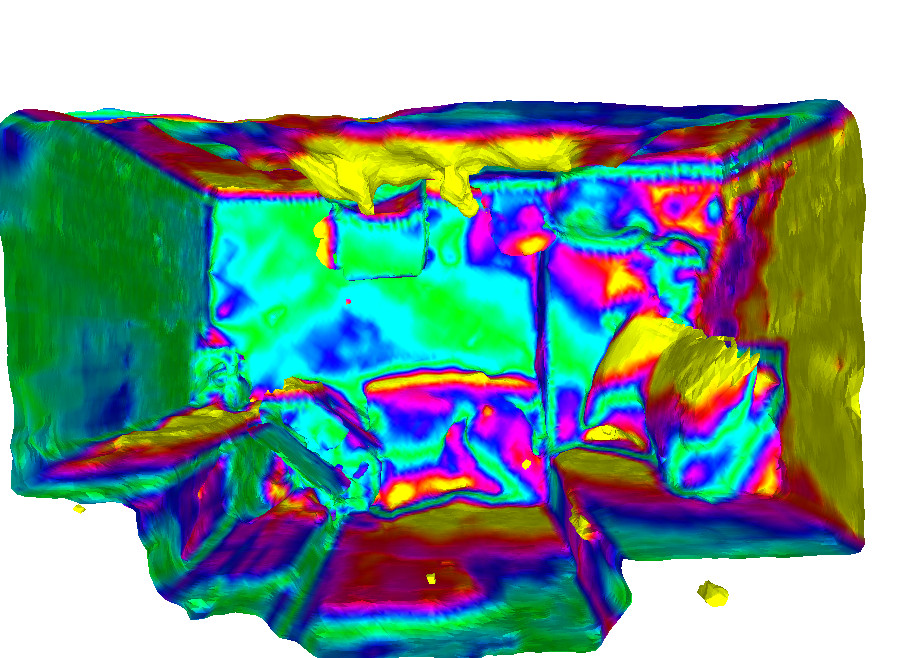} &
\includegraphics[trim={0 0cm 0 0cm},clip=true, valign=c, width=\sz\linewidth]{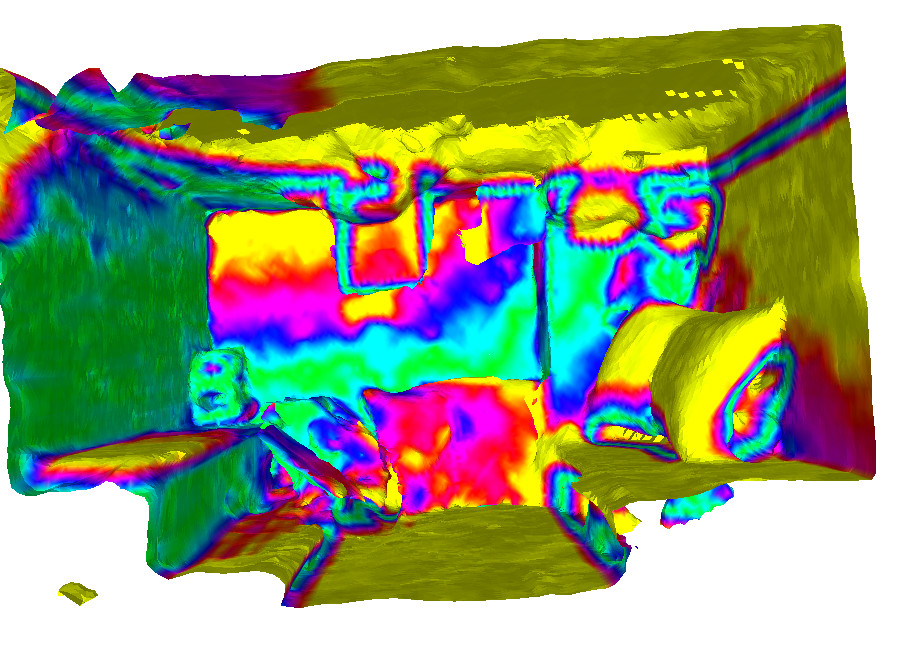} & 
\includegraphics[trim={0 0cm 0 0cm},clip=true, valign=c, width=\sz\linewidth]{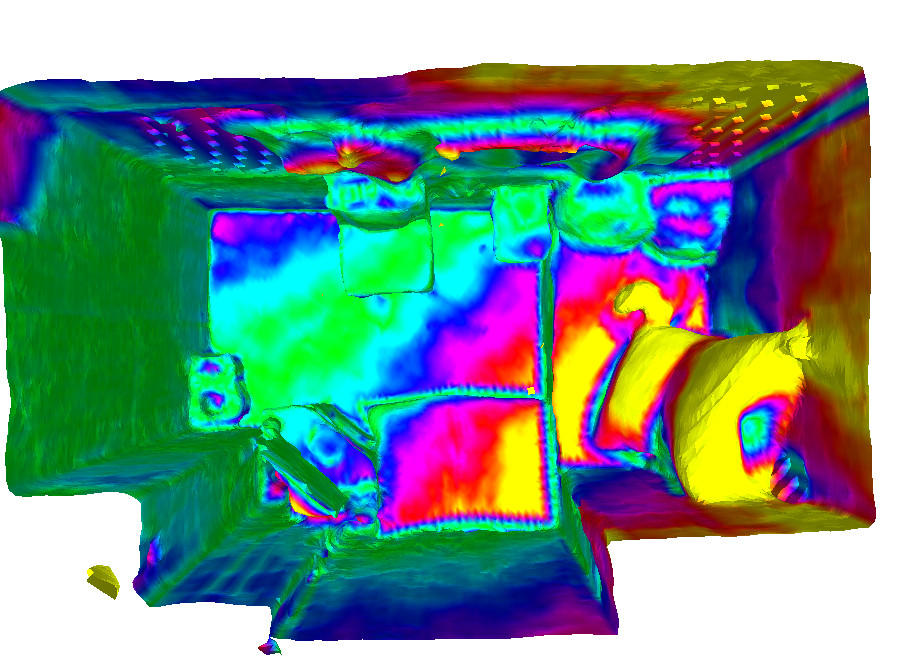} & \\
&  \includegraphics[trim={0 0cm 0 0cm},clip=true, valign=c, width=\sz\linewidth]{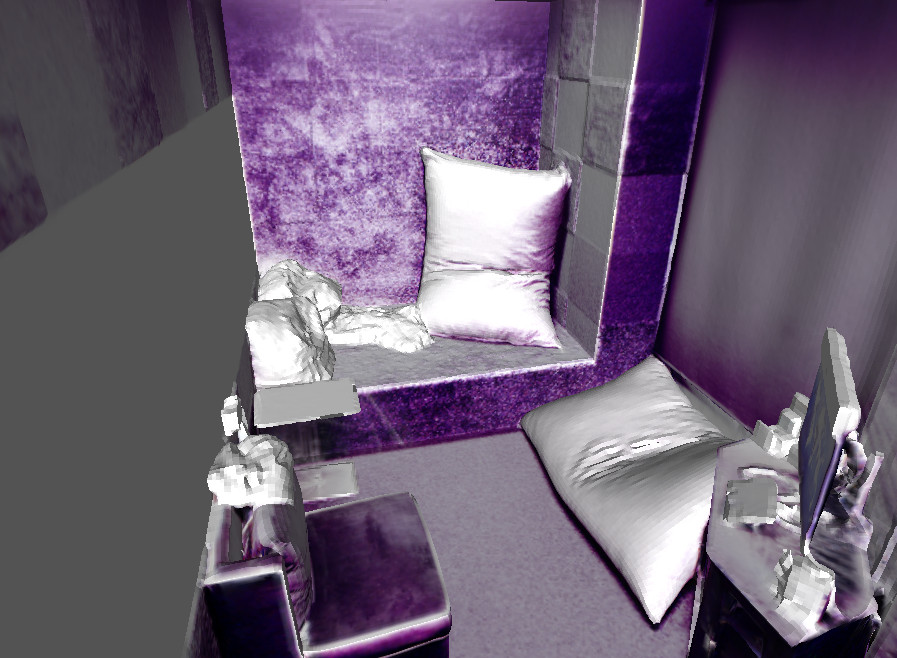} & 
\includegraphics[trim={0 0cm 0 0cm},clip=true, valign=c, width=\sz\linewidth]{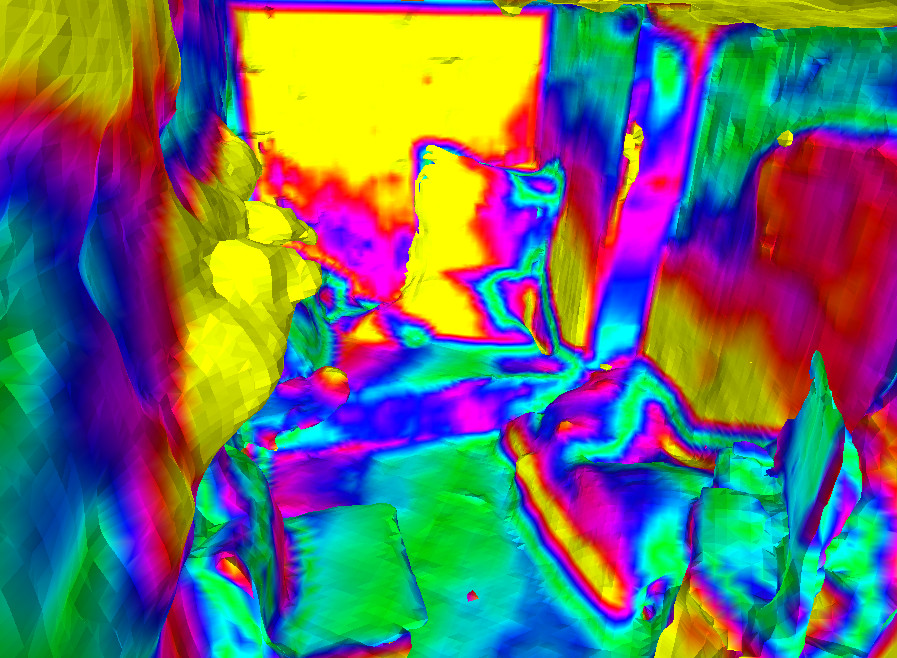} &
\includegraphics[trim={0 0cm 0 0cm},clip=true, valign=c, width=\sz\linewidth]{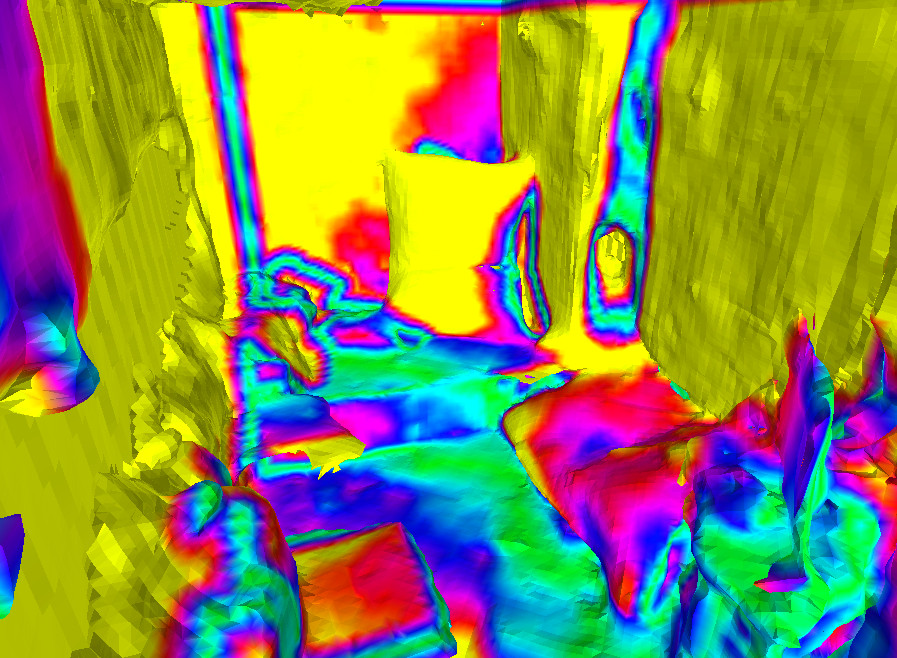} & 
\includegraphics[trim={0 0cm 0 0cm},clip=true, valign=c, width=\sz\linewidth]{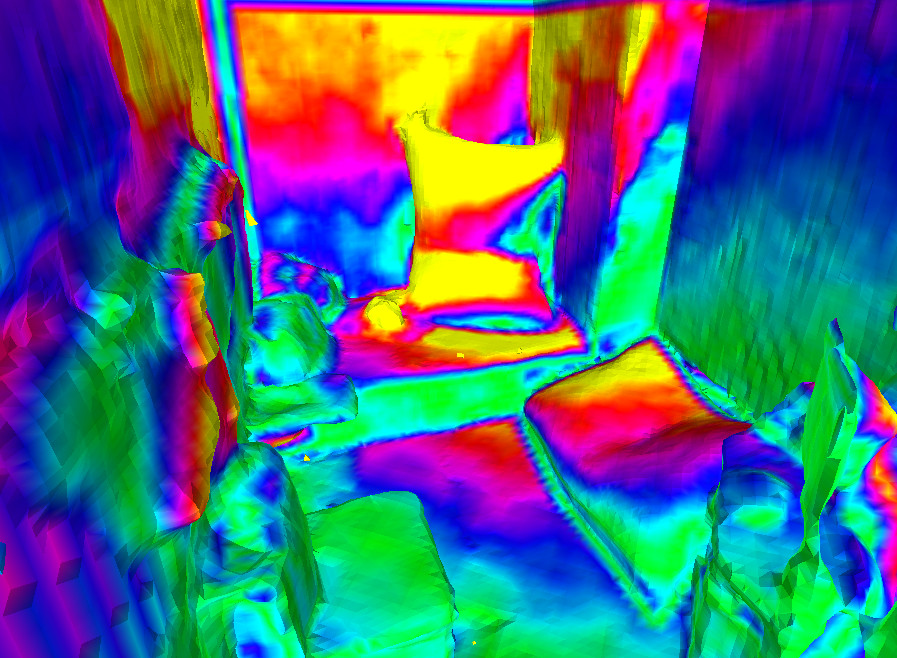} & \\
\multirow{1}{*}[-8.5pt]{\rotatebox[origin=c]{90}{\texttt{Room 2}}} & 
\includegraphics[trim={0 0cm 0 0cm},clip=true, valign=c, width=\sz\linewidth]{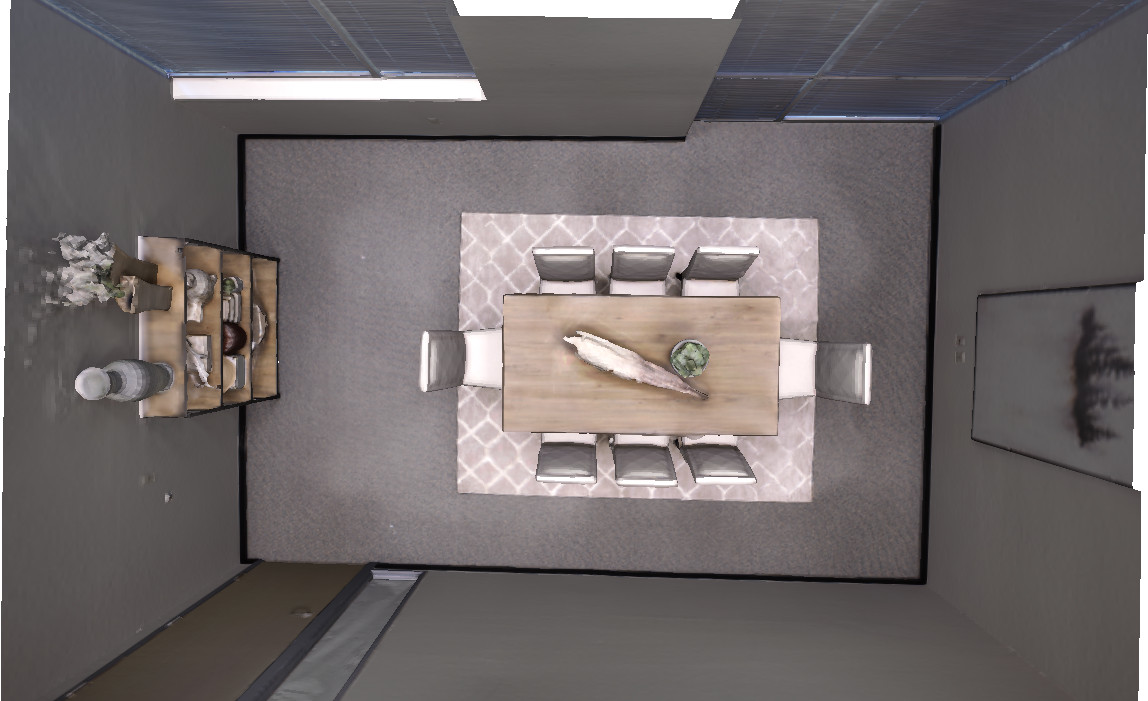} & 
\includegraphics[trim={0 0cm 0 0cm},clip=true, valign=c, width=\sz\linewidth]{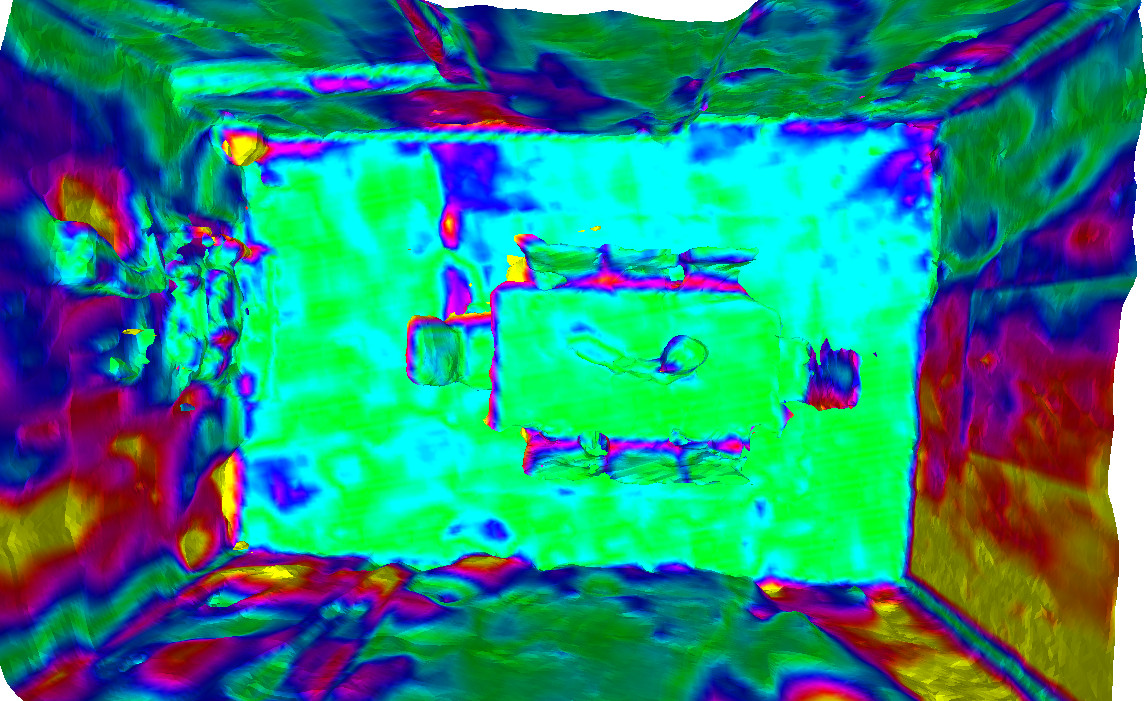} &
\includegraphics[trim={0 0cm 0 0cm},clip=true, valign=c, width=\sz\linewidth]{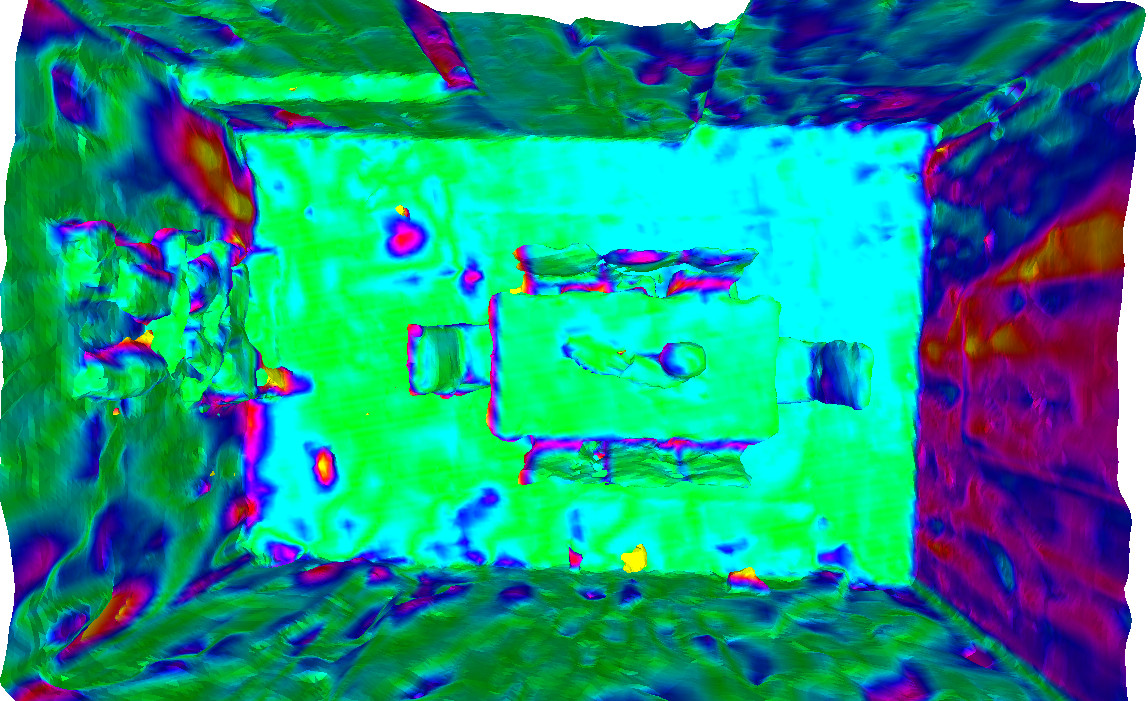} & 
\includegraphics[trim={0 0cm 0 0cm},clip=true, valign=c, width=\sz\linewidth]{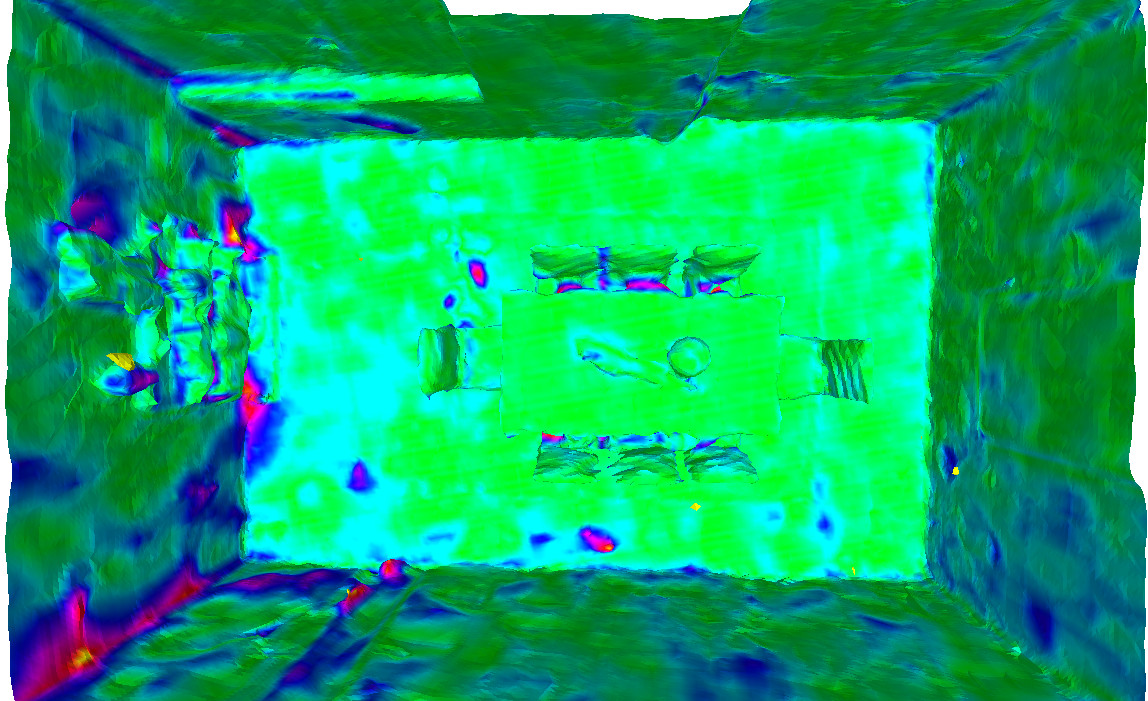} & \\
&  \includegraphics[trim={0 0cm 0 0cm},clip=true, valign=c, width=\sz\linewidth]{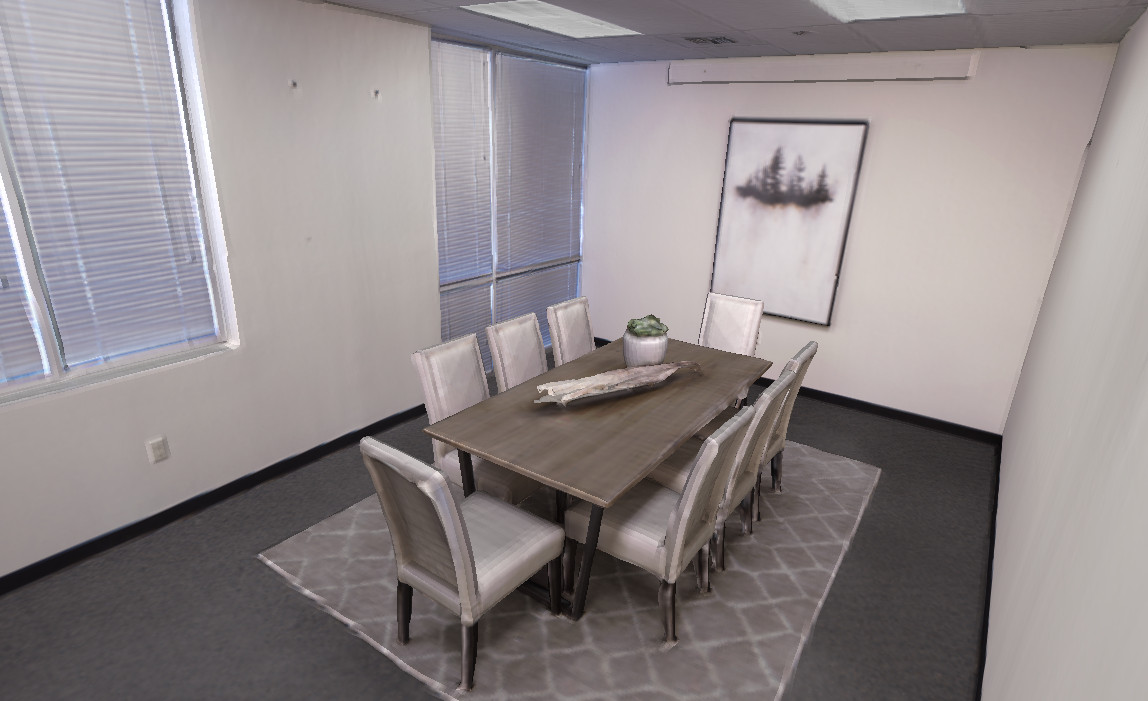} & 
\includegraphics[trim={0 0cm 0 0cm},clip=true, valign=c, width=\sz\linewidth]{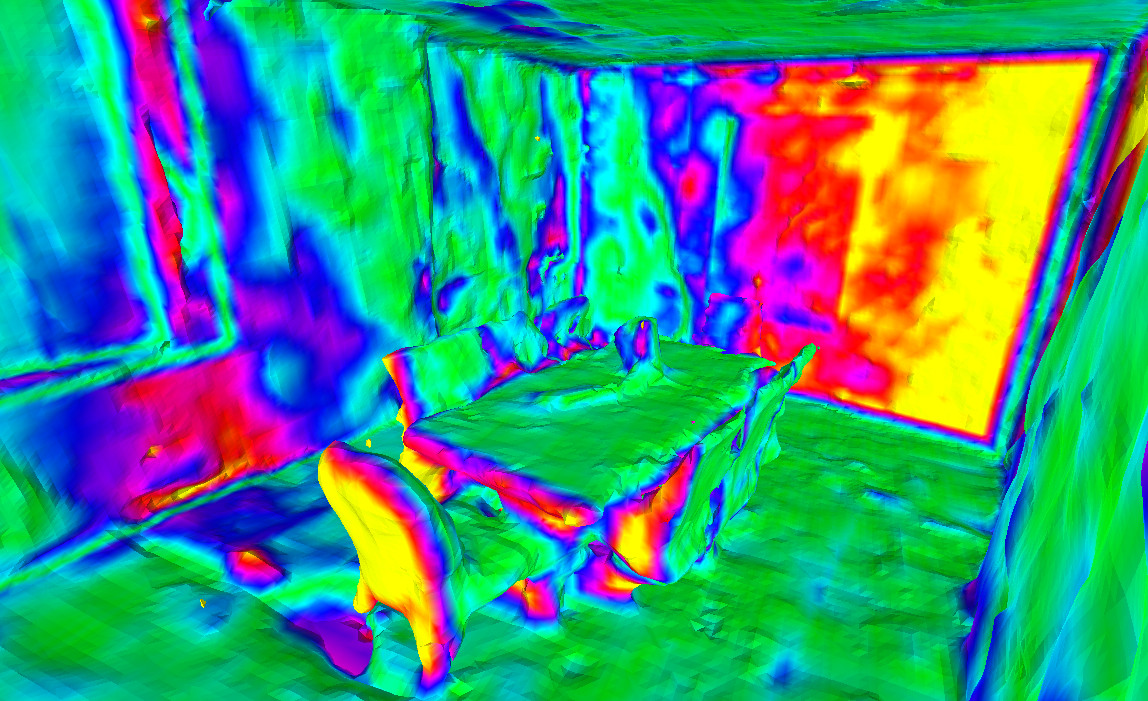} &
\includegraphics[trim={0 0cm 0 0cm},clip=true, valign=c, width=\sz\linewidth]{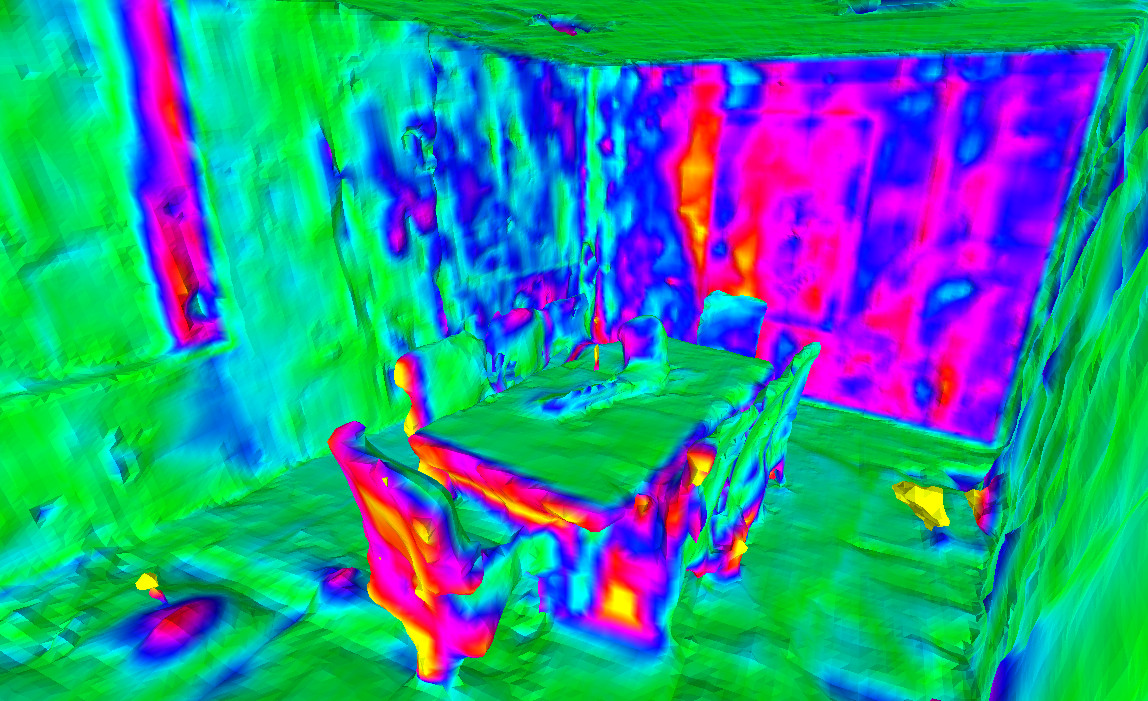} & 
\includegraphics[trim={0 0cm 0 0cm},clip=true, valign=c, width=\sz\linewidth]{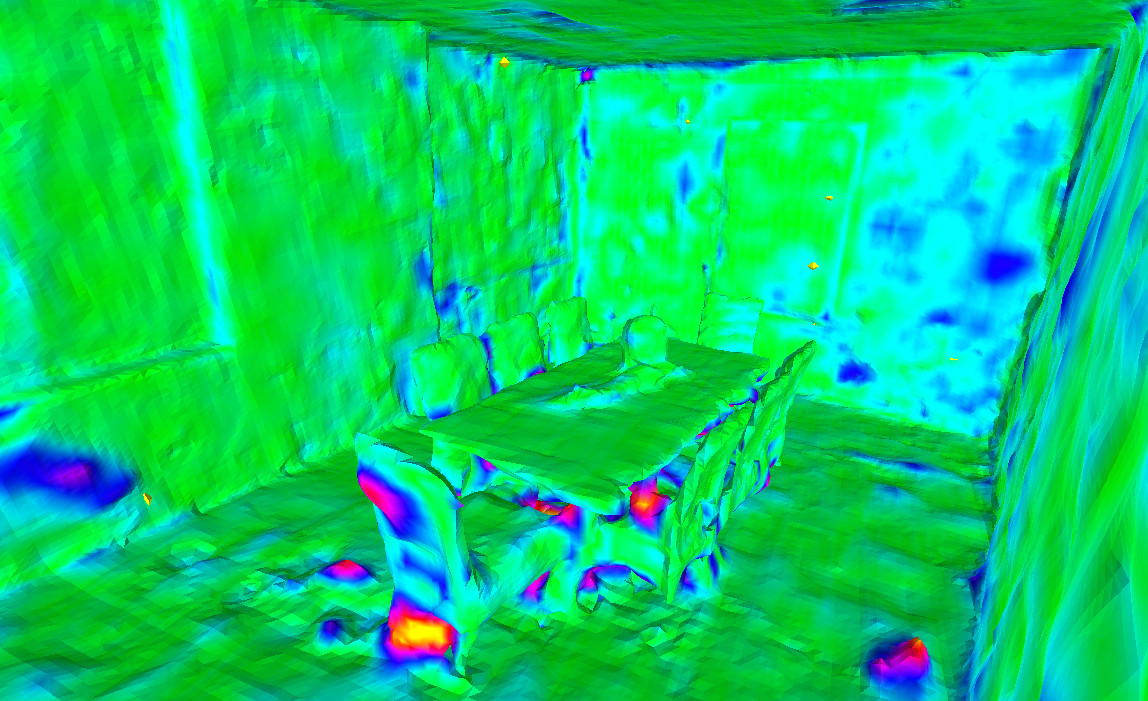} & \\
 & Ground Truth & PSMNet~\cite{chang2018pyramid} & SGM~\cite{hirschmuller2007stereo} & Fused (ours) & \\ 
\end{tabular}
}
\caption{\textbf{Multi-Sensor Reconstruction on Replica~\cite{straub2019replica}}. The two middle columns show single sensor reconstructions while the rightmost column shows the result when both sensors are jointly fused into the same geometry using our proposed \ours. Our uncertainty modeling helps on average to achieve more accurate reconstructions in the multi-sensor setting compared to the single sensor reconstructions. The colorbar displays the deviation from the ground truth mesh.}
\label{fig:multi_sensor_recon}
\end{figure}

\subsection{Memory and Runtime}
Due to the low number of parameters in our uncertainty MLP $h_w$ (5409), we add 43 kB to the already allocated 421 kB for the decoders in NICE-SLAM. This is negligible in comparison to the 95.86 MB allocated for the dense grids for the \texttt{office 0} scene. We report a 15 $\%$ increase in runtime over NICE-SLAM which can be compared to the average gain of 38$\%$ and 27$\%$ in terms of ATE RMSE on the 7-Scenes and TUM-RGBD datasets respectively and 11$\%$ and 32$\%$ in terms of the F1-score on single sensor RGBD SLAM and multi-sensor depth SLAM.

\subsection{Limitations} Our framework uses patch based modeling of uncertainty which may not hold in the general case along with the cheaply available features we feed as input to the uncertainty decoder. Simply using a more expressive model with learned features is not straight forward though, as shown by our results with the pretrained model and we leave this as future work. Finally, we believe that the relatively large voxel size we use can prevent efficient uncertainty learning from fine geometric details due to the high degree of averaging. We believe that our method can benefit from a scene representation that allows for resolving finer details.

%% file: tex/conclusion.tex
\section{Conclusion}
\label{sec:conclusion}
 The paper presents a way to learn per pixel depth uncertainties for dense neural SLAM. This allows the mapping and tracking re-rendering losses to be re-weighted such that trustworthy sensor readings are used to track the camera and to update the map. We believe this is a useful instrument in closing the gap in tracking accuracy to traditional sparse SLAM methods. We show that modeling depth uncertainty generally results in improvements both in terms of mapping and tracking accuracy and often performs better than alternatives that require ground truth depth or 3D. The paper also provides one of the initial solutions that utilizes more than one depth sensing modality for dense neural SLAM.

{\small
\boldparagraph{Acknowledgements.}
This work was supported by a VIVO collaboration project on real-time scene reconstruction, as well as by a research grants from FIFA. We thank Suryansh Kumar for fruitful discussions.
}

%% file: tex/supplementary_abstract.tex
\begin{abstract}
    This supplementary material accompanies the main paper by providing further information for better reproducibility as well as additional evaluations and qualitative results.
\end{abstract}

%% file: tex/supplementary.tex
\section{Video}
\label{sec:videos}

We provide a video that shows the predicted depth uncertainty along the trajectory of the \texttt{Office 1} scene from the Replica dataset. For reference, the video also contains the absolute depth error. Video link: \url{https://youtu.be/jsbZx3A7Y74}

\section{Method}
\label{sec:method}
In the following, we provide more details about our proposed method, specifically the decoder network architecture and details regarding the multi-sensor experiments.

\begin{figure*}[ht]
  \centering
    \includegraphics[width=1.0\linewidth]{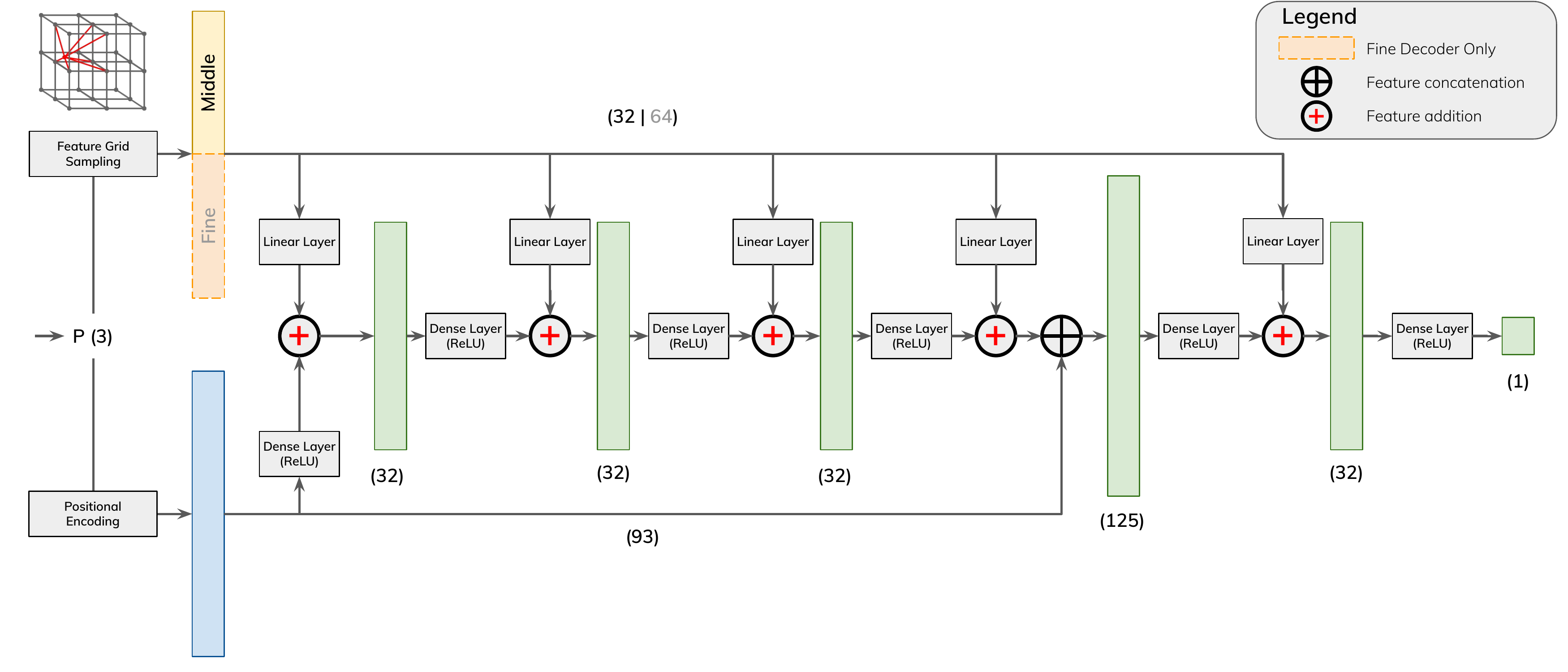}
  \caption{\textbf{Decoder Architecture.} The geometric middle and fine MLP architecture. The middle decoder  $f^1$ takes only the middle feature encoding as input, while the fine decoder $f^2$ takes the concatenation of the middle and fine geometric features as input.}
  \label{fig:og-mlp-arch}
\end{figure*}

\boldparagraph{Decoder Network Architecture.}
Each feature grid $\phi^l_\theta$ has an associated decoder $f^l$, where $l \in \{1,2\}$. Additionally, the color is encoded in a third feature grid $\psi_\omega$ with decoder $g_w$, used for further scene refinement after initial stages of geometric optimization. The observed scene geometry is reconstructed from the middle and fine resolution feature grids, with the fine feature grid output being added to the middle grid occupancy in a residual manner. We use the same geometric decoder architecture as proposed by NICE-SLAM \cite{zhu2022nice} detailed in \cref{fig:og-mlp-arch}. The color decoder $g_w$ follows the same general architecture as $f^1$, but outputs RGB instead of occupancy.

\boldparagraph{Multi-Sensor Modifications.}
We detail the main considerations that are required to handle two-sensor input.
When provided with more than one input sensor, the feature set for optimization extends to the furthest depth reading of the sensors. This ensures that the optimizable parameters include the set of feature points in the grid that would theoretically be observed by any sensor.
We modify the keyframe selection strategy to accept an \emph{averaged} depth map between the two input sensors for determining the relevant keyframes used during the mapping process. A future extension could include using the uncertainty maps to form a weighted depth map for this purpose.
NICE-SLAM samples 16 points near the depth measurements as well as 32 points a long the ray. With the addition of another sensor, we sample 16 points at around each sensor measurement on top of the 32 points along the ray for a total of 64 points. The 32 points sampled equally throughout the ray are determined by the maximum depth of both sensors. 

\section{Implementation Details}
\label{sec:imp_details}
We use PyTorch 1.11.0 and Python 3.7.11 to implement the pipeline. 
Training is done with the Adam optimizer using various Nvidia GPUs with a maximum 12 GB of memory.
For the uncertainty decoder $h_w$, we use a learning rate of 3e-4. For all optimizers, we use the default Adam hyperparameters \emph{betas} $= (0.9, 0.999)$, \emph{eps} = $1e$-$08$ and \emph{weight$\_$decay} = 0. The tracking learning rate for the camera pose is set to 0.001.

\section{Evaluation Metrics}
\label{sec:metrics}
\boldparagraph{Mapping.}
We use the following five metrics to quantify the reconstruction performance. 
We run marching cubes~\cite{lorensen1987marching} on the predicted occupancy grid $V$ and compare to the ground truth mesh. 
The F-score is defined as the harmonic mean between Recall (R) and Precision (P), $F = 2\frac{PR}{P+R}$. 
Precision is defined as the percentage of points on the predicted mesh which lie within some distance $\tau$ from a point on the ground truth mesh.
Vice versa, Recall is defined as the percentage of points on the ground truth mesh which lie within the same distance $\tau$ from a point on the predicted mesh. In all our experiments, we use a distance threshold $\tau = 0.05$ m. In addition to the F-score, Recall and Precision, we report the mean Precision and Recall which we define as the mean distance to all points. 
We use the evaluation script provided by the authors of \cite{Sandstrom2022LearningFusion}\footnote{\url{https://github.com/eriksandstroem/evaluate_3d_reconstruction_lib}}.

Finally, we report the depth L1 metric which renders depth maps from randomly sampled view points from the reconstructed and ground truth meshes. The depth maps are then compared and the L1 error is reported and averaged over 1000 sampled novel view points.

\boldparagraph{Tracking.} We use the absolute trajectory error (ATE) RMSE~\cite{Sturm2012ASystems} to compare tracking error across methods. This error normally computes the translation difference of the trajectories after alignment. We disable the alignment on Replica to better analyze camera pose drift, as the initial pose is fixed at the ground-truth pose. For the real-world experiments we keep the alignment enabled to be comparable to other methods.

\section{Baselines}
\label{sec:baselines}
\boldparagraph{SenFuNet.}
To make the comparison to our method fair, we increase the voxel size from $0.01$ m to $0.16$ m and train SenFuNet on the following scenes of the Replica dataset: $\{$\texttt{apartment 1, frl apartment 0, office 3, room 0, office 4, hotel 0}$\}$ and validate using the scene $\{$\texttt{frl apartment 1}$\}$.

\boldparagraph{Pretrained Confidence Network.}
We use the identical network as described by Weder~\etal~ \cite{Weder2020RoutedFusionLR} (called routing network), but make the following modification. We remove the refinement decoder of the network and only keep the confidence decoder of the network. This means that the network predicts the confidence of the input depth rather than the refined depth. We train on the following scenes of the Replica dataset: $\{$\texttt{apartment 1, frl apartment 0, office 3, room 0, office 4, hotel 0}$\}$ and validate using the scene $\{$\texttt{frl apartment 1}$\}$.

\section{More Experiments}
\label{sec:exp}

\subsection*{Single Sensor Evaluation}
\boldparagraph{Replica.} We provide experimental evaluations on the SL~\cite{handa2014benchmark} depth sensor in three different settings: 1. Depth only with ground truth poses \ie pure mapping from noisy data. 2. Depth with estimated camera poses (\ie with tracking) and 3. RGBD with tracking. In \cref{tab:sl} we find consistent improvements in the settings where tracking is enabled. In the mapping only setting, our model performs best in terms of precision for the SL sensor. When tracking is turned on, we perform better than NICE-SLAM in both the depth only setting and in the RGBD setting. \cref{fig:single_sensor_recon} shows a visualization from the \texttt{office 1} scene from the Replica dataset depicting lower surface reconstruction errors compared to NICE-SLAM~\cite{zhu2022nice}. For a visualization of the predicted uncertainty from the SL sensor compared to the pretrained model from NICE-SLAm+Pre, we refer to \cref{fig:unc_vis}.

\begin{table}[tb]
\centering
\setlength{\tabcolsep}{3pt}
\resizebox{\columnwidth}{!}
{
\begin{tabular}{l|lllllll}
\cellcolor{gray}       & \cellcolor{gray}Depth L1$\downarrow$      & \cellcolor{gray}mP$\downarrow$  & \cellcolor{gray}mR$\downarrow$     & \cellcolor{gray}P$\uparrow$ & \cellcolor{gray}R$\uparrow$ & \cellcolor{gray}F$\uparrow$ & \cellcolor{gray}ATE$\downarrow$\\
\multirow{-2}{*}{\cellcolor{gray} ${\text{Model}\downarrow | \text{Metric}\rightarrow}$} & \cellcolor{gray}[cm] & \cellcolor{gray}[cm] & \cellcolor{gray}[cm] & \cellcolor{gray}$[\%]$ & \cellcolor{gray}$[\%]$ & \cellcolor{gray}$[\%]$ & \cellcolor{gray}[cm]\\\hline
\multicolumn{8}{c}{\emph{Depth Only + Ground Truth Poses}} \\ \hline
NICE-SLAM~\cite{zhu2022nice} & \fs \textbf{1.79} & \nd 2.23 & \fs \textbf{1.69} & \nd 91.06 & \fs \textbf{93.97} & \fs \textbf{92.47} & \multicolumn{1}{c}{-} \\
NICE-SLAM+Pre & \rd 1.88 & \rd 2.25 & \rd 1.77 & \rd 90.68 & \rd 93.35 & \rd 91.98 & \multicolumn{1}{c}{-} \\
Ours & \nd 1.85  & \fs \textbf{2.19} & \nd 1.75  & \fs \textbf{91.12} & \nd 93.50 & \nd 92.28 & \multicolumn{1}{c}{-} \\ \hline 
\multicolumn{8}{c}{\emph{Depth + Tracking}} \\ \hline
NICE-SLAM~\cite{zhu2022nice} & \rd 16.47 & \rd 13.36 & \rd 10.31 & \rd 42.56 & \rd 46.34 & \rd 44.29 & \rd 41.51 \\
NICE-SLAM+Pre & \fs \textbf{8.62} & \fs \textbf{7.98} & \fs \textbf{7.13} & \fs \textbf{53.89}  & \fs \textbf{56.88}  & \fs \textbf{55.32}  & \fs \textbf{26.11}  \\
Ours & \nd 11.48   & \nd 9.40 & \nd 8.89 & \nd 49.63  & \nd 52.08  & \nd 50.80  & \nd 29.99  \\ \hline 
\multicolumn{8}{c}{\emph{RGB-D + Tracking}} \\ \hline
NICE-SLAM~\cite{zhu2022nice} & \nd 14.39 & \nd 11.80 & \nd 9.51 & \nd 44.31 & \nd 47.81 & \nd 45.96 & \nd 36.03 \\
Ours & \fs \textbf{9.78}  & \fs \textbf{8.74} & \fs \textbf{8.46} & \fs \textbf{49.78}  & \fs \textbf{51.78}  & \fs \textbf{50.73}  & \fs \textbf{25.64}  \\ 
\hline
\end{tabular}
}
\caption{\textbf{Reconstruction Performance on Replica~\cite{straub2019replica}: SL~\cite{handa2014benchmark}.} Our model is able to outperform NICE-SLAM when tracking is enabled. In the mapping only setting, our model favors precision over recall compared to NICE-SLAM~\cite{zhu2022nice}. Best results are highlighted as \colorbox{colorFst}{\bf first}, \colorbox{colorSnd}{second}, and \colorbox{colorTrd}{third}.}
\label{tab:sl}
\end{table}

\begin{figure}[t]
\centering
{\footnotesize
\setlength{\tabcolsep}{1pt}
\renewcommand{\arraystretch}{1}
\newcommand{\sz}{0.29}
\begin{tabular}{ccccc}
\rotatebox[origin=c]{90}{\texttt{Office 1}} & 
\includegraphics[trim={0 1cm 0 1cm},clip=true, valign=c, width=\sz\linewidth]{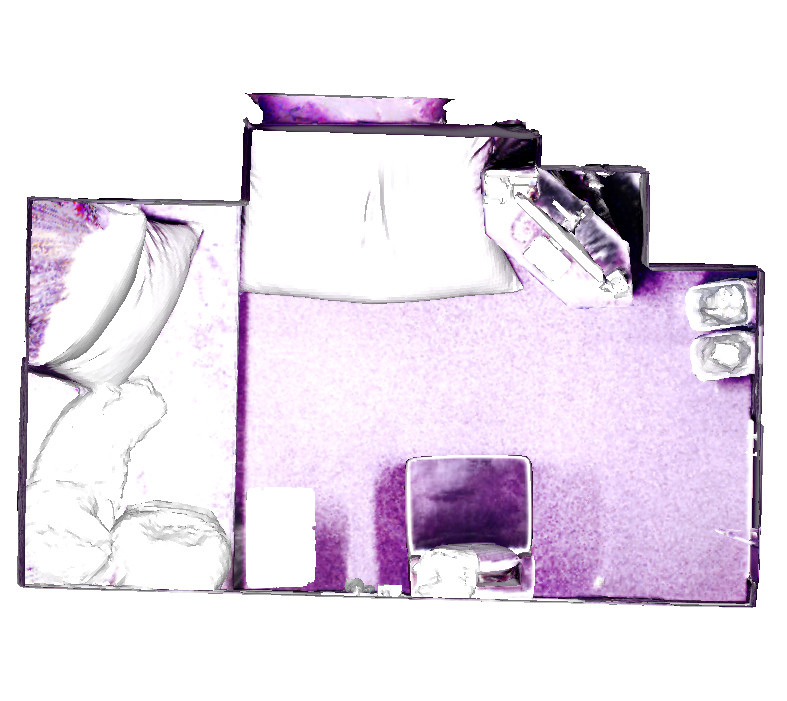} & 
\includegraphics[trim={0 1cm 0 1cm},clip=true, valign=c, width=\sz\linewidth]{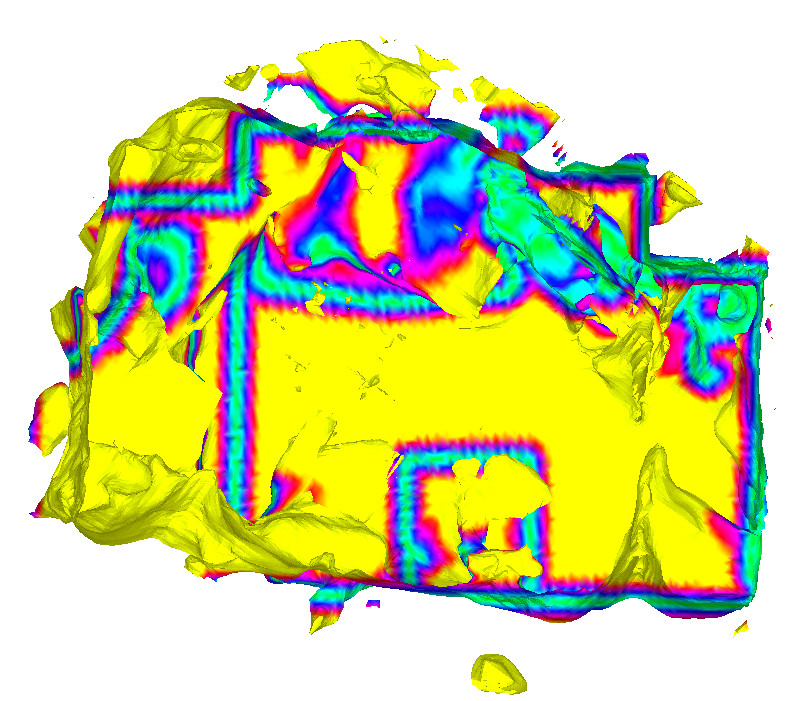} &
\includegraphics[trim={0 1cm 0 1cm},clip=true, valign=c, width=\sz\linewidth]{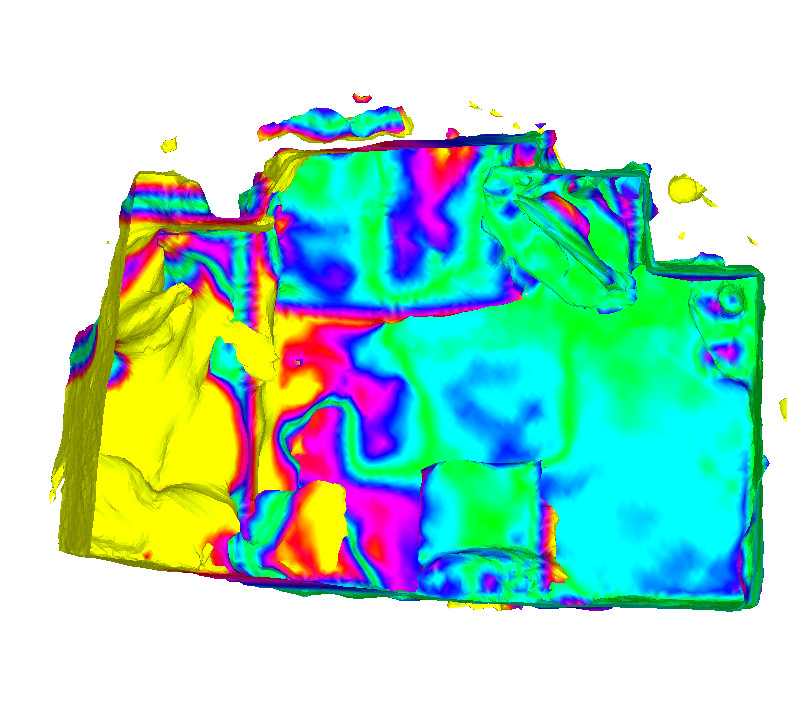} &  \multirow{1}{*}[20.0pt]{\includegraphics[width=.07\linewidth]{figs/colorbar_outlier_filter.jpg}}\\
 & Ground Truth & NICE-SLAM~\cite{zhu2022nice} & Ours & 
\end{tabular}
}
\caption{\textbf{Single Sensor Reconstruction on Replica~\cite{straub2019replica}.} We show that our uncertainty modeling on average helps to achieve more accurate reconstructions when the SL~\cite{handa2014benchmark} sensor is provided as input. This experiment uses RGBD input with tracking enabled. The colorbar displays the deviation from the ground truth mesh.}
\label{fig:single_sensor_recon}
\end{figure}

\begin{figure}[t]
\centering
{\tiny
\setlength{\tabcolsep}{1pt}
\renewcommand{\arraystretch}{1}
\newcommand{\sz}{0.95}
\begin{tabular}{cc}
 &  \hspace{2.8cm} Estimated Uncertainty \hspace{1.65cm} Ground Truth  \\
 & \hspace{-0.2cm} Sensor Depth \hspace{1.1cm} NICE-SLAM+Pre \hspace{0.75cm} Ours \hspace{1.65cm} Depth Error\\[0.0cm]
\raisebox{-0.75cm}{\rotatebox[origin=c]{90}{\footnotesize SL~\cite{handa2014benchmark}\tiny}} & \multirow{1}{*}{\includegraphics[width=\sz\linewidth]{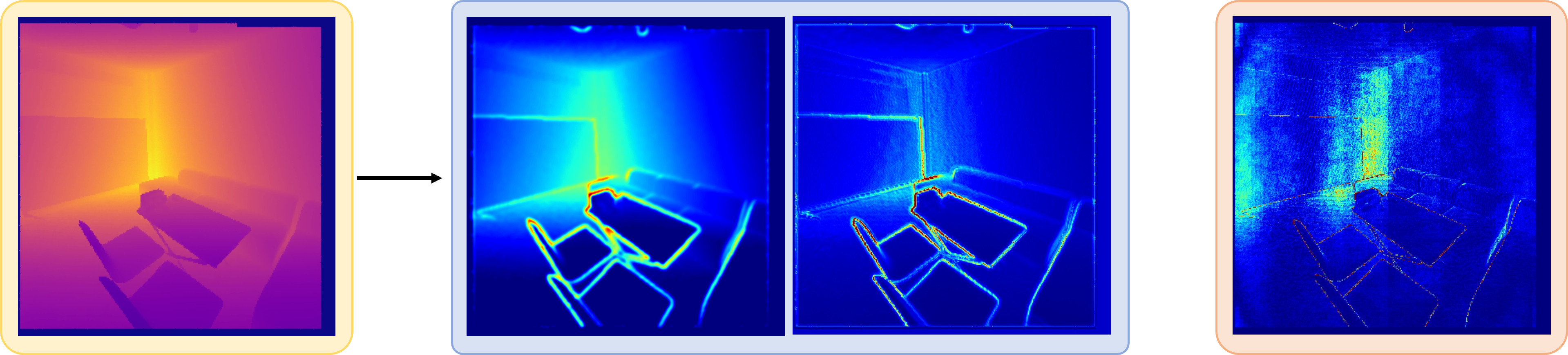}} \\
\end{tabular}
}
\vspace{0.5cm}
\caption{\textbf{Uncertainty Visualization.} From left to right, the depth map from the SL sensor with the associated uncertainty estimation from the pretrained network model and ours. As reference, the ground truth absolute depth error is shown in the last column. We find that our model reproduces the error map with less smoothing than the pretrained model. Blue: low uncertainty, red: high uncertainty.}
\label{fig:unc_vis}
\end{figure}

\boldparagraph{TUM-RGBD.}
In \cref{tab:track-TUM}, we provide additional evaluation metrics on the real-world TUM-RGBD dataset~\cite{Sturm2012ASystems} over the main paper. Specifically, we provide median and minimum ATE RMSE over 10 runs. On average, camera pose tracking is greatly benefited by our uncertainty aware strategy. Due to the randomness of the pipeline, we find that the best performance (minimum ATE) is similar to NICE-SLAM.

\begin{table}[ht]
\vspace{-0.2cm}
\centering
\footnotesize
\setlength{\tabcolsep}{2pt}
\renewcommand{\arraystretch}{1.1}
\resizebox{\columnwidth}{!}
{
\begin{tabular}{llccc}
\toprule
Scene & Model & Mean ATE [\si{cm}]$\downarrow$ & Median ATE [\si{cm}]$\downarrow$  & Min. ATE [\si{cm}]$\downarrow$ \\
\midrule
\multirow{2}{*}{Fr1 Desk}       & NICE-SLAM & \nd \textrm{40.40} & \nd 31.74 & \nd 6.27  \\
       & Ours & \fs 29.04 & \fs 6.50 & \fs 5.54   \\
\midrule
\multirow{2}{*}{Fr1 Desk2}        & NICE-SLAM & \nd 47.81 & \nd 48.07 & \fs 16.73    \\
        & Ours & \fs 36.57 & \fs 27.00 & \nd 18.42   \\
\midrule
\multirow{2}{*}{Fr1 xyz}        & NICE-SLAM & \nd 5.11 & \nd 2.78 & \nd 2.65   \\
       & Ours & \fs 2.71 & \fs 2.71 & \fs 2.61  \\
\midrule
\multirow{2}{*}{Average}        & NICE-SLAM & \nd 31.11 & \nd 27.53 & \fs 8.55   \\
       & Ours & \fs 22.77 & \fs 12.07 & \nd 8.86 \\
\bottomrule

\end{tabular}}
\caption{\textbf{Tracking Evaluation on TUM-RGBD.} We report the average and median ATE RMSE and the minimum ATE RMSE over 10 runs for each scene by mapping every 2nd frame. Best results are highlighted as \colorbox{colorFst}{\bf first} and \colorbox{colorSnd}{second}.
}
\label{tab:track-TUM}
\end{table}

\boldparagraph{7-Scenes.} We provide additional evaluation metrics over the main paper per scene in \cref{tab:track-7} on the 7-Scenes dataset~\cite{glocker2013real}. We use sequence 1 for all scenes. We find that NICE-SLAM~\cite{zhu2022nice} consistently yields worse tracking results suggesting the effectiveness of our depth uncertainty when it comes to maintaining robust camera pose tracking. On average, our method yields a 38 $\%$ gain in terms of the mean ATE. Due to the randomness of the pipeline, we find that the best performance (minimum ATE) is similar to NICE-SLAM.

\begin{table}[ht]
\centering
\footnotesize
\setlength{\tabcolsep}{2pt}
\renewcommand{\arraystretch}{1.1}
\resizebox{\columnwidth}{!}
{
\begin{tabular}{llccc}
\toprule
Scene & Model & Mean ATE [\si{cm}]$\downarrow$ & Median ATE [\si{cm}]$\downarrow$  & Min. ATE [\si{cm}]$\downarrow$ \\
\midrule
\multirow{2}{*}{Chess}       & NICE-SLAM~\cite{zhu2022nice} & \nd \textrm{40.30} & \nd \textrm{15.77} & \fs \textbf{\textrm{10.38}}  \\
       & Ours & \fs \textbf{\textrm{14.85}} & \fs \textbf{\textrm{12.49}} & \nd \textrm{10.88}   \\
\midrule
\multirow{2}{*}{Fire}        & NICE-SLAM~\cite{zhu2022nice} & \nd \textrm{47.67} & \nd \textrm{41.24} & \nd \textrm{7.89}  \\
        & Ours & \fs \textbf{\textrm{25.47}} & \fs \textbf{\textrm{11.88} }& \fs \textbf{\textrm{7.00} } \\
\midrule
\multirow{2}{*}{Head}        & NICE-SLAM~\cite{zhu2022nice} & \nd \textrm{20.55} & \fs \textbf{12.95} & \nd \textrm{8.46}  \\
       & Ours & \fs \textbf{\textrm{13.12}} & \nd \textrm{\textrm{14.53} }& \fs \textbf{\textrm{8.31} }  \\
\midrule
\multirow{2}{*}{Office}      & NICE-SLAM~\cite{zhu2022nice} & \nd \textrm{8.49}  & \nd \textrm{8.40} & \nd \textrm{6.99}   \\
     & Ours & \fs \textbf{\textrm{7.83} } & \fs \textbf{\textrm{8.05} }& \fs \textbf{\textrm{6.57} } \\
\midrule
\multirow{2}{*}{Pumpkin}     & NICE-SLAM~\cite{zhu2022nice} & \nd \textrm{33.11} & \fs \textbf{27.63} & \fs \textbf{\textrm{25.92}}  \\
    & Ours & \fs \textbf{\textrm{29.32}} & \nd \textrm{\textrm{28.47} }& \nd \textrm{27.59}  \\
\midrule
\multirow{2}{*}{Red Kitchen} & NICE-SLAM~\cite{zhu2022nice} & \nd \textrm{24.39}  & \nd \textrm{7.56} & \nd \textrm{6.61}  \\
 & Ours & \fs \textbf{\textrm{6.21} } & \fs \textbf{\textrm{6.07} }& \fs \textbf{\textrm{5.42} } \\
\midrule
\multirow{2}{*}{Stairs}      & NICE-SLAM~\cite{zhu2022nice} & \nd \textrm{9.18}  & \nd \textrm{\textrm{8.81}} & \nd \textrm{7.80}   \\
     & Ours & \fs \textbf{\textrm{8.53} } & \fs \textbf{8.05} & \fs \textbf{\textrm{6.24} }  \\
\midrule
\multirow{2}{*}{Average}      & NICE-SLAM~\cite{zhu2022nice} & \nd \textrm{24.24}  & \nd \textrm{17.48} & \nd \textrm{10.58} \\
     & Ours & \fs \textbf{\textrm{15.05}}  & \fs \textbf{\textrm{12.79}} & \fs \textbf{\textrm{10.29}}  \\  
\bottomrule
\end{tabular}}
\caption{\textbf{Tracking Evaluation on 7-Scenes.} We report the average ATE RMSE, the median ATE RMSE as well as the minimum ATE RMSE over 5 runs for each scene. With our depth uncertainty modeling, we achieve significantly better tracking compared to NICE-SLAM. On average, our method yields a 38 $\%$ gain in terms of the mean ATE.}
\label{tab:track-7}
\end{table}

\subsection*{Multi-Sensor Evaluation}

\begin{table}[tb]
\centering
\setlength{\tabcolsep}{3pt}
\resizebox{\columnwidth}{!}
{
\begin{tabular}{l|lllllll}
\cellcolor{gray}       & \cellcolor{gray}Depth L1$\downarrow$      & \cellcolor{gray}mP$\downarrow$  & \cellcolor{gray}mR$\downarrow$     & \cellcolor{gray}P$\uparrow$ & \cellcolor{gray}R$\uparrow$ & \cellcolor{gray}F$\uparrow$ & \cellcolor{gray}ATE$\downarrow$\\
\multirow{-2}{*}{\cellcolor{gray} ${\text{Model}\downarrow | \text{Metric}\rightarrow}$} & \cellcolor{gray}[cm] & \cellcolor{gray}[cm] & \cellcolor{gray}[cm] & \cellcolor{gray}$[\%]$ & \cellcolor{gray}$[\%]$ & \cellcolor{gray}$[\%]$ & \cellcolor{gray}[cm]\\\hline
\multicolumn{8}{c}{\emph{Single Sensor Ours: Depth + Ground Truth Poses}} \\ \hline
PSMNet~\cite{chang2018pyramid} & 2.42  & 2.58 & 2.29 & 89.14 & 88.70 & 88.89 & -\\ 
SL~\cite{handa2014benchmark}   & 1.85  & 2.19   & \fs \textbf{1.75}  & 91.12 & \nd 93.50 & \nd 92.28 & -\\ 
\hline
\multicolumn{8}{c}{\emph{Multi-Sensor: Depth + Ground Truth Poses}} \\ \hline
NICE-SLAM~\cite{zhu2022nice} & \rd 1.76 & \rd 2.16 & \nd 1.78 & \rd 91.73 & 92.69 & \rd 92.19 & -\\
SenFuNet~\cite{Sandstrom2022LearningFusion} & 14.43 & 9.65   & 13.26 & 35.70 & 27.32 & 30.85 & -\\ 
Vox-Fusion~\cite{yang2022vox} & 3.32 & 31.10 & 1.97 & 60.18 & \fs \textbf{94.10} & 72.57& - \\ 
NICE-SLAM+Pre & \nd 1.62 & \nd 2.09 & \nd 1.78 & \nd 91.78 & 92.49 & 92.11 & -\\
Ours & \fs \textbf{1.60}& \fs \textbf{2.02}& \fs \textbf{1.75}& \fs \textbf{92.20} & \rd 92.77 & \fs \textbf{92.46} & -  \\ 
\hline
\multicolumn{8}{c}{\emph{Single Sensor Ours: Depth + Tracking}} \\ \hline
PSMNet~\cite{chang2018pyramid} & 7.39 & 6.56& 6.20& 57.30 & 57.57 & 57.41 & \nd 19.36 \\ 
SL~\cite{handa2014benchmark} &  11.48   &  9.40 & 8.89 &  49.63  &  52.08  &  50.80  &  29.99  \\ \hline %
\multicolumn{8}{c}{\emph{Multi-Sensor: Depth + Tracking}} \\ \hline
NICE-SLAM~\cite{zhu2022nice} & \rd 4.70 & \rd 6.39 & \rd 5.74 & \rd 63.61 & \rd 64.58 & \rd 64.08 &  22.16 \\
NICE-SLAM+Pre & \fs 3.57  & \fs 4.63 & \fs 4.30 & \fs 70.53 & \fs 70.90 & \fs 70.70 & \rd 20.43 \\
Ours & \nd 3.87 & \nd 5.48 & \nd 4.90 & \nd 66.20 & \nd 66.59 & \nd 66.36 & \fs 16.86 \\
\hline
\end{tabular}
}
\caption{\textbf{Reconstruction Performance on Replica~\cite{straub2019replica}: SL~\cite{handa2014benchmark}\bplus{}PSMNet~\cite{chang2018pyramid}.} Our multi-sensor reconstruction performance improves over the single sensor results in isolation and we outperform the baseline methods. The experiment was conducted in the depth only setting with known camera poses.}
\label{tab:sl_psmnet}
\end{table}

\begin{figure}[t]
\centering
{\footnotesize
\setlength{\tabcolsep}{1pt}
\renewcommand{\arraystretch}{1}
\newcommand{\sz}{0.21}
\begin{tabular}{cccccc}
\rotatebox[origin=c]{90}{\texttt{Office 0}} & 
\includegraphics[trim={0 0cm 0 0cm},clip=true, valign=c, width=\sz\linewidth]{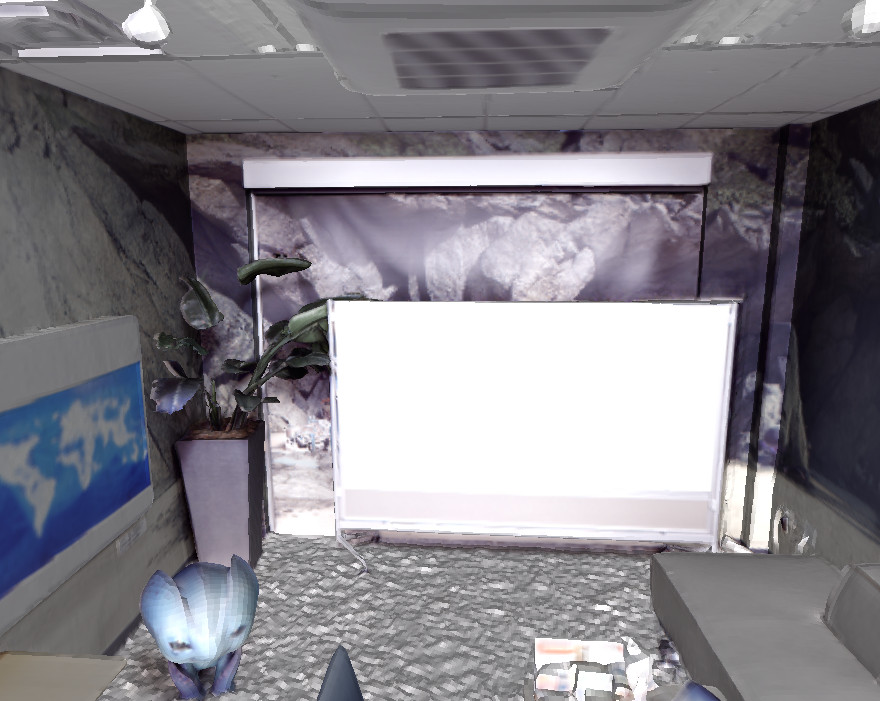} & 
\includegraphics[trim={0 0cm 0 0cm},clip=true, valign=c, width=\sz\linewidth]{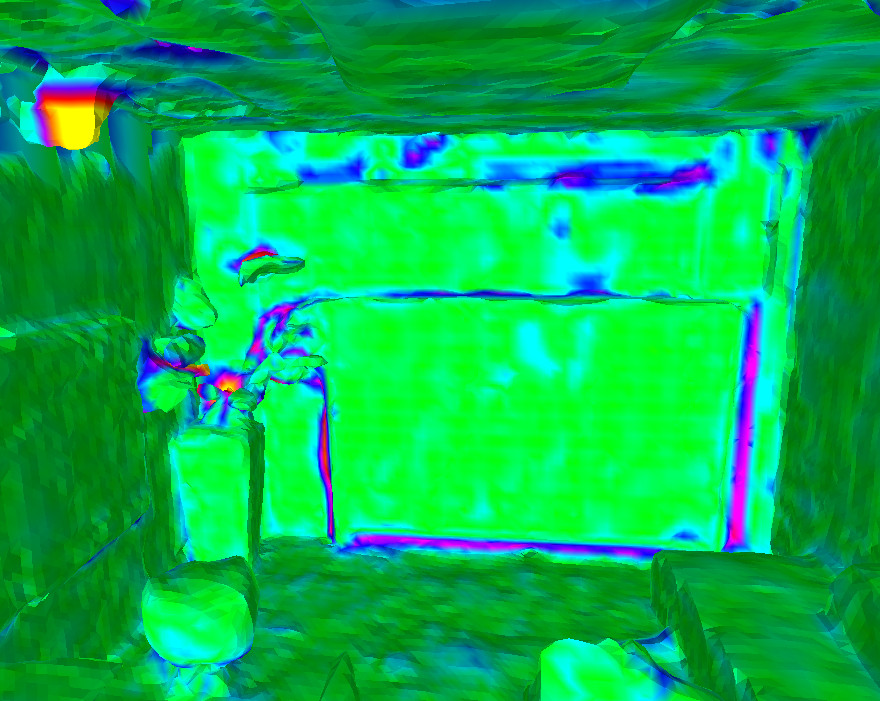} &
\includegraphics[trim={0 0cm 0 0cm},clip=true, valign=c, width=\sz\linewidth]{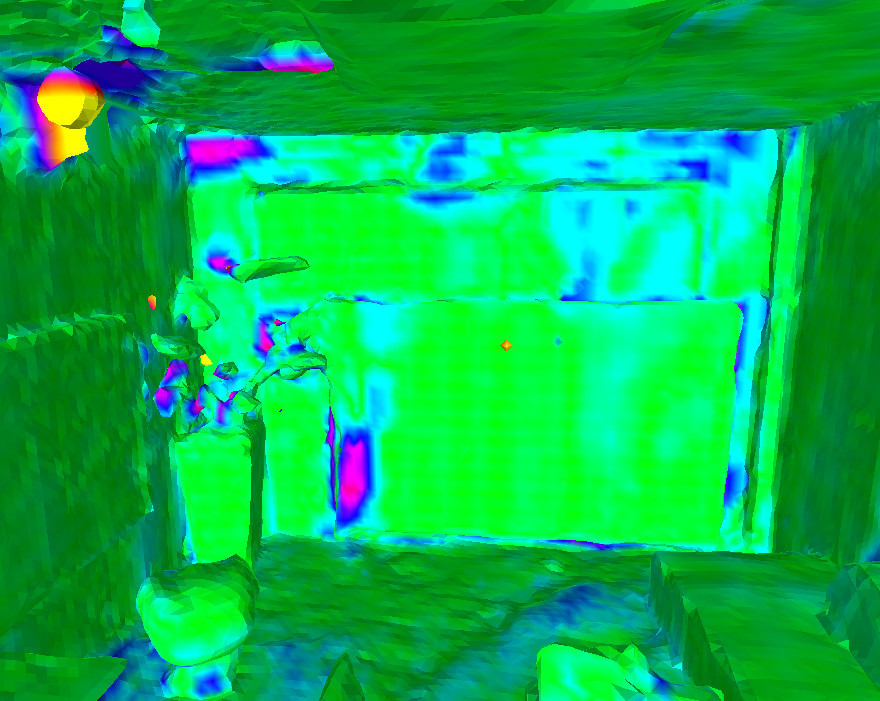} & 
\includegraphics[trim={0 0cm 0 0cm},clip=true, valign=c, width=\sz\linewidth]{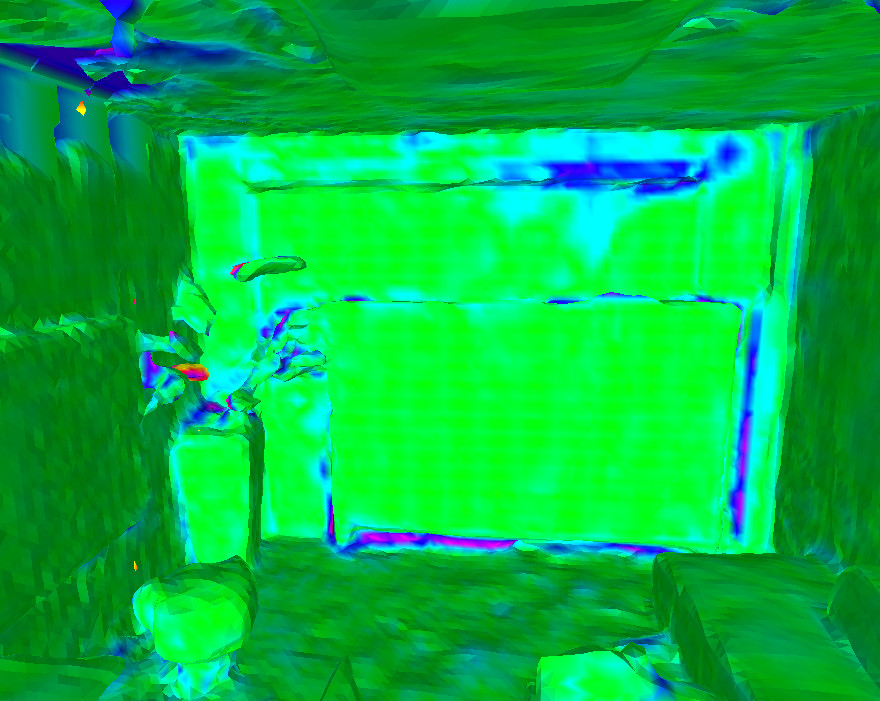} &
\multirow{1}{*}[-8.5pt]{\includegraphics[width=.1\linewidth]{figs/colorbar_outlier_filter.jpg}} \\
 & Ground Truth & PSMNet~\cite{chang2018pyramid} & SL~\cite{handa2014benchmark} & Fused (ours) & \\ 
\rotatebox[origin=c]{90}{\texttt{Room 2}} & 
\includegraphics[trim={0 0cm 0 0cm},clip=true, valign=c, width=\sz\linewidth]{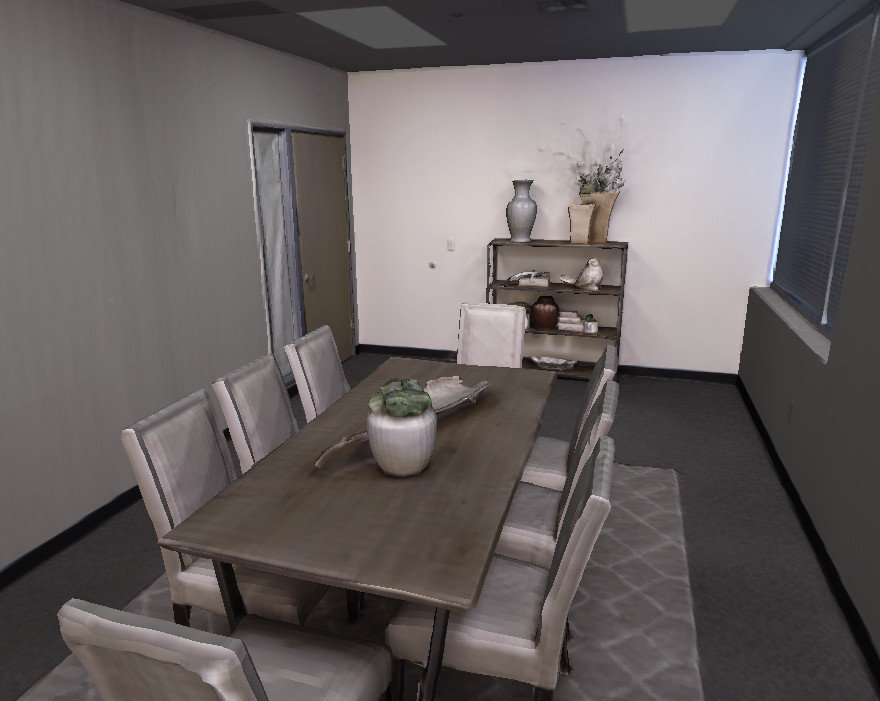} & 
\includegraphics[trim={0 0cm 0 0cm},clip=true, valign=c, width=\sz\linewidth]{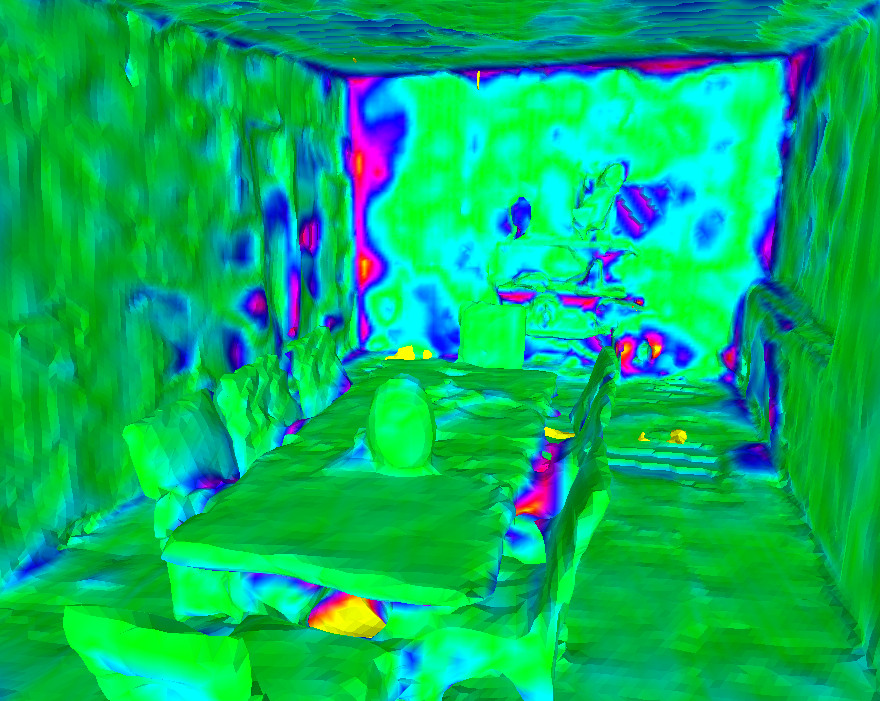} &
\includegraphics[trim={0 0cm 0 0cm},clip=true, valign=c, width=\sz\linewidth]{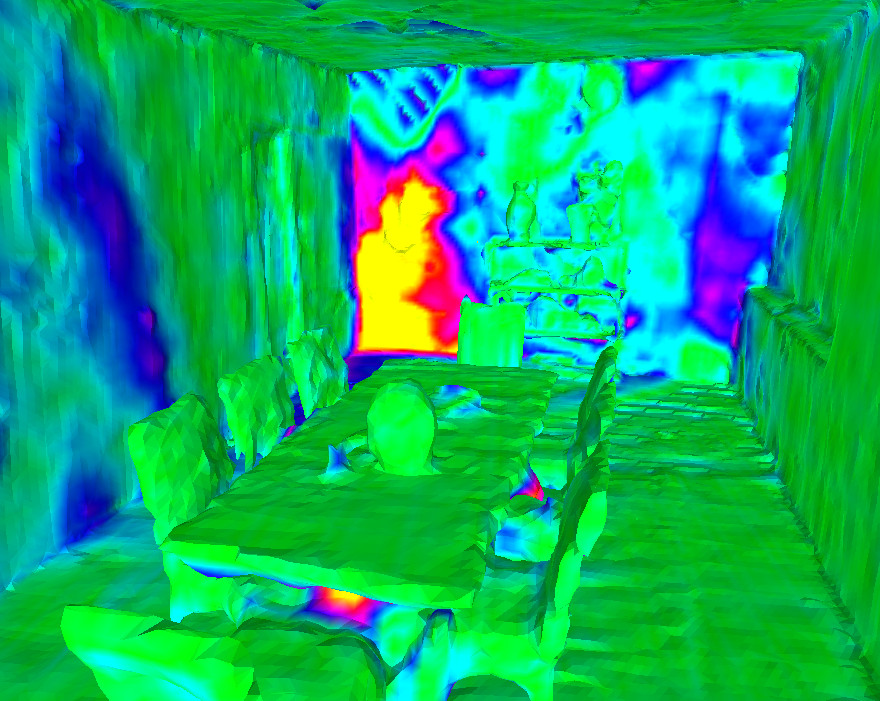} &
\includegraphics[trim={0 0cm 0 0cm},clip=true, valign=c, width=\sz\linewidth]{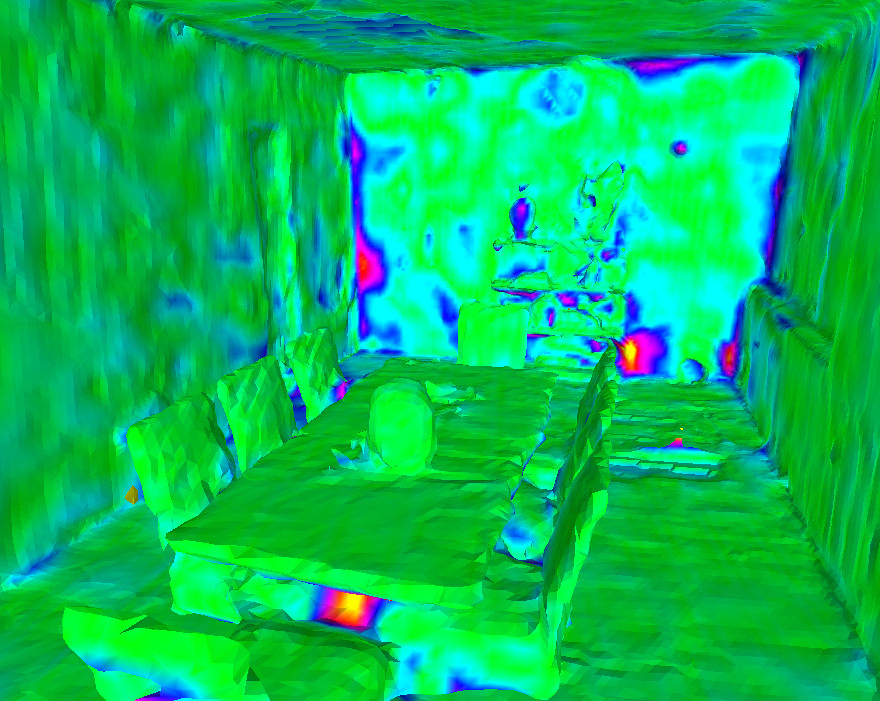} & \\
 & Ground Truth & PSMNet~\cite{chang2018pyramid} & SL~\cite{handa2014benchmark} & Fused (ours) & \\  
\rotatebox[origin=c]{90}{\texttt{Room 2}} & 
\includegraphics[trim={0 0cm 0 0cm},clip=true, valign=c, width=\sz\linewidth]{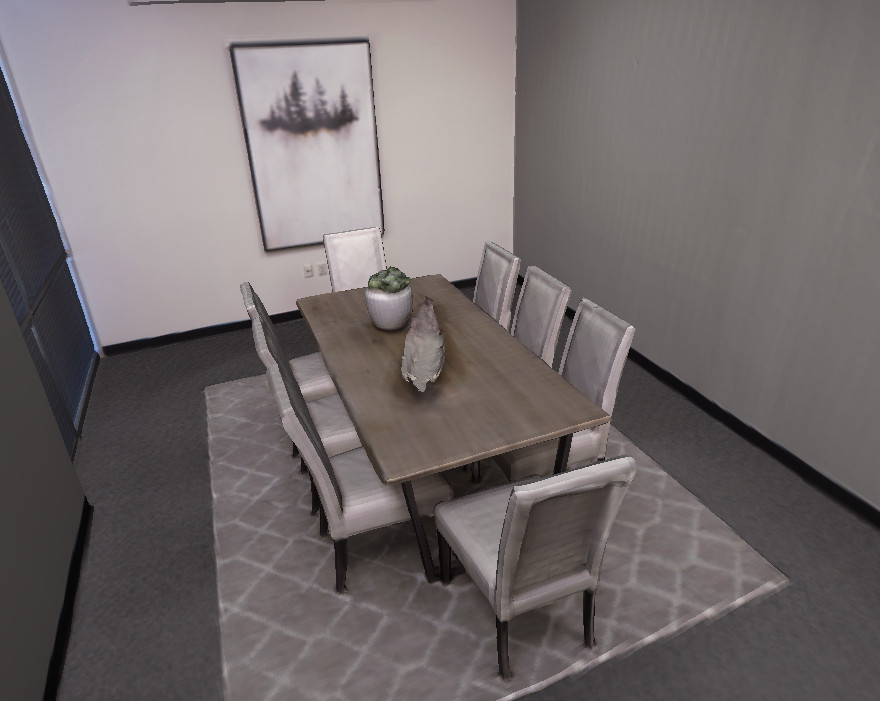} & 
\includegraphics[trim={0 0cm 0 0cm},clip=true, valign=c, width=\sz\linewidth]{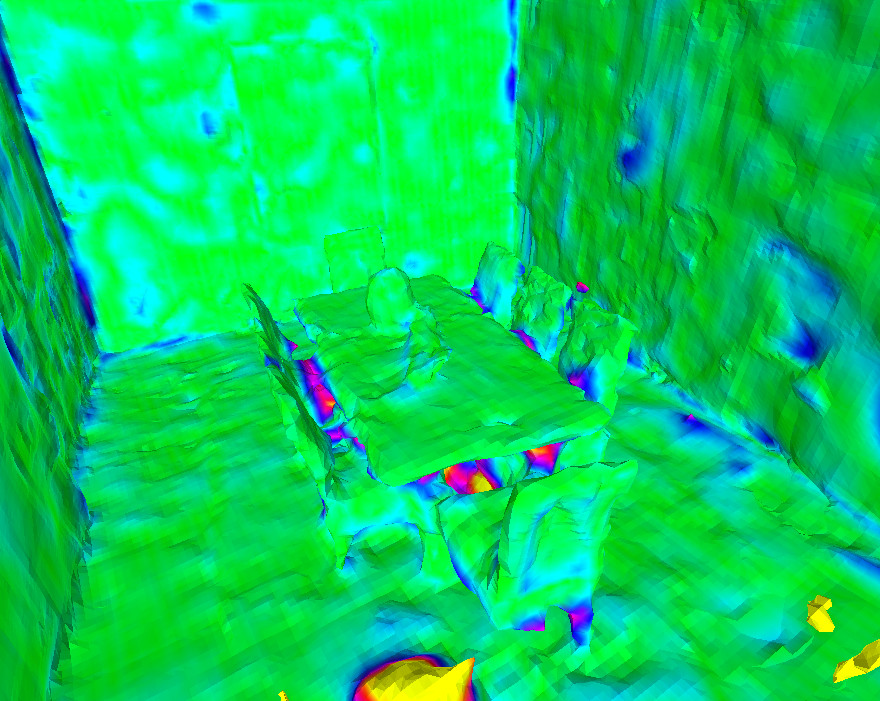} &
\includegraphics[trim={0 0cm 0 0cm},clip=true, valign=c, width=\sz\linewidth]{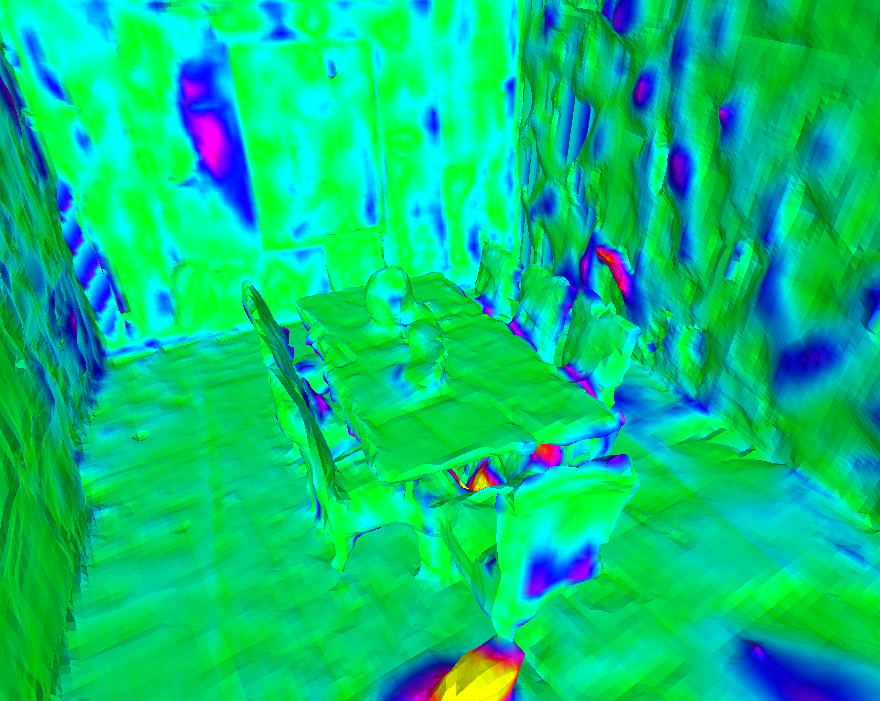} & 
\includegraphics[trim={0 0cm 0 0cm},clip=true, valign=c, width=\sz\linewidth]{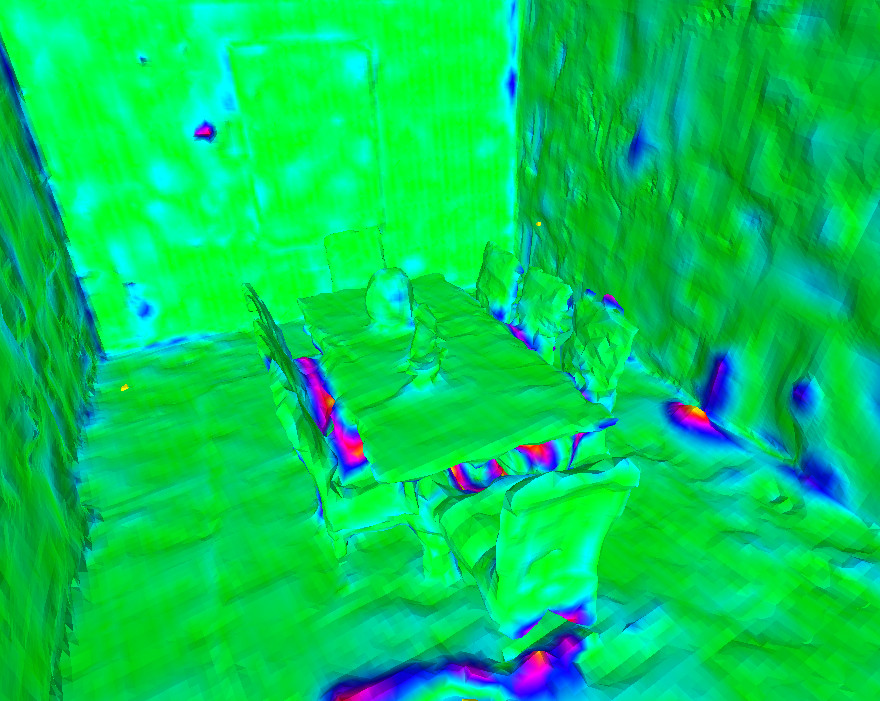} & \\
 & Ground Truth & PSMNet~\cite{chang2018pyramid} & SGM~\cite{hirschmuller2007stereo} & Fused (ours) &  
\end{tabular}
}
\caption{\textbf{Multi-Sensor Reconstruction on Replica~\cite{straub2019replica} with Ground Truth Poses}. The two middle columns show single sensor reconstructions while the rightmost column shows the result when both sensors are jointly fused into the same geometry using our proposed \ours. Our uncertainty modeling helps on average to achieve more accurate reconstructions in the multi-sensor setting compared to the single sensor reconstructions. The colorbar displays the deviation from the ground truth mesh.}
\label{fig:multi_sensor_recon}
\end{figure}

\cref{fig:multi_sensor_recon} shows multi-sensor reconstruction results with ground truth poses. Also in this case, a visual improvement can on average be observed over the single sensor reconstructions.

\subsection*{Architecture Ablation}

We provide architecture ablations on the Replica dataset under the SL noise model and ground truth poses. For the ablations, we use the same evaluation protocol and metrics are reported by NICE-SLAM \cite{zhu2022nice} \ie the mean Precision (mP), mean Recall (mR),  Recall (R) and depth L1. 

We select four variables to vary in these experiments to understand which architecture provides the most promising results. We vary the minimum uncertainty value $\beta_{\mathrm{min}}$: 1e-1~\si{m} or 1e-3~\si{m}. We vary the kernel size or patch size: 1$\times$1 or 5$\times$5. Lastly we select two options for the informative features. The first option is to use the depth and the incident angle. This yields a pixel feature dimension of two. The second option is to use the depth, the normal direction, the image gradients, and the incident angle yielding a pixel feature dimension of seven. In total, we perform 8 ablations, whose details are provided in \cref{tab:archabl}\footnote{We provide trial names based on the ablation parameters. These involve the patch size, the number of features, and the use of a ``small'' or ``large'' regularizer: [1K/5K][2F/7F][S/L]}. 

\begin{table}[ht]
\centering
\footnotesize
\setlength{\tabcolsep}{7pt}
\renewcommand{\arraystretch}{1.0}
\begin{tabular}{ccccccc}
\toprule
 Name & Patch-size & $D_m$ & $\mathbf{N}_m$ & $\mathrm{dx,dy}$ & $\theta$ & $\beta_{min}$ \\
\midrule
1K7FS   & 1 & \checkmark & \checkmark  & \checkmark & \checkmark & 1e-3 \\
1K2FS   & 1 & \checkmark & -           & -          & \checkmark & 1e-3  \\
1K7FL   & 1 & \checkmark & \checkmark  & \checkmark & \checkmark & 1e-1 \\
1K2FL   & 1 & \checkmark & -           & -          & \checkmark & 1e-1 \\
5K7FS & 5 & \checkmark & \checkmark  & \checkmark & \checkmark & 1e-3 \\
5K2FS & 5 & \checkmark & -           & -          & \checkmark & 1e-3 \\
5K7FL & 5 & \checkmark & \checkmark  & \checkmark & \checkmark & 1e-1 \\
5K2FL & 5 & \checkmark & -           & -          & \checkmark & 1e-1 \\
\bottomrule
\end{tabular}
\caption{\textbf{Naming Table.} Description of different ablations for understanding the effect of different architectural and loss methods.}
\label{tab:archabl}
\end{table}

We first present results using a 1$\times$1 patch. Within this subset, we have four ablation results between the choice of regularizer and the number of input features. These results are summarized in \cref{tab:unc-sl-abalate-2d1}.

\begin{table}[ht]
\centering
\footnotesize
\setlength{\tabcolsep}{2pt}
\renewcommand{\arraystretch}{1.1}
\resizebox{\columnwidth}{!}{%
\begin{tabular}{lccccccc}
\toprule
Scene & Sensor & Map Loss & N & mP [\si{cm}] & mR [\si{cm}] & R [\%] & Depth L1 [\si{cm}]\\
\midrule
Office 0 & SL & 1K7FS & 10 & \fs 2.97 (0.19) & \fs 2.06 (0.02) & \fs 94.7 (0.3) & \fs 1.79 (0.03)\\
Office 1 & SL & 1K7FS & 10 & \degr 2.75 (0.19) & \fs 1.64 (0.04) & \fs 96.3 (0.3) & \fs 1.35 (0.05)\\
Room 2   & SL & 1K7FS & 10 & \degr 2.88 (0.12) & \degr 2.41 (0.02) & \degr 92.4 (0.2) & \degr 2.14 (0.05)\\
\midrule
Office 0 & SL & 1K7FL & 10 & \degr 3.24 (0.34) & \fs 2.07 (0.03) & \fs 94.7 (0.3) & \degr 2.73 (1.44)\\
Office 1 & SL & 1K7FL & 10 & \degr 2.80 (0.21) & \fs 1.65 (0.03) & 96.2 (0.3) & \fs 1.36 (0.03)\\
Room 2   & SL & 1K7FL &  9 & \degr 2.81 (0.06) & \textrm{2.39} (0.03) & \degr 92.5 (0.4) & \degr 2.12 (0.03)\\
\midrule
Office 0 & SL & 1K2FS & 10 & \fs 2.93 (0.13) & \degr 2.10 (0.02) & \degr 94.4 (0.2) & \degr 1.90 (0.03)\\
Office 1 & SL & 1K2FS & 10 & \fs 2.63 (0.20) & \fs 1.62 (0.03) & \fs 96.4 (0.3) & \degr 1.39 (0.05)\\
Room 2   & SL & 1K2FS & 10 & \degr 2.91 (0.14) & \textit{2.49} (0.02) & \degr 91.7 (0.4) & \degr 2.22 (0.02)\\
\midrule
Office 0 & SL & 1K2FL & 10 & \fs 2.85 (0.09) & \fs 2.05 (0.02) & \fs 94.8 (0.2) & \textrm{1.80} (0.03)\\
Office 1 & SL & 1K2FL & 10 & \degr 2.80 (0.15) & \fs 1.61 (0.03) & \fs 96.5 (0.3) & \degr 1.40 (0.10)\\
Room 2   & SL & 1K2FL & 10 & \fs 2.76 (0.11) & \fs 2.38 (0.03) & \textrm{92.6} (0.3) & \fs 2.09 (0.04)\\
\bottomrule
\end{tabular}}
\caption{\textbf{Patch Size 1$\times$1 Ablation.} \colorbox{colorFst}{\bf Green} denotes improvement over no uncertainty estimation \ie{} defaulting to \cite{zhu2022nice}. \colorbox{colorDeg}{Orange} shows degradation. No color denotes no change. ``1K2FL'' achieves the most consistent improvement across metrics. $N$ denotes the number of runs. The number in parenthesis denotes the standard deviation across the N runs.}
\label{tab:unc-sl-abalate-2d1}
\end{table}

Using the 1$\times$1 patch-based approach, we find improvement over some parameters and degradation in others. We find that the simplest architecture ``1K2FL'', employing two input features and a larger $\beta_{\textrm{min}}$, has one of the better performances within this subset of ablations. This method improves across eight metrics and observes degradation in two metrics. The remaining two metrics are within rounding error. With few input parameters and a larger regularizer, the chance of overfitting may be limited by this particular architecture, preventing the more wide-spread degradation we observe across other trials. 

We next present results using a 5$\times$5 patch. Within this subset, we have four ablation results between the use of regularizer and the number of input features. These results are summarized in \cref{tab:unc-sl-abalate-2d5}.

\begin{table}[ht]
\centering
\footnotesize
\setlength{\tabcolsep}{2pt}
\renewcommand{\arraystretch}{1.1}
\resizebox{\columnwidth}{!}
{
\begin{tabular}{cccccccc}
\toprule
Scene & Sensor & Map Loss & N & mP [\si{cm}] & mR [\si{cm}] & R [\%] & Depth L1 [\si{cm}]\\
\midrule
Office 0 & SL & 5K7FS & 10 & \degr 3.06 (0.09) & \textrm{2.08} (0.03) & \textrm{94.5} (0.4) & \degr 1.86 (0.02)\\
Office 1 & SL & 5K7FS & 10 & \degr 2.72 (0.27) & \fs 1.62 (0.03) & \fs 96.4 (0.3) & \degr 1.39 (0.06)\\
Room 2   & SL & 5K7FS & 10 & \degr 2.87 (0.17) & \degr 2.45 (0.01) & \degr 91.9 (0.2) & \degr 2.56 (1.25)\\
\midrule
Office 0 & SL & 5K7FL & 10 & \degr 3.08 (0.18) & \fs 2.06 (0.03) & \fs 94.7 (0.3) & \degr 1.82 (0.04)\\
Office 1 & SL & 5K7FL & 10 & \fs 2.68 (0.20) & \fs 1.61 (0.02) & \fs 96.6 (0.2) & \fs 1.36 (0.03)\\
Room 2   & SL & 5K7FL &  7 & \degr 2.85 (0.06) & \fs 2.37 (0.04) & \fs 92.8 (0.4) & \textrm{2.11} (0.01)\\
\midrule
Office 0 & SL & 5K2FS & 10 & \fs 2.87 (0.09) & \fs 2.06 (0.04) & \fs 94.7 (0.3) & \fs 1.79 (0.02)\\
Office 1 & SL & 5K2FS & 10 & \degr 2.75 (0.21) & \fs 1.61 (0.03) & \fs 96.5 (0.3) & \fs 1.36 (0.04)\\
Room 2   & SL & 5K2FS & 10 & \fs 2.79 (0.15) & \textrm{2.39} (0.02) & \textrm{92.6} (0.3) & \textrm{2.11} (0.05)\\
\midrule
Office 0 & SL & 5K2FL & 10 & \textrm{3.03} (0.19) & \fs 2.07 (0.02) & \fs 94.6 (0.2) & \textrm{1.80} (0.03)\\
Office 1 & SL & 5K2FL & 10 & \degr 2.84 (0.19) & \fs 1.60 (0.03) & \fs 96.7 (0.3) & \degr 1.39 (0.04)\\
Room 2   & SL & 5K2FL & 10 & \degr 2.81 (0.14) & \degr 2.40 (0.03) & \degr 92.4 (0.4) & \textrm{2.11} (0.02)\\
\bottomrule
\end{tabular}}
\caption{\textbf{Patch Size 5$\times$5 Ablation.} \colorbox{colorFst}{\bf Green} denotes improvement over no uncertainty estimation \ie{} defaulting to \cite{zhu2022nice}. \colorbox{colorDeg}{Orange} shows degradation. No color denotes no change. ``5K2FS'' achieves the most consistent improvement across metrics. $N$ denotes the number of runs. The number in parenthesis denotes the standard deviation across the N runs.}
\label{tab:unc-sl-abalate-2d5}
\end{table}

We find that two methods achieve positive improvement across a majority of metrics. Trials ``5K7FL'' and ``5K2FS'' both see improvements across eight metrics. ``5K2FS'' saw fewer metrics degrade in performance after discounting rounding errors. Overall, however, both these methods achieve marginal improvement over the baseline methods. The ``5K7FS'' trial experienced a strong outlier where the tracking error was high. This outlier run skewed the results in the Room 2 scene. The other three trials congregate around a similar performance cluster. We see again that the use of a larger regularizer $\beta_{\mathrm{min}}$ may be beneficial within the single sensor framework in improving metrics. When using a smaller regularizer, the inclusion of fewer features may improve results.

However, the results we have are inconclusive and the evidence is limited in the above claims. We see that that ``5K2FL'' appears to perform worse than ``5K2FS'', which contradicts our belief that a strong regularizer should be beneficial in 3D reconstruction. Amongst the architectures, we decide to use ''5K2FS`` as the architecture of choice. 

\subsection*{Ablation Statistical Significance} 
\label{sec:ablate-sig}

The default implementation of NICE-SLAM is non-deterministic due to the backward pass of the \texttt{\small grid\_sample()} function in PyTorch. This function is used for interpolating on the voxel grids of features. We note that a deterministic implementation of the backwards pass through should be possible, but that no plug-and-play implementation exists.

A first strategy to address the variance in output is simple aggregate statistics across a number of runs. A second strategy to determine the significance of our work is through an unpaired t-test (see Section G) that reports the likelihood that both results exhibit the same mean. Such tests give insight into the effectiveness of various implementations. 

We perform a cursory analysis on the total number of improved metrics in \cref{tab:unc-sl-abalate-2d5,tab:unc-sl-abalate-2d1}, showing the general trends of improvement. We now present the detailed results of the significance analysis of the various ablations in \cref{tab:unc-sl-abalate-2d5,tab:unc-sl-abalate-2d1}. The summary can be found in \cref{tab:ablate-stats}.

\begin{table}[ht]
\centering
\footnotesize
\setlength{\tabcolsep}{6pt}
\renewcommand{\arraystretch}{1.1}
\begin{tabular}{cccccc}
\toprule
Trial & Scene & P mP & P mR & P R & P Depth L1  \\ 
\midrule
& Office 0
& 65.8\% $\uparrow$ & \fs 2.7\% $\uparrow$ & 6.8\% $\uparrow$ & 36.9\% $\uparrow$   \\
1K7FS & Office 1  & 91.2\% $\downarrow$ & 62.4\% $\uparrow$ & 46.7\% $\uparrow$ & 41.1\% $\uparrow$  \\
& Room 2
& 33.6\% $\downarrow$ & \degr 4.1\% $\downarrow$ & 31.2\% $\downarrow$ & 22.4\% $\downarrow$  \\

\midrule
& Office 0 
& 20.1\% $\uparrow$ & \degr 1.2\% $\downarrow$ & 12.9\% $\downarrow$ & 0.0\% $\downarrow$ \\
1K2FS & Office 1  
& 21.0\% $\uparrow$ & 13.3\% $\uparrow$ & 13.9\% $\uparrow$ & 33.1\% $\downarrow$  \\
& Room 2 
& 11.3\% $\downarrow$ & \degr 0.0\% $\downarrow$ & \degr 0.0\% $\downarrow$ & \degr 0.0\% $\downarrow$  \\

\midrule
& Office 0 
& 6.7\% $\downarrow$ & 71.1\% $\uparrow$ & 15.9\% $\uparrow$ & 7.1\% $\downarrow$\\
1K7FL & Office 1  
& 51.3\% $\downarrow$ & 61.8\% $\uparrow$ & 89.0\% $\uparrow$ &42.3\% $\uparrow$  \\
& Room 2 
& 62.3\% $\downarrow$ & 94.4\% $\downarrow$ & 81.7\% $\downarrow$ & 61.9\% $\downarrow$ \\

\midrule
& Office 0 
& \fs 0.5\% $\uparrow$ & \fs 0.0\% $\uparrow$ & \fs 0.2\% $\uparrow$ & 78.0\% $\uparrow$  \\
1K2FL & Office 1  
& 39.5\% $\downarrow$ & 5.1\% $\uparrow$ & \fs 3.4\% $\uparrow$ & 35.7\% $\downarrow$  \\
& Room 2 
& 16.3\% $\uparrow$ & 47.9\% $\uparrow$ & 37.2\% $\uparrow$ & 20.2\% $\uparrow$ \\

\midrule
& Office 0 
& 24.9\% $\downarrow$ & 61.4\% $\downarrow$ & 78.4\% $\uparrow$ & \degr 0.0\% $\downarrow$ \\
5K7FS & Office 1  
& 87.5\% $\downarrow$ & 19.2\% $\uparrow$ & 12.7\% $\uparrow$ & 34.0\% $\downarrow$  \\
& Room 2 
& 18.3\% $\downarrow$ & \degr 0.0\% $\downarrow$ & \degr 0.0\% $\downarrow$ & 28.7\% $\downarrow$  \\

\midrule
& Office 0 
& \fs  0.7\% $\uparrow$ & 26.7\% $\uparrow$ & 10.2\% $\uparrow$ & 17.1\% $\uparrow$   \\
5K2FS & Office 1  
& 89.1\% $\downarrow$ & \fs 3.6\% $\uparrow$ & \fs 3.7\% $\uparrow$ & 57.9\% $\uparrow$ \\
& Room 2 
& 50.9\% $\uparrow$ & 63.8\% $\downarrow$ & 55.6\% $\uparrow$ & 88.8\% $\uparrow$ \\

\midrule
& Office 0 
& 29.6\% $\downarrow$ & 10.8\% $\uparrow$ & 14.8\% $\uparrow$ & 18.4\% $\downarrow$  \\
5K7FL & Office 1  
& 48.7\% $\uparrow$ & \fs 2.4\% $\uparrow$ & \fs 0.3\% $\uparrow$ & 80.1\% $\uparrow$ \\
& Room 2 
& 60.4\% $\downarrow$ & 30.6\% $\uparrow$ & 14.0\% $\uparrow$ & 67.2\% $\uparrow$ \\

\midrule
& Office 0 
& 75.0\% $\downarrow$ & 16.8\% $\uparrow$ & 8.6\% $\uparrow$ & 61.8\% $\uparrow$ 
\\
5K2FL & Office 1  
& 25.0\% $\downarrow$ & \fs 0.9\% $\uparrow$ & \fs 0.1\% $\uparrow$ & 14.3\% $\downarrow$  \\
& Room 2 
& 78.9\% $\downarrow$ & 22.5\% $\downarrow$ & 67.0\% $\downarrow$ & 44.4\% $\uparrow$  \\
\bottomrule
\end{tabular}
\caption{\textbf{Statistical t-Test.} Statistical significance of ablation differences based on Welch's t-test. \colorbox{colorFst}{\bf Green} denotes statistically significant improvements over no uncertainty estimation \ie defaulting to \cite{zhu2022nice}. \colorbox{colorDeg}{Orange} shows statistically significant degradations. No color means no statistically significant change (P $>$ 0.05). $\uparrow$ means improvement, $\downarrow$ means degradation.}
\label{tab:ablate-stats}
\end{table}

We can see that many of the results are not statistically significant. We present in \cref{tab:arch-net-sig} the number of significant improvements, the number of significant degradations, and the \emph{net total} of significant improvements.

\begin{table}[ht]
\centering
\footnotesize
\setlength{\tabcolsep}{2pt}
\renewcommand{\arraystretch}{1.1}
\centering
\footnotesize
\begin{tabular}{ccccccc}
\toprule
Run &  Patch-size & $\#$ Features & $\beta_{min}$ [$\si{m}$] & $\#$ Sig$\uparrow$ & $\#$ Sig$\downarrow$  & $\#$ Net Sig\\
\midrule
1K7FS   & 1 & 7 & 1e-3 & 1 & 1 & 0 \\
1K2FS   & 1 & 2 & 1e-3 & 0 & 5 & -5 \\
1K7FL   & 1 & 7 & 1e-1 & 0 & 0 & 0 \\
1K2FL   & 1 & 2 & 1e-1 & 3 & 0 & 3 \\
5K7FS & 5 & 7 & 1e-3 & 0 & 3 & -3 \\
5K2FS & 5 & 2 & 1e-3 & 3 & 0 & 3 \\
5K7FL & 5 & 7 & 1e-1 & 2 & 0 & 2 \\
5K2FL & 5 & 2 & 1e-1 & 2 & 0 & 2 \\
\bottomrule
\end{tabular}
\caption{\textbf{Statistical t-Test Summary.} Summary of the effect of different ablations in understanding different uncertainty-aware architectures.}
\label{tab:arch-net-sig}
\end{table}

The two best performing methods, ``1K2FL'' and ``5K2FS'', each have three significant improvements and no significant degradations.

\section{t-test}
\label{sec:t-test}
The unpaired t-test is a two sample location test that compares if two sample populations have the same mean. This analysis is performed by determining the statistic $t$ and the degrees-of-freedom $\nu$. Given the sample means $\overline{X}_{\{1,2\}}$ and the standard errors $s_{\bar{X}_{\{1,2\}}}$, $t$ and $\nu$ can be calculated using \cref{eq:t,eq:nu}.

\begin{align}
t \ &= \ {\frac {\Delta {\overline {X}}}{s_{\Delta {\bar {X}}}}} \ = \ {\frac {{\overline {X}}_{1}-{\overline {X}}_{2}}{\sqrt {{s_{{\bar {X}}_{1}}^{2}}+{s_{{\bar {X}}_{2}}^{2}}}}} \label{eq:t}
\\ 
\nu \ &\approx \ {\frac {\left(\;{\frac {s_{1}^{2}}{N_{1}}}\;+\;{\frac {s_{2}^{2}}{N_{2}}}\;\right)^{2}}{\quad {\frac {s_{1}^{4}}{N_{1}^{2}\nu _{1}}}\;+\;{\frac {s_{2}^{4}}{N_{2}^{2}\nu _{2}}}\quad }}
\label{eq:nu}
\end{align}

These values can then be used to identify the probability given by the Student's t-distribution that the two sample means are equal. If we assume significance at $P<0.05$, we can determine if an improvement\textemdash i.e. an increase in the mean performance\textemdash should be considered significant. 

\section{Experiments per Scene}
\label{sec:detexp}

In the main paper, we show the average over the test scenes \texttt{Office 0}, \texttt{Office 1} and \texttt{Room 2}. In the following, we show the per scene results for the same experiments conducted on the Replica dataset.

\subsection*{Replica Single-Sensor Mapping}
\cref{tab:map-ss-sl}, \cref{tab:map-ss-psm} and \cref{tab:map-ss-sgm} show the per scene results when only depth and ground truth poses are used.

\begin{table}[tb]
\centering
\setlength{\tabcolsep}{3pt}
\resizebox{\columnwidth}{!}
{
\begin{tabular}{l|llllll}
\cellcolor{gray}       & \cellcolor{gray}Depth L1$\downarrow$     & \cellcolor{gray}mP$\downarrow$  & \cellcolor{gray}mR$\downarrow$     & \cellcolor{gray}P$\uparrow$ & \cellcolor{gray}R$\uparrow$ & \cellcolor{gray}F$\uparrow$    \\
\multirow{-2}{*}{\cellcolor{gray} ${\text{Model}\downarrow | \text{Metric}\rightarrow}$} & \cellcolor{gray}[cm] & \cellcolor{gray}[cm] & \cellcolor{gray}[cm] & \cellcolor{gray}$[\%]$ & \cellcolor{gray}$[\%]$ & \cellcolor{gray}$[\%]$ \\\hline
\multicolumn{7}{c}{\emph{Depth + Ground Truth Poses Office 0}} \\ \hline
NICE-SLAM~\cite{zhu2022nice} &\fs \textrm{1.76}& \rd \textrm{2.26} & \fs \textrm{1.69} & \nd  \textrm{91.1} & \fs \textrm{94.3} & \nd \textrm{92.7} \\
NICE-SLAM+Pre &\nd \textrm{1.85} & \nd \textrm{2.23} & \rd \textrm{1.75} & \rd \textrm{90.9} & \rd \textrm{94.0} & \rd \textrm{92.4} \\
Ours &  \rd \textrm{1.86}& \fs \textrm{2.13} & \nd \textrm{1.74} &\fs \textrm{91.7} & \nd \textrm{94.2} & \fs \textrm{93.0} \\
\hline
\multicolumn{7}{c}{\emph{Depth + Ground Truth Poses Office 1}} \\ \hline
NICE-SLAM~\cite{zhu2022nice} & \nd \textrm{1.52} & \nd \textrm{2.36} & \fs \textrm{1.26} & \nd \textrm{90.8} & \fs \textrm{96.6} & \fs \textrm{93.6}\\
NICE-SLAM+Pre &\rd \textrm{1.57}&\nd \textrm{2.36} & \rd \textrm{1.30} & \rd \textrm{90.5} & \rd \textrm{96.3} & \rd \textrm{93.3} \\
Ours & \fs \textrm{1.50}& \fs \textrm{2.27} & \nd \textrm{1.28} & \fs \textrm{91.0} & \nd \textrm{96.5} & \fs \textrm{93.6} \\
\hline
\multicolumn{7}{c}{\emph{Depth + Ground Truth Poses Room 2}} \\ \hline
NICE-SLAM~\cite{zhu2022nice} &\fs \textrm{2.09}& \fs \textrm{2.07} & \fs \textrm{2.12} &  \fs \textrm{91.3} & \fs \textrm{90.9} & \fs \textrm{91.1} \\
NICE-SLAM+Pre &\rd \textrm{2.21}& \nd \textrm{2.16} & \rd \textrm{2.25} & \nd \textrm{90.6} & \rd \textrm{89.8} & \nd \textrm{90.2} \\
Ours & \nd \textrm{2.19}& \nd \textrm{2.16} & \nd \textrm{2.24} & \nd  \textrm{90.6} & \nd \textrm{89.9} & \nd \textrm{90.2} \\
\hline
\end{tabular}}
\caption{\textbf{Depth + Ground Truth Poses: SL Sensor.}}
\label{tab:map-ss-sl}
\end{table}

\begin{table}[tb]
\centering
\setlength{\tabcolsep}{3pt}
\resizebox{\columnwidth}{!}
{
\begin{tabular}{l|llllll}
\cellcolor{gray}       & \cellcolor{gray}Depth L1$\downarrow$     & \cellcolor{gray}mP$\downarrow$  & \cellcolor{gray}mR$\downarrow$     & \cellcolor{gray}P$\uparrow$ & \cellcolor{gray}R$\uparrow$ & \cellcolor{gray}F$\uparrow$    \\
\multirow{-2}{*}{\cellcolor{gray} ${\text{Model}\downarrow | \text{Metric}\rightarrow}$} & \cellcolor{gray}[cm] & \cellcolor{gray}[cm] & \cellcolor{gray}[cm] & \cellcolor{gray}$[\%]$ & \cellcolor{gray}$[\%]$ & \cellcolor{gray}$[\%]$ \\\hline
\multicolumn{7}{c}{\emph{Depth + Ground Truth Poses Office 0}} \\ \hline
NICE-SLAM~\cite{zhu2022nice} & \rd  \textrm{2.15}& \nd \textrm{2.32} & \rd  \textrm{2.18} &\nd \textrm{90.9} & \rd \textrm{89.9} & \rd \textrm{90.4} \\
NICE-SLAM+Pre &\fs  \textrm{1.96}& \nd \textrm{2.32} & \nd \textrm{2.15} &  \nd \textrm{90.9} & \nd \textrm{90.1} & \nd \textrm{90.5} \\
Ours & \nd \textrm{1.97}& \fs \textrm{2.28} & \fs \textrm{2.14} & \fs \textrm{91.2} &\fs  \textrm{90.2} & \fs \textrm{90.7} \\
\hline
\multicolumn{7}{c}{\emph{Depth + Ground Truth Poses Office 1}} \\ \hline
NICE-SLAM~\cite{zhu2022nice} &\nd \textrm{3.38}& \nd \textrm{3.46} & \rd \textrm{2.34} & \rd \textrm{83.9} & \rd \textrm{87.4} & \rd \textrm{85.6} \\
NICE-SLAM+Pre &\rd \textrm{3.75}& \rd \textrm{3.53} & \nd \textrm{2.31} & \nd \textrm{84.1} & \nd \textrm{87.8} & \nd \textrm{85.9} \\
Ours & \fs \textrm{3.01}& \fs \textrm{3.34} & \fs \textrm{2.28} & \fs \textrm{84.4} & \fs \textrm{87.9} & \fs \textrm{86.1} \\
\hline
\multicolumn{7}{c}{\emph{Depth + Ground Truth Poses Room 2}} \\ \hline
NICE-SLAM~\cite{zhu2022nice} &\rd \textrm{2.38}& \rd \textrm{2.17} & \rd \textrm{2.54} & \rd \textrm{91.5} & \rd \textrm{87.3} & \rd \textrm{89.3} \\
NICE-SLAM+Pre &\nd \textrm{2.30}& \fs \textrm{2.10} & \nd \textrm{2.48} & \fs \textrm{91.9} & \fs \textrm{88.0} & \fs \textrm{89.9} \\
Ours & \fs \textrm{2.27} & \nd \textrm{2.12} & \fs \textrm{2.47} & \fs \textrm{91.9} & \fs \textrm{88.0} & \fs \textrm{89.9} \\
\hline
\end{tabular}}
\caption{\textbf{Depth + Ground Truth Poses: PSMNet Sensor.}}
\label{tab:map-ss-psm}
\end{table}

\begin{table}[tb]
\centering
\setlength{\tabcolsep}{3pt}
\resizebox{\columnwidth}{!}
{
\begin{tabular}{l|llllll}
\cellcolor{gray}       & \cellcolor{gray}Depth L1$\downarrow$     & \cellcolor{gray}mP$\downarrow$  & \cellcolor{gray}mR$\downarrow$     & \cellcolor{gray}P$\uparrow$ & \cellcolor{gray}R$\uparrow$ & \cellcolor{gray}F$\uparrow$    \\
\multirow{-2}{*}{\cellcolor{gray} ${\text{Model}\downarrow | \text{Metric}\rightarrow}$} & \cellcolor{gray}[cm] & \cellcolor{gray}[cm] & \cellcolor{gray}[cm] & \cellcolor{gray}$[\%]$ & \cellcolor{gray}$[\%]$ & \cellcolor{gray}$[\%]$ \\\hline
\multicolumn{7}{c}{\emph{Depth + Ground Truth Poses Office 0}} \\ \hline
NICE-SLAM~\cite{zhu2022nice} & \rd \textrm{2.38}& \nd \textrm{2.75} & \rd \textrm{2.18} &\nd  \textrm{88.3} & \nd \textrm{91.1} & \rd \textrm{89.7} \\
NICE-SLAM+Pre & \fs \textrm{2.28} & \fs \textrm{2.70} & \fs \textrm{2.14} & \fs \textrm{88.5} & \nd \textrm{91.1} & \nd \textrm{89.8} \\
Ours & \nd \textrm{2.30}& \rd \textrm{2.76} & \nd \textrm{2.15} & \fs \textrm{88.5} & \fs \textrm{91.3} & \fs \textrm{89.9} \\
\hline
\multicolumn{7}{c}{\emph{Depth + Ground Truth Poses Office 1}} \\ \hline
NICE-SLAM~\cite{zhu2022nice} & \fs \textrm{1.76}& \nd \textrm{2.38} & \nd \textrm{1.63} &  \fs \textrm{91.5} & \nd \textrm{94.2} & \fs \textrm{92.8} \\
NICE-SLAM+Pre &\nd \textrm{1.79} & \fs \textrm{2.37} & \fs  \textrm{1.62} & \nd \textrm{91.1} & \fs \textrm{94.3} & \nd \textrm{92.7} \\
Ours & \nd \textrm{1.79}&\rd  \textrm{2.44} & \rd \textrm{1.64} &\nd \textrm{91.1} &\rd  \textrm{94.1} & \rd \textrm{92.6} \\
\hline
\multicolumn{7}{c}{\emph{Depth + Ground Truth Poses Room 2}} \\ \hline
NICE-SLAM~\cite{zhu2022nice} &\rd \textrm{2.90}& \rd \textrm{2.53} &\rd  \textrm{2.56} & \rd \textrm{88.8} & \rd \textrm{87.9} &\rd  \textrm{88.3} \\
NICE-SLAM+Pre &\fs \textrm{2.66} & \fs \textrm{2.40} & \fs \textrm{2.48} &\fs  \textrm{89.9} & \fs \textrm{88.8} & \fs \textrm{89.3}\\
Ours & \nd \textrm{2.72} & \nd \textrm{2.48} & \nd \textrm{2.50} & \nd  \textrm{89.2} &\nd  \textrm{88.4} & \nd \textrm{88.8}\\
\hline
\end{tabular}}
\caption{\textbf{Depth + Ground Truth Poses: SGM Sensor.}}
\label{tab:map-ss-sgm}
\end{table}

\subsection*{Replica Single-Sensor Tracking}
\cref{tab:map-ss-sl-t}, \cref{tab:map-ss-psm-t} and \cref{tab:map-ss-sgm-t} show the per scene results when camera pose estimation is enabled. We provide results with depth as the only input and with RGBD. 

\begin{table}[tb]
\centering
\setlength{\tabcolsep}{3pt}
\resizebox{\columnwidth}{!}
{
\begin{tabular}{l|lllllll}
\cellcolor{gray}       & \cellcolor{gray}Depth L1$\downarrow$     & \cellcolor{gray}mP$\downarrow$  & \cellcolor{gray}mR$\downarrow$     & \cellcolor{gray}P$\uparrow$ & \cellcolor{gray}R$\uparrow$ & \cellcolor{gray}F$\uparrow$  & \cellcolor{gray}ATE$\downarrow$  \\
\multirow{-2}{*}{\cellcolor{gray} ${\text{Model}\downarrow | \text{Metric}\rightarrow}$} & \cellcolor{gray}[cm] & \cellcolor{gray}[cm] & \cellcolor{gray}[cm] & \cellcolor{gray}$[\%]$ & \cellcolor{gray}$[\%]$ & \cellcolor{gray}$[\%]$ & \cellcolor{gray}[cm]\\\hline
\multicolumn{8}{c}{\emph{Depth + Tracking Office 0}} \\ \hline
NICE-SLAM~\cite{zhu2022nice} & \nd \textrm{12.00}& \nd \textrm{8.21} & \nd  \textrm{7.47}  & \nd \textrm{52.5} & \nd \textrm{56.9} & \nd \textrm{54.6} &  \fs \textrm{24.68}\\
NICE-SLAM+Pre & \fs \textrm{9.30}& \fs \textrm{7.81} & \fs \textrm{7.16}  & \fs \textrm{54.6} &\fs  \textrm{59.1} & \fs \textrm{56.8} & \nd \textrm{25.20}\\
Ours & \rd \textrm{13.81} & \rd \textrm{9.40} & \rd\textrm{9.21}  & \rd\textrm{51.4} & \rd\textrm{55.1} & \rd\textrm{53.2} & \rd\textrm{33.36}\\
\hline
\multicolumn{8}{c}{\emph{RGBD + Tracking Office 0}} \\ \hline
NICE-SLAM~\cite{zhu2022nice} & \nd \textrm{11.89}& \fs \textrm{8.31} &\fs  \textrm{7.39}  & \nd \textrm{51.5} & \nd \textrm{56.1} & \nd \textrm{53.7} & \fs  \textrm{25.72}\\
Ours &\fs \textrm{11.26}&\nd  \textrm{8.67} & \nd \textrm{7.96}  & \fs \textrm{52.1} & \fs \textrm{56.4} & \fs \textrm{54.2} & \nd \textrm{29.25}\\
\hline
\multicolumn{8}{c}{\emph{Depth + Tracking Office 1}} \\ \hline
NICE-SLAM~\cite{zhu2022nice} &\rd  \textrm{20.29}& \rd \textrm{17.19} & \rd \textrm{12.24}  & \rd \textrm{27.8} & \rd \textrm{32.3} & \rd \textrm{29.8} &\rd  \textrm{66.27}\\
NICE-SLAM+Pre &\fs \textrm{10.61} &\fs  \textrm{10.04} & \fs \textrm{8.65}  & \fs \textrm{44.6} & \fs \textrm{46.7} & \fs \textrm{45.6} &\fs  \textrm{31.95}\\
Ours &\nd \textrm{14.00}& \nd \textrm{11.49} & \nd \textrm{11.01}  &\nd  \textrm{40.0} & \nd \textrm{40.9} & \nd \textrm{40.4} &\nd  \textrm{38.52}\\
\hline
\multicolumn{8}{c}{\emph{RGBD + Tracking Office 1}} \\ \hline
NICE-SLAM~\cite{zhu2022nice} &\nd \textrm{16.67} & \nd \textrm{13.59} & \nd \textrm{11.15}  &\nd  \textrm{34.1} & \nd \textrm{36.6} & \nd \textrm{35.2}& \nd \textrm{52.17} \\
Ours &\fs  \textrm{11.50} & \fs \textrm{10.05} & \fs \textrm{10.42}  &\fs  \textrm{41.5} &\fs  \textrm{41.0} & \fs \textrm{41.2}&\fs \textrm{29.46} \\
\hline
\multicolumn{8}{c}{\emph{Depth + Tracking Room 2}} \\ \hline
NICE-SLAM~\cite{zhu2022nice} &\rd \textrm{17.12} & \rd \textrm{14.68} & \rd \textrm{11.24}  &\rd  \textrm{47.3} & \rd \textrm{49.8} & \rd \textrm{48.5} &  \rd \textrm{33.58}\\
NICE-SLAM+Pre &\fs \textrm{5.94}& \fs \textrm{6.10} & \fs \textrm{5.56}  & \fs \textrm{62.5} & \fs \textrm{64.8} & \fs \textrm{63.6} & \nd \textrm{21.18}\\
Ours & \nd \textrm{6.63} & \nd \textrm{7.32} & \nd \textrm{6.44} & \nd \textrm{57.6} & \nd \textrm{60.3} & \nd \textrm{58.9}& \fs \textrm{18.10}  \\
\hline
\multicolumn{8}{c}{\emph{RGBD + Tracking Room 2}} \\ \hline
NICE-SLAM~\cite{zhu2022nice} &\nd \textrm{14.59} & \nd \textrm{13.50} & \nd \textrm{9.99}  & \nd \textrm{47.3} & \nd \textrm{50.7} & \nd \textrm{48.9} & \nd \textrm{45.67}\\
Ours & \fs \textrm{6.59}& \fs \textrm{7.51} & \fs \textrm{6.98}  & \fs \textrm{55.7} & \fs \textrm{57.9} & \fs \textrm{56.8} &  \fs \textrm{18.20}\\
\hline
\end{tabular}}
\caption{\textbf{Depth and RGBD + Tracking: SL Sensor.}}
\label{tab:map-ss-sl-t}
\end{table}

\begin{table}[tb]
\centering
\setlength{\tabcolsep}{3pt}
\resizebox{\columnwidth}{!}
{
\begin{tabular}{l|lllllll}
\cellcolor{gray}       & \cellcolor{gray}Depth L1$\downarrow$     & \cellcolor{gray}mP$\downarrow$  & \cellcolor{gray}mR$\downarrow$     & \cellcolor{gray}P$\uparrow$ & \cellcolor{gray}R$\uparrow$ & \cellcolor{gray}F$\uparrow$  & \cellcolor{gray}ATE$\downarrow$  \\
\multirow{-2}{*}{\cellcolor{gray} ${\text{Model}\downarrow | \text{Metric}\rightarrow}$} & \cellcolor{gray}[cm] & \cellcolor{gray}[cm] & \cellcolor{gray}[cm] & \cellcolor{gray}$[\%]$ & \cellcolor{gray}$[\%]$ & \cellcolor{gray}$[\%]$ & \cellcolor{gray}[cm]\\\hline
\multicolumn{8}{c}{\emph{Depth + Tracking Office 0}} \\ \hline
NICE-SLAM~\cite{zhu2022nice} &\rd \textrm{11.08}& \rd \textrm{9.15} & \rd \textrm{6.93}  & \rd \textrm{49.1} & \rd \textrm{53.3} & \rd \textrm{51.1} & \rd \textrm{23.47}\\
NICE-SLAM+Pre & \fs \textrm{6.93}& \fs \textrm{6.31} & \fs \textrm{5.70}  & \fs \textrm{60.5} & \fs \textrm{64.0} & \fs \textrm{62.2} & \fs \textrm{19.94}\\
Ours & \nd \textrm{8.91} & \nd \textrm{7.41} & \nd \textrm{6.73}  &\nd  \textrm{53.0} & \nd \textrm{55.5} & \nd \textrm{54.2}& \nd \textrm{21.28}\\
\hline
\multicolumn{8}{c}{\emph{RGBD + Tracking Office 0}} \\ \hline
NICE-SLAM~\cite{zhu2022nice} & \fs \textrm{8.01}& \fs \textrm{7.04} & \fs \textrm{5.72}  &\fs  \textrm{55.9} & \fs \textrm{59.5} & \fs \textrm{57.6} & \fs \textrm{17.16}\\
Ours &\nd \textrm{8.43} & \nd \textrm{8.34} & \nd \textrm{6.95}  & \nd \textrm{49.9} &\nd  \textrm{55.8} & \nd \textrm{51.8} & \nd \textrm{23.06}\\
\hline
\multicolumn{8}{c}{\emph{Depth + Tracking Office 1}} \\ \hline
NICE-SLAM~\cite{zhu2022nice} &\nd \textrm{15.72}& \nd \textrm{15.39} & \nd \textrm{9.09}  &\nd  \textrm{34.9} & \rd \textrm{39.8} & \nd \textrm{37.0} & \nd \textrm{43.17}\\
NICE-SLAM+Pre & \rd \textrm{18.05}& \rd \textrm{31.22} & \rd \textrm{10.31}  &\rd  \textrm{28.7} & \nd \textrm{41.7} & \rd \textrm{33.7} & \rd \textrm{74.65}\\
Ours &\fs \textrm{8.83}& \fs \textrm{8.33} & \fs \textrm{7.88}  & \fs \textrm{44.8} &\fs  \textrm{44.5} & \fs \textrm{44.6} &\fs  \textrm{22.58}\\
\hline
\multicolumn{8}{c}{\emph{RGBD + Tracking Office 1}} \\ \hline
NICE-SLAM~\cite{zhu2022nice} &\nd \textrm{11.24} & \nd \textrm{10.58} & \nd \textrm{9.01} & \nd \textrm{38.2} &\nd  \textrm{40.2} & \nd \textrm{39.2} & \nd \textrm{25.08} \\
Ours &\fs  \textrm{7.32}& \fs \textrm{7.62} & \fs \textrm{7.24}  & \fs \textrm{48.0} &\fs  \textrm{48.0} &\fs  \textrm{48.0} & \fs \textrm{21.90}\\
\hline
\multicolumn{8}{c}{\emph{Depth + Tracking Room 2}} \\ \hline
NICE-SLAM~\cite{zhu2022nice} & \rd \textrm{5.14}& \rd \textrm{5.58} & \rd \textrm{5.47}  & \rd \textrm{61.3} &\rd  \textrm{61.1} & \rd \textrm{61.2} & \rd \textrm{17.05}\\
NICE-SLAM+Pre &\nd \textrm{4.71} & \nd \textrm{4.44} &\nd  \textrm{4.52}  & \nd \textrm{68.1} & \nd \textrm{67.4} &\nd  \textrm{67.7}& \nd \textrm{16.27} \\
Ours & \fs \textrm{4.42}& \fs \textrm{3.95} & \fs \textrm{3.98} & \fs \textrm{74.1} & \fs \textrm{72.7} & \fs \textrm{73.4}& \fs \textrm{14.22}  \\
\hline
\multicolumn{8}{c}{\emph{RGBD + Tracking Room 2}} \\ \hline
NICE-SLAM~\cite{zhu2022nice} & \nd \textrm{5.09}& \nd \textrm{5.82} & \nd \textrm{5.57}  & \nd \textrm{61.3} & \nd \textrm{61.0} & \nd \textrm{61.1} & \nd  \textrm{18.52}\\
Ours &\fs  \textrm{3.70}& \fs \textrm{3.33} & \fs \textrm{3.60}  & \fs \textrm{78.8} & \fs \textrm{76.4} & \fs \textrm{77.5} &  \fs \textrm{11.80}\\
\hline
\end{tabular}}
\caption{\textbf{Depth and RGBD + Tracking: PSMNet Sensor.}}
\label{tab:map-ss-psm-t}
\end{table}

\begin{table}[tb]
\centering
\setlength{\tabcolsep}{3pt}
\resizebox{\columnwidth}{!}
{
\begin{tabular}{l|lllllll}
\cellcolor{gray}       & \cellcolor{gray}Depth L1$\downarrow$     & \cellcolor{gray}mP$\downarrow$  & \cellcolor{gray}mR$\downarrow$     & \cellcolor{gray}P$\uparrow$ & \cellcolor{gray}R$\uparrow$ & \cellcolor{gray}F$\uparrow$  & \cellcolor{gray}ATE$\downarrow$  \\
\multirow{-2}{*}{\cellcolor{gray} ${\text{Model}\downarrow | \text{Metric}\rightarrow}$} & \cellcolor{gray}[cm] & \cellcolor{gray}[cm] & \cellcolor{gray}[cm] & \cellcolor{gray}$[\%]$ & \cellcolor{gray}$[\%]$ & \cellcolor{gray}$[\%]$ & \cellcolor{gray}[cm]\\\hline
\multicolumn{8}{c}{\emph{Depth + Tracking Office 0}} \\ \hline
NICE-SLAM~\cite{zhu2022nice} &\rd \textrm{9.75} & \rd \textrm{8.21} & \rd \textrm{6.75}  &\rd  \textrm{49.3} & \rd \textrm{55.5} & \rd \textrm{52.2} & \nd \textrm{19.17}\\
NICE-SLAM+Pre &\fs \textrm{5.88} &\fs  \textrm{5.96} & \fs \textrm{4.63}  & \fs \textrm{61.1} & \fs \textrm{67.5} & \fs \textrm{64.1} & \fs \textrm{12.62}\\
Ours & \nd \textrm{9.32}& \nd \textrm{7.08} & \nd \textrm{5.83}  & \nd \textrm{55.3} & \nd \textrm{61.8} & \nd \textrm{58.4} & \rd \textrm{19.93}\\
\hline
\multicolumn{8}{c}{\emph{RGBD + Tracking Office 0}} \\ \hline
NICE-SLAM~\cite{zhu2022nice} & \nd \textrm{8.84}& \nd \textrm{7.37} &\nd  \textrm{6.09}  & \nd \textrm{54.5} & \nd \textrm{60.5} & \nd \textrm{57.3} &  \nd \textrm{17.86}\\
Ours & \fs \textrm{6.20}& \fs \textrm{6.15} & \fs \textrm{4.87}  & \fs \textrm{59.6} &\fs  \textrm{65.0} & \fs \textrm{62.2} & \fs \textrm{14.05}\\
\hline
\multicolumn{8}{c}{\emph{Depth + Tracking Office 1}} \\ \hline
NICE-SLAM~\cite{zhu2022nice} &\fs \textrm{16.63}& \fs \textrm{13.48} &\rd \textrm{10.46}  &\fs  \textrm{34.3} & \nd \textrm{36.0} & \nd \textrm{35.0} & \fs \textrm{44.13}\\
NICE-SLAM+Pre &\rd \textrm{38.77}& \rd \textrm{35.14} & \fs \textrm{9.66}  & \rd \textrm{29.2} & \fs \textrm{44.3} & \fs \textrm{35.1} & \rd \textrm{60.76}\\
Ours & \nd \textrm{17.56} & \nd \textrm{16.68} & \nd \textrm{10.19} & \nd \textrm{29.5} & \rd \textrm{34.6} & \rd \textrm{31.7} & \nd \textrm{52.72} \\
\hline
\multicolumn{8}{c}{\emph{RGBD + Tracking Office 1}} \\ \hline
NICE-SLAM~\cite{zhu2022nice} & \nd  \textrm{14.74}& \fs  \textrm{17.86} & \nd \textrm{9.25} & \nd \textrm{32.3} & \nd \textrm{37.0} & \nd \textrm{33.9} & \fs \textrm{42.47}\\
Ours & \fs \textrm{12.06}& \nd \textrm{22.48} & \fs \textrm{8.43}  & \fs \textrm{33.6} & \fs \textrm{42.9} & \fs \textrm{37.1} & \nd \textrm{54.61}\\
\hline
\multicolumn{8}{c}{\emph{Depth + Tracking Room 2}} \\ \hline
NICE-SLAM~\cite{zhu2022nice} & \nd \textrm{9.71} & \nd \textrm{8.95} & \nd \textrm{6.05}  & \rd \textrm{54.4} & \rd \textrm{60.3} & \rd \textrm{57.1} &\nd  \textrm{28.87}\\
NICE-SLAM+Pre &\rd \textrm{12.23}& \rd \textrm{11.95} &\rd  \textrm{6.41} & \nd \textrm{56.5} & \nd \textrm{61.0} & \nd \textrm{58.4}&  \rd \textrm{44.04}  \\
Ours &\fs \textrm{4.92} & \fs \textrm{4.36} & \fs \textrm{3.72}  & \fs \textrm{73.1} & \fs \textrm{75.3} & \fs \textrm{74.2}& \fs \textrm{14.69} \\
\hline
\multicolumn{8}{c}{\emph{RGBD + Tracking Room 2}} \\ \hline
NICE-SLAM~\cite{zhu2022nice} & \nd \textrm{6.14}& \nd \textrm{5.88} & \nd \textrm{5.12}  & \nd \textrm{63.5} &\nd  \textrm{66.0} & \nd \textrm{64.7} &  \nd \textrm{19.35}\\
Ours & \fs\textrm{5.11}& \fs\textrm{4.39} & \fs\textrm{4.09}  & \fs\textrm{75.1} & \fs\textrm{75.6} & \fs\textrm{75.3} & \fs\textrm{13.58}\\
\hline
\end{tabular}
}
\caption{\textbf{Depth and RGBD + Tracking: SGM Sensor.}}
\label{tab:map-ss-sgm-t}
\end{table}

\subsection*{Replica Multi-Sensor Mapping}
\cref{tab:map-ms-sl-psm} and \cref{tab:map-ms-psm-sgm} show the per scene results for the multi-sensor experiments when we only provide depth and ground truth poses. \cref{tab:psmnet_sgm_track} and \cref{tab:sl_psm_track} show the per scene results when tracking is enabled.

\begin{table}[tb]
\centering
\setlength{\tabcolsep}{3pt}
\resizebox{\columnwidth}{!}
{
\begin{tabular}{l|llllll}
\cellcolor{gray}       & \cellcolor{gray}Depth L1$\downarrow$     & \cellcolor{gray}mP$\downarrow$  & \cellcolor{gray}mR$\downarrow$     & \cellcolor{gray}P$\uparrow$ & \cellcolor{gray}R$\uparrow$ & \cellcolor{gray}F$\uparrow$ \\
\multirow{-2}{*}{\cellcolor{gray} ${\text{Model}\downarrow | \text{Metric}\rightarrow}$} & \cellcolor{gray}[cm] & \cellcolor{gray}[cm] & \cellcolor{gray}[cm] & \cellcolor{gray}$[\%]$ & \cellcolor{gray}$[\%]$ & \cellcolor{gray}$[\%]$ \\\hline
\multicolumn{7}{c}{\emph{SL + PSMNet Office 0}} \\ \hline
NICE-SLAM~\cite{zhu2022nice} & \rd \textrm{1.86}& \rd \textrm{2.24} & \fs \textrm{1.66} & \rd \textrm{91.7} & \fs \textrm{93.6} & \nd \textrm{92.6} \\
NICE-SLAM+Pre & \fs \textrm{1.52}& \nd \textrm{2.07} & \nd \textrm{1.68} & \nd \textrm{92.1} & \rd \textrm{93.1} & \nd \textrm{92.6} \\
Ours & \nd \textrm{1.58}  & \fs \textrm{2.02} & \nd \textrm{1.68} & \fs \textrm{92.2} & \nd \textrm{93.3} & \fs \textrm{92.7} \\
\hline
\multicolumn{7}{c}{\emph{SL + PSMNet Office 1}} \\ \hline
NICE-SLAM~\cite{zhu2022nice} &\rd \textrm{1.66} & \rd \textrm{2.43} & \rd \textrm{1.64} & \nd \textrm{89.7} & \rd \textrm{92.9} & \nd \textrm{91.3} \\
NICE-SLAM+Pre &\fs \textrm{1.57}& \nd \textrm{2.40} & \nd \textrm{1.62} & \rd \textrm{89.5} & \nd \textrm{93.0} & \rd \textrm{91.2} \\
Ours & \nd   \textrm{1.58}& \fs \textrm{2.26} & \fs \textrm{1.48} &\fs \textrm{90.5} & \fs \textrm{94.0} & \fs \textrm{92.2} \\
\hline
\multicolumn{7}{c}{\emph{SL + PSMNet Room 2}} \\ \hline
NICE-SLAM~\cite{zhu2022nice} & \fs \textrm{1.75}& \rd \textrm{1.82} & \fs \textrm{2.04} & \nd \textrm{93.7} & \fs \textrm{91.6} & \fs \textrm{92.7} \\
NICE-SLAM+Pre & \nd \textrm{1.78}& \nd \textrm{1.79} & \fs \textrm{2.04} &\nd \textrm{93.7} &\nd \textrm{91.3} & \nd \textrm{92.5} \\
Ours & \rd \textrm{1.79}& \fs \textrm{1.78} & \nd \textrm{2.09} & \fs \textrm{93.9} & \rd \textrm{91.0} & \rd \textrm{92.4} \\
\hline
\end{tabular}
}
\caption{\textbf{Depth + Ground Truth Poses: SL+PSMNet Sensor Fusion.}}
\label{tab:map-ms-sl-psm}
\end{table}

\begin{table}[tb]
\centering
\setlength{\tabcolsep}{3pt}
\resizebox{\columnwidth}{!}
{
\begin{tabular}{l|llllll}
\cellcolor{gray}       & \cellcolor{gray}Depth L1$\downarrow$     & \cellcolor{gray}mP$\downarrow$  & \cellcolor{gray}mR$\downarrow$     & \cellcolor{gray}P$\uparrow$ & \cellcolor{gray}R$\uparrow$ & \cellcolor{gray}F$\uparrow$ \\
\multirow{-2}{*}{\cellcolor{gray} ${\text{Model}\downarrow | \text{Metric}\rightarrow}$} & \cellcolor{gray}[cm] & \cellcolor{gray}[cm] & \cellcolor{gray}[cm] & \cellcolor{gray}$[\%]$ & \cellcolor{gray}$[\%]$ & \cellcolor{gray}$[\%]$ \\\hline
\multicolumn{7}{c}{\emph{SGM + PSMNet Office 0}} \\ \hline
NICE-SLAM~\cite{zhu2022nice} & \fs \textrm{1.74} & \nd \textrm{2.29} & \rd \textrm{1.88}  & \fs \textrm{91.0} & \fs \textrm{91.8} & \fs \textrm{91.4} \\
NICE-SLAM+Pre & \rd \textrm{1.78} & \rd \textrm{2.37} & \nd \textrm{1.92} & \nd \textrm{90.5} & \nd \textrm{91.5} & \rd \textrm{91.0} \\
Ours & \nd \textrm{1.76} & \fs \textrm{2.26} & \fs \textrm{1.93} & \fs \textrm{91.0} & \rd \textrm{91.4} & \nd \textrm{91.2} \\
\hline
\multicolumn{7}{c}{\emph{SGM + PSMNet Office 1}} \\ \hline
NICE-SLAM~\cite{zhu2022nice} & \nd \textrm{2.34} & \fs \textrm{2.78} & \fs \textrm{1.84} & \fs \textrm{87.7} & \fs \textrm{91.1} & \fs \textrm{89.4} \\
NICE-SLAM+Pre &\rd \textrm{2.80} & \rd  \textrm{3.02} & \rd \textrm{1.87} & \rd \textrm{86.3} & \rd \textrm{90.8} & \rd \textrm{88.5} \\
Ours & \fs \textrm{2.15}& \nd \textrm{2.85} & \nd \textrm{1.85} & \nd \textrm{86.6} & \fs \textrm{91.1} & \nd \textrm{88.8} \\
\hline
\multicolumn{7}{c}{\emph{SGM + PSMNet Room 2}} \\ \hline
NICE-SLAM~\cite{zhu2022nice} & \nd \textrm{1.99}& \nd \textrm{1.95} & \rd \textrm{2.26} & \nd \textrm{92.9} & \rd \textrm{89.7} & \nd \textrm{91.3} \\
NICE-SLAM+Pre & \fs \textrm{1.97}& \fs  \textrm{1.93} & \fs \textrm{2.24} & \fs \textrm{93.0} & \fs \textrm{90.0} & \fs \textrm{91.4} \\
Ours & \rd \textrm{2.00} & \rd \textrm{1.96} & \nd \textrm{2.25} & \rd \textrm{92.8} & \nd \textrm{89.8} & \nd \textrm{91.3} \\
\hline
\end{tabular}
}
\caption{\textbf{Depth + Ground Truth Poses: SGM+PSMNet Sensor Fusion.}}
\label{tab:map-ms-psm-sgm}
\end{table}

\begin{table}[tb]
\centering
\setlength{\tabcolsep}{3pt}
\resizebox{\columnwidth}{!}
{
\begin{tabular}{l|lllllll}
\cellcolor{gray}       & \cellcolor{gray}Depth L1$\downarrow$     & \cellcolor{gray}mP$\downarrow$  & \cellcolor{gray}mR$\downarrow$     & \cellcolor{gray}P$\uparrow$ & \cellcolor{gray}R$\uparrow$ & \cellcolor{gray}F$\uparrow$ & \cellcolor{gray}ATE$\downarrow$\\
\multirow{-2}{*}{\cellcolor{gray} ${\text{Model}\downarrow | \text{Metric}\rightarrow}$} & \cellcolor{gray}[cm] & \cellcolor{gray}[cm] & \cellcolor{gray}[cm] & \cellcolor{gray}$[\%]$ & \cellcolor{gray}$[\%]$ & \cellcolor{gray}$[\%]$ & \cellcolor{gray}[cm]\\\hline
\multicolumn{8}{c}{\emph{PSMNet + SGM Office 0}} \\ \hline
NICE-SLAM~\cite{zhu2022nice} & \nd 4.20 & \rd 4.83 & \nd 4.03 & \nd 69.07 & \nd 72.41 & \nd 70.69 & \nd 12.99 \\
NICE-SLAM+Pre & \fs 2.73 & \fs 3.55 & \fs 3.49 & \fs 79.05 & \fs 78.89 & \fs 78.97 & \fs 10.90 \\
Ours & \rd 4.21  & \nd 4.82 & \rd 4.09 & \rd 67.70 & \rd 70.54 & \rd 69.07 & \rd 13.72 \\
\hline
\multicolumn{8}{c}{\emph{PSMNet + SGM Office 1}} \\ \hline
NICE-SLAM~\cite{zhu2022nice} & \nd 21.52 & \nd 31.35 & \rd 13.06 & \nd 27.73 & \rd 32.05 & \nd 29.34 & \nd 67.72 \\
NICE-SLAM+Pre & \rd 28.49 & \rd 34.42 & \nd 11.73 & \rd 21.42 & \nd 36.60 & \rd 26.97 & \rd 83.36 \\
Ours & \fs 5.44  & \fs 6.45 & \fs 6.00 & \fs 55.50 & \fs 54.08 & \fs 54.78 & \fs 34.48 \\
\hline
\multicolumn{8}{c}{\emph{PSMNet + SGM Room 2}} \\ \hline
NICE-SLAM~\cite{zhu2022nice} & \rd 15.02 & \rd 14.11 & \rd 6.42 & \rd 56.77 & \rd 61.89 & \rd 58.39 & \rd 40.90 \\
NICE-SLAM+Pre & \nd 2.63  & \nd 2.80 & \nd 3.14 & \nd 85.60 & \nd 82.37 & \nd 83.95 & \nd 12.38 \\
Ours & \fs 2.75 & \fs 2.54 & \fs 2.97 & \fs 87.71 & \fs 83.28 & \fs 85.43 & \fs 11.44 \\
\hline
\multicolumn{8}{c}{\emph{PSMNet + SGM Overall}} \\ \hline
NICE-SLAM~\cite{zhu2022nice} & \rd 13.58 & \rd 16.76 & \rd 7.84 & \rd 51.19 & \rd 55.45 & \rd 52.81 & \rd 40.37 \\
NICE-SLAM+Pre & \nd 11.29  & \nd 13.59 & \nd 6.12 & \nd 62.02 & \nd 65.95 & \nd 63.30 & \nd 35.55 \\
Ours & \fs 4.13 & \fs 4.60 & \fs 4.35 & \fs 70.30 & \fs 69.30 & \fs 69.76 & \fs 19.88 \\
\hline
\end{tabular}
}
\caption{\textbf{Depth + Tracking: PSMNet+SGM Sensor Fusion.} Average of 5 runs.}
\label{tab:psmnet_sgm_track}
\end{table}

\begin{table}[tb]
\centering
\setlength{\tabcolsep}{3pt}
\resizebox{\columnwidth}{!}
{
\begin{tabular}{l|lllllll}
\cellcolor{gray}       & \cellcolor{gray}Depth L1$\downarrow$     & \cellcolor{gray}mP$\downarrow$  & \cellcolor{gray}mR$\downarrow$     & \cellcolor{gray}P$\uparrow$ & \cellcolor{gray}R$\uparrow$ & \cellcolor{gray}F$\uparrow$ & \cellcolor{gray}ATE$\downarrow$\\
\multirow{-2}{*}{\cellcolor{gray} ${\text{Model}\downarrow | \text{Metric}\rightarrow}$} & \cellcolor{gray}[cm] & \cellcolor{gray}[cm] & \cellcolor{gray}[cm] & \cellcolor{gray}$[\%]$ & \cellcolor{gray}$[\%]$ & \cellcolor{gray}$[\%]$ & \cellcolor{gray}[cm]\\\hline
\multicolumn{8}{c}{\emph{SL + PSMNet Office 0}} \\ \hline
NICE-SLAM~\cite{zhu2022nice} & \fs 3.72 & \fs 3.97 & \fs 3.43 & \fs 75.47 & \fs 77.23 & \fs 76.34 & \fs 14.00 \\
NICE-SLAM+Pre & \nd 4.63 & \nd 5.32 & \rd 4.89 & \nd 66.05 & \nd 68.27 & \nd 67.14 & \rd 20.48 \\
Ours & \rd 5.58  & \rd 5.38 & \nd 4.71 & \rd 64.04 & \rd 66.12 & \rd 65.06 & \nd 16.32 \\
\hline
\multicolumn{8}{c}{\emph{SL + PSMNet Office 1}} \\ \hline
NICE-SLAM~\cite{zhu2022nice} & \rd 4.77 & \rd 9.16 & \rd 8.01 & \nd 49.46 & \nd 50.36 & \nd 49.88 & \rd 28.58 \\
NICE-SLAM+Pre & \fs 3.41 & \fs 4.94 & \fs 4.13 & \fs 70.72 & \fs 70.78 & \fs 70.74 & \fs 23.89 \\
Ours & \nd 3.95  & \nd 8.69 & \nd 7.36 & \rd 44.44 & \rd 46.74 & \rd 45.54 & \nd 26.09 \\
\hline
\multicolumn{8}{c}{\emph{SL + PSMNet Room 2}} \\ \hline
NICE-SLAM~\cite{zhu2022nice} & \rd 5.61 & \rd 6.03 & \rd 5.78 & \rd 65.90 & \rd 66.17 & \rd 66.01 & \rd 23.90 \\
NICE-SLAM+Pre & \nd 2.66  & \nd 3.62 & \nd 3.88 & \nd 74.83 & \nd 73.64 & \nd 74.23 & \nd 16.93 \\
Ours & \fs 2.09 & \fs 2.36 & \fs 2.64 & \fs 90.14 & \fs 86.91 & \fs 88.49 & \fs 8.18 \\
\hline
\multicolumn{8}{c}{\emph{SL + PSMNet Overall}} \\ \hline
NICE-SLAM~\cite{zhu2022nice} & \rd 4.70 & \rd 6.39 & \rd 5.74 & \rd 63.61 & \rd 64.58 & \rd 64.08 & \rd 22.16 \\
NICE-SLAM+Pre & \fs 3.57  & \fs 4.63 & \fs 4.30 & \fs 70.53 & \fs 70.90 & \fs 70.70 & \nd 20.43 \\
Ours & \nd 3.87 & \nd 5.48 & \nd 4.90 & \nd 66.20 & \nd 66.59 & \nd 66.36 & \fs 16.86 \\
\hline
\end{tabular}
}
\caption{\textbf{Depth + Tracking: SL+PSMNet Sensor Fusion.} Average of 5 runs.}
\label{tab:sl_psm_track}
\end{table}